\documentclass[10pt,twocolumn,letterpaper]{article}
\usepackage{makecell} 
\usepackage{graphicx}
\usepackage{multirow}
\usepackage{capt-of}  
\usepackage{cuted}     
\usepackage[table]{xcolor}
\usepackage{tabularx,booktabs,array}
\usepackage{pifont}
\usepackage{subcaption}
\usepackage{marvosym}
\usepackage{cite}
\newcolumntype{Y}{>{\centering\arraybackslash}X}

\usepackage{cvpr}    

\definecolor{cvprblue}{rgb}{0.21,0.49,0.74}
\usepackage[pagebackref,breaklinks,colorlinks,allcolors=cvprblue]{hyperref}

\def\paperID{9841}
\def\confName{CVPR}
\def\confYear{2026}

\title{Towards Cross-View Point Correspondence in Vision-Language Models}

\author{
 \small Yipu Wang$^{1,2*}$, Yuheng Ji$^{1,3*}$, Yuyang Liu$^{1,3*}$, Enshen Zhou$^{4}$, Ziqiang Yang$^{5}$,Yuxuan Tian$^{1,3}$, Ziheng Qin$^{1,3}$,\\
 \small Yue Liu$^{6}$, Huajie Tan$^{7}$,
  Cheng Chi$^{8}$, Zhiyuan Ma$^{9}$, Daniel Dajun Zeng$^{1,2,3}$, Xiaolong Zheng$^{1,2,3,\text{\Letter}}$ \\
\\
$^1$ \small Institute of Automation, Chinese Academy of Sciences\\
$^2$ \small School of Advanced Interdisciplinary Sciences, University of Chinese Academy of Sciences \\
$^3$ \small School of Artificial Intelligence, University of Chinese Academy of Sciences
$^4$ \small Beihang University \\
$^5$ \small Jilin University 
$^6$ \small National University of Singapore
$^7$ \small Peking University\\ 
$^8$ \small Beijing Academy of Artificial Intelligence
$^9$ \small Huazhong University of Science And Technology
}

\begin{document}
\twocolumn[{%
\maketitle
\vspace{-0.9cm}
\begin{center}
    \centering
    \captionsetup{type=figure}
    \includegraphics[width=1\linewidth]{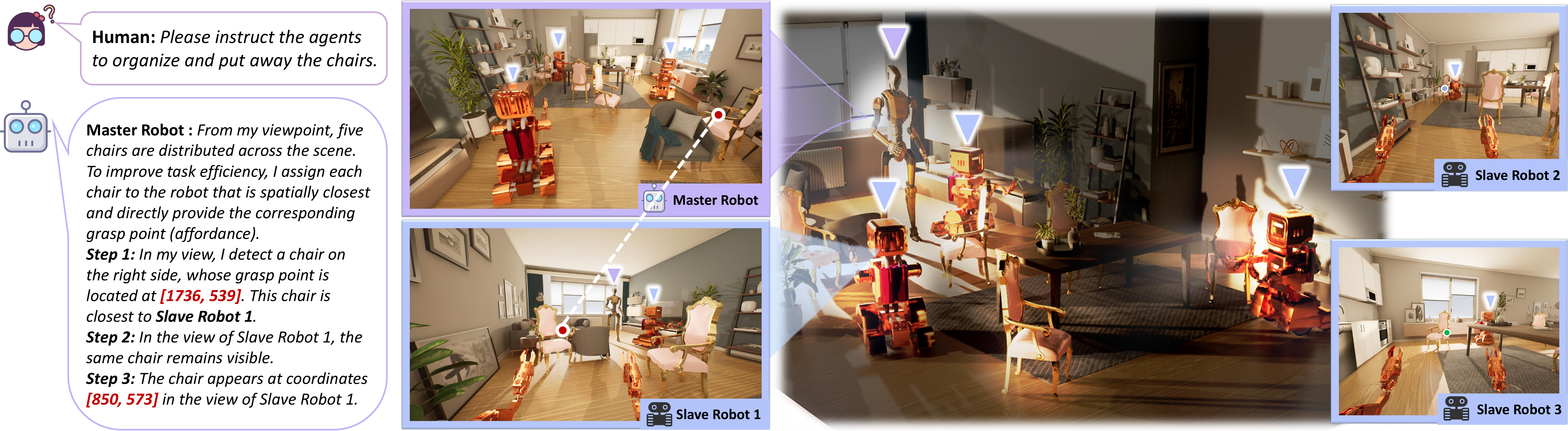}
    \captionof{figure}{\textbf{Cross-View Point Correspondence (CVPC) builds point-level geometric correspondence across views.} It involves: \textbf{(a)} understanding spatial instructions and locating targets (e.g., ground the location where the chair can be grasped); \textbf{(b)} reasoning about targets visibility (e.g., determining whether the red point is visible from slave robot 1’s view); \textbf{(c)} establishing point correspondence across views (e.g., corresponding this point from the master to slave robot 1). Such CVPC capability is pivotal for embodied robots to comprehend spatial layouts and execute downstream tasks. (e.g., multi-agent chair organization).} 
    \label{fig:teaser}
\end{center}
}]

\let\thefootnote\relax\footnotetext{$^{*}$ Equal contribution.}
\let\thefootnote\relax\footnotetext{$^{\text{\Letter}}$ Corresponding author.}

\maketitle

\begin{abstract}
Cross-view correspondence is a fundamental capability for spatial understanding and embodied AI.
However, it is still far from being realized in Vision-Language Models (VLMs), especially in achieving precise point-level correspondence, which is crucial for precise affordance interaction.
So we propose the \textbf{Cross-View Point Correspondence (CVPC)} task and \textbf{CrossPoint-Bench}, a comprehensive benchmark with hierarchical design, inspired by the human cognitive process of ``perceive'', ``reason'', and ``correspond''.
Our evaluation shows the state-of-the-art models (e.g.,  Gemini-2.5-Pro) still fall far behind humans, with a gap of over 54.65\% in overall accuracy, exposing a challenge in transitioning from coarse-grained judgement to fine-grained coordinate prediction.
To address this problem, we construct \textbf{CrossPoint-378K}, a dataset with 378K question-answering pairs across 900 scenes, focused on actionable affordance regions that better reflect real-world manipulation and interaction scenarios.
Furthermore, we propose \textbf{CroPond} that trained on the CrossPoint-378K dataset. Our CroPond achieves state-of-the-art performance on CrossPoint-Bench, surpassing Gemini-2.5-Pro by 39.7\% accuracy, which offers a foundation for advancing future work on cross-view correspondence.
The benchmark, dataset, and model are publicly available at https://github.com/WangYipu2002/CrossPoint.
\end{abstract}

\section{Introduction}
\label{sec:intro}

\begin{table}[htbp]
  \centering
  \caption{\textbf{Comparison of cross-view benchmarks across key evaluation dimensions.} Cross. denotes Cross-View; Aff. denotes Affordance Awareness; Occ. denotes Occlusion Robustness. Corr. denotes Correspondence Granularity.}
  \setlength{\tabcolsep}{6pt}
  \renewcommand{\arraystretch}{1.1}
  \fontsize{8}{11}\selectfont 
  \begin{tabular}{lccccc}
        \toprule
        \textbf{Benchmark}                                  & \textbf{Cross.}   &\textbf{Aff.}  &\textbf{Occ.}  &\textbf{Corr.} \\
        \midrule
        All-Angles Bench~\cite{yeh2025seeing-All-Angles}    & \ding{51}         & \ding{55}     & \ding{55}     & \ding{55} \\
        MMSI-Bench~\cite{yang2025mmsi}                      & \ding{51}         & \ding{55}     & \ding{55}     & \ding{55}\\
        University-1652~\cite{zheng2020university}          & \ding{51}         & \ding{55}     & \ding{55}     & \textit{image-choice} \\
        BLINK~\cite{fu2024blink}                            & \textit{partial}  & \ding{55}     & \ding{55}     & \textit{point-choice}\\
        SPAR-Bench~\cite{zhang2025from}                     & \ding{51}         & \ding{55}     & \ding{55}     & \textit{bbox-choice} \\
        \textbf{CrossPoint-Bench}                           & \ding{51}         & \ding{51}     & \ding{51}     & \textit{point-prediction}\\
        \bottomrule
  \end{tabular}
  \label{tab:benchmark_comparison}
  \vspace{-1em}
\end{table}

Humans possess a remarkable ability for cross-view spatial correspondence: given multiple viewpoints, we can reliably infer where the same physical point lies across views~\cite{shepard1971mental,wang2002human}. This capability underpins embodied tasks such as navigation, grasping, and multi-agent collaboration~\cite{li2024comprehensive,chen2025survey}, forming a critical bridge from visual understanding to physical execution. As illustrated in Fig.~\ref{fig:teaser}, in a ``chairs organization'' scenario a master robot identifies manipulable points on the chairs and maps them into the egocentric views of other agents to coordinate their actions. Despite progress in spatial reasoning~\cite{chen2024spatialvlm,cheng2024spatialrgpt,zhang2024groundhog,he2024improved,tanreason,ji2025robobrain,team2025robobrain,an2025llava}, current Vision–Language Models (VLMs) still lack robust cross-view correspondence. Their understanding remains tied to static single views, creating a reality gap where models may identify a plausible region but fail to predict the precise point needed for real robot motion. In everyday interaction, people naturally refer at the point level, for example saying ``put this one here'' or ``grab that handle''. In practical deployments, downstream agents likewise require only point-level guidance, while a master agent handles task interpretation and generates these targets.

To systematically address this limitation and drive progress, we introduce and formalize the task of \textbf{Cross-View Point Correspondence (CVPC)}. The CVPC task requires models to go beyond coarse region-level correspondence and predict the precise geometric coordinates of a target point across different views. This ability is essential for collaborative manipulation and other real-world applications. Although prior studies have explored cross-view understanding in VLMs, existing work~\cite{GeminiRobotics,zhang2025from,yeh2025seeing-All-Angles,yang2025mmsi,zheng2020university,fu2024blink} still fall short of systematically evaluating their correspondence capabilities. As shown in Tab.~\ref{tab:benchmark_comparison}, they exhibit several shortcomings: 1) \textit{\textbf{Limited and Coarse Task Granularity}}: The majority of benchmarks are restricted to a multiple-choice format (\textit{choice-level}), which fails to effectively evaluate a model's ability to predict continuous spatial coordinates, a \textit{point-level} localization skill is crucial for real-world applications. 2) \textit{\textbf{Lack of Semantic Relevance}}: Most tasks do not focus on functionally significant affordances, which are directly tied to physical interaction and thus create a disconnect between the evaluation and embodied applications. 3) \textit{\textbf{Insufficient Evaluation Dimensions}}: Previous works rarely provide a systematic assessment of model robustness against real-world challenges such as scale variations induced by viewpoint distance and object occlusion.

To address these gaps, we introduce \textbf{CrossPoint-Bench}, the first systematic benchmark designed to directly couple with these limitations. Inspired by the human cognitive process for cross-view reasoning, our benchmark is structured as a hierarchical progression of four evaluation dimensions: \textit{Fine-grained Grounding}, \textit{Visibility Reasoning}, \textit{Correspondence-Judgement}, and \textit{Correspondence-Pointing}. This hierarchical design systematically evaluates the entire reasoning chain from single-view localization to cross-view correspondence, directly addressing the \textit{task granularity} gap. Furthermore, to ensure \textit{semantic relevance}, all tasks are centered on functionally significant affordance regions, which we categorize into fine-grained \textit{semantic parts} and \textit{general objects}. The benchmark also systematically evaluates robustness against \textit{scale variations and occlusion}, tackling the final limitation. Our comprehensive evaluation on CrossPoint-Bench reveals a profound gap between current VLMs capabilities and human-level performance. For instance, the state-of-the-art (SOTA) Gemini-2.5-Pro achieves only 16.41\% accuracy on the critical point-level prediction task, in stark contrast to the human average of 93.63\%. Further analysis highlights two critical bottlenecks: 1) \textit{\textbf{Continuous Coordinate Degradation}}: We observe a substantial performance degradation when models transition from selecting a discrete region to predicting a continuous coordinate, exposing an inherent weakness in their geometric consistency modeling. 2) \textit{\textbf{Actionable Regions Gap}}: Models consistently underperform on fine-grained semantic parts compared to general objects, which indicates that their spatial understanding of functionally critical areas remains underdeveloped.

Motivated by these findings, which highlight a clear need for data that targets geometric consistency and fine-grained affordance, we construct the \textbf{CrossPoint-378K} dataset. It contains approximately 378k question-answering pairs, spanning 900 real-world indoor scenes, 162 interaction types, and 8k object instances. The dataset is uniquely organized around affordance regions to teach physically meaningful spatial representations. Building upon this, we develop \textbf{CroPond}, a strong baseline model whose performance demonstrates the effectiveness of our approach. Experimental results show that CroPond achieves 76.8\% accuracy on CrossPoint-Bench, substantially outperforming existing VLMs. 

The main contributions of this work are summarized as follows:
\begin{itemize}

    \item We systematically define the CVPC task and introduce CrossPoint-Bench, the first comprehensive benchmark for CVPC, which evaluates VLMs capabilities through a hierarchical task design and reveals their critical limitations in geometric consistency.

    \item We construct CrossPoint-378K, the first large-scale dataset for CVPC, which is focused on affordance regions, spanning 900 real-world indoor scenes, 162 interaction types, and 8k object instance.

    \item We develop CroPond, a strong baseline model that achieves SOTA performance on CrossPoint-Bench and serves as a reliable starting point for future research on cross-view correspondence.
    
\end{itemize}

\section{Related Work}
\label{sec:formatting}

\subsection{Spatial Understanding with VLMs}

Spatial understanding concerns the spatial attributes of objects, including position, orientation and scale, as well as relations among objects such as distance, direction and topological adjacency~\cite{song2025hazards,zhang-etal-2025-sphere,cheng2024spatialrgpt,yang2025thinking,ji2025visualtrans,lyu2025egoprompt,bai2025towards,ji2025mathsticks}. Early studies aligned semantics with vision at the region level in single-view images, typically predicting bounding boxes or masks~\cite{kang2025your,goto2025referring,yang2024boosting,song2025maniplvm}. More recent work progresses to point-level grounding, often incorporating depth, camera pose, and three-dimensional scene priors to bridge the gap between perception and executable pointing or manipulation~\cite{cheng2024spatialrgpt,park2025r,xu2025mc}. Representative efforts include RoboPoint~\cite{yuan2025robopoint}, which maps natural language to affordance points, and RoboRefer~\cite{zhou2025roborefer}, which achieves multi step spatial reasoning for referring. Despite progress across these directions, existing studies have yet to systematically address fine-grained point-level correspondence, leaving a critical gap between spatial understanding and executable embodied applications.

\subsection{Multi-view Understanding for VLMs}

Multi-view understanding aims to keep geometry and semantics consistent across viewpoints so answers and localizations remain stable under camera changes. It is critical for embodied manipulation~\cite{li2024manipllm,liang2024rapid,zhao2025cot,lee2025dynscene,wang2024scaling}, teleoperation~\cite{kuang2024towards,wu2024surgicai,chen2025robo2vlm}, augmented reality and virtual reality~\cite{luo2024real,lee2025rewind,hollidt2024egosim,song2025geolocation,cai2024benchlmm}, digital twin inspection~\cite{dong2025digital,li2025multi,gupta2025embodied}, and multi-agent cooperative perception~\cite{xu2024v2x,xia2025learning,liu2025mmcooper,zhong2025cooptrack,hong2024multi,tan2025roboosnext,tan2025roboos}. Prior work falls into three threads: 1) Multi-view geometry and consistency, which recovers cross-view structure or enforces appearance–geometry agreement during generation and editing~\cite{wang2025vggt,cabon2025must3r,wang2024dust3r,an2025cross,cao2024mvinpainter,zhang20244diffusion,asim2025met3r}. 2) 3D-aware representations and semantic grounding, which build persistent, queryable scene memory and align language to 3D anchors so identities persist across views~\cite{fan2024large,zhou2025cross,grauman2024ego}. 3) Multi-view cognition and action, which target view-stable reasoning, viewpoint management, and embodied decision-making, alongside evaluations of multi-view coherence~\cite{majumder2025viewpoint,zhang2025from,wang2025facebench,fu2024blink,yeh2025seeing-All-Angles,yang2025mmsi,zheng2020university}. Despite progress across these directions, existing studies have yet to systematically address multi-view consistency and establish correspondence within precise coordinate space. To bridge this gap, we formalize the Cross-View Point Correspondence (CVPC) task and introduce benchmark and dataset specifically tailored for this task.

\section{Cross-View Point Correspondence}

\subsection{Problem Definition}\label{sec:definition}

Inspired by the human cognitive process of cross-view reasoning, which involves perceiving a point from instruction, reasoning its visibility from another view, and corresponding it across views, we define Cross-View Point Correspondence (CVPC) as a multi-stage cognitive task.

\begin{itemize}
    \item \textbf{Instruction-conditioned Grounding.} 
    Given an image view \( I_a \) and a natural language instruction \( T \), the model first interprets the instruction semantics and localizes the corresponding target position in the image:
    \[
    f_{\text{loc}} : (I_a, T) \rightarrow p_a,
    \]
    where \( p_a\) denotes the target 2D point in \( I_a \).
 This stage serves as the foundation of CVPC.
 
    \item \textbf{Visibility Reasoning.} 
    For another view \( I_b \), the model determines whether the target point is visible in this view:
    \[
    f_{\text{vis}} : (I_a, I_b, p_a) \rightarrow y_{\text{vis}},
    \]
    where \( y_{\text{vis}} \in \{0, 1\} \) indicates the visibility of the target in \( I_b \). 
    This step requires the model to reason about occlusion and depth relations.

    \item \textbf{Correspondence.} 
    If the target is visible in \( I_b \), the model further predicts the corresponding 2D point:
    \[
    f_{\text{map}} : (I_a, I_b, p_a) \rightarrow p_b,
    \]
    where \( p_b \) represents the target 2D point in view \( I_b \).
    In 3D space, the two 2D points \( p_a \) and \( p_b \) are projections of the same 3D point \( P \) onto the imaging planes of two cameras. 

    \end{itemize}
For example, when instructed to pick up the left cup, the model first localizes the cup’s handle in the first view, then reasons about its visibility in the second view, and, if not occluded, maps the localized point to the corresponding position in the second view.

\subsection{CrossPoint-Bench}

\begin{figure}[t]  
  \centering
  \includegraphics[width=\linewidth]{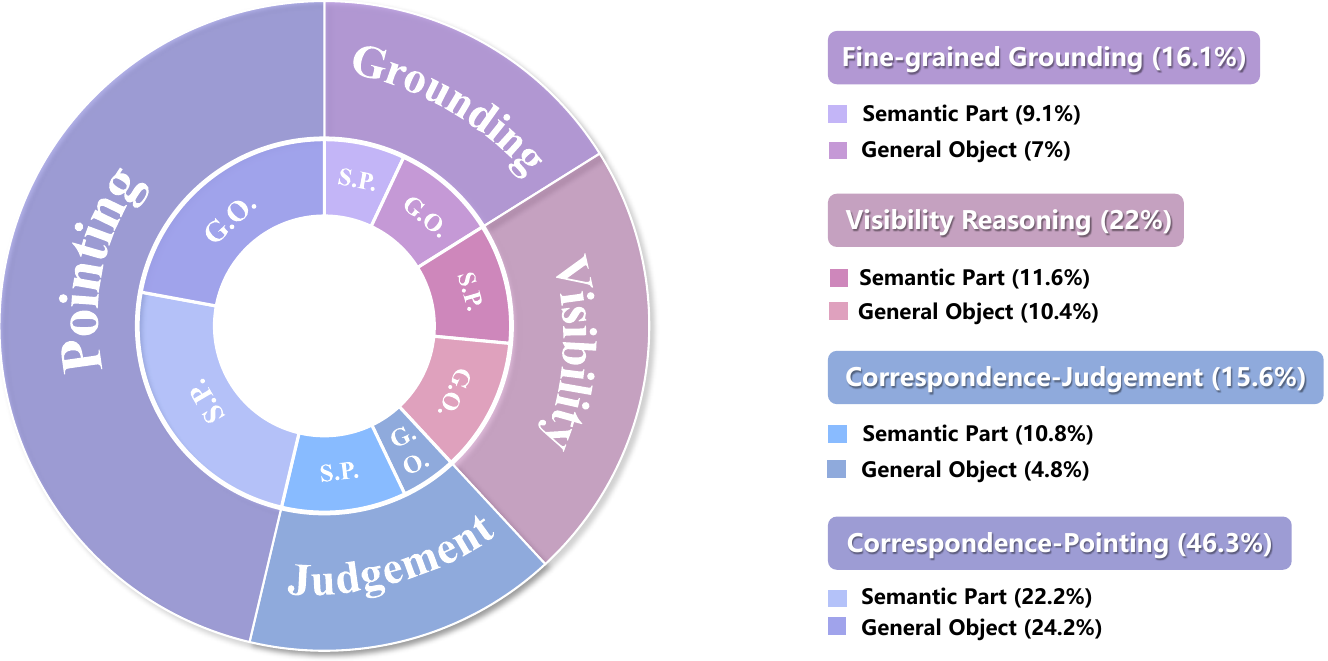} 
  \caption{\textbf{Overview of CrossPoint-Bench,} which is divided into four categories, each covering two levels of affordance.}
  \label{fig:CrossPoint-Bench}
\end{figure}

\begin{figure*}[htbp]
    \centering
    \includegraphics[width=\textwidth]{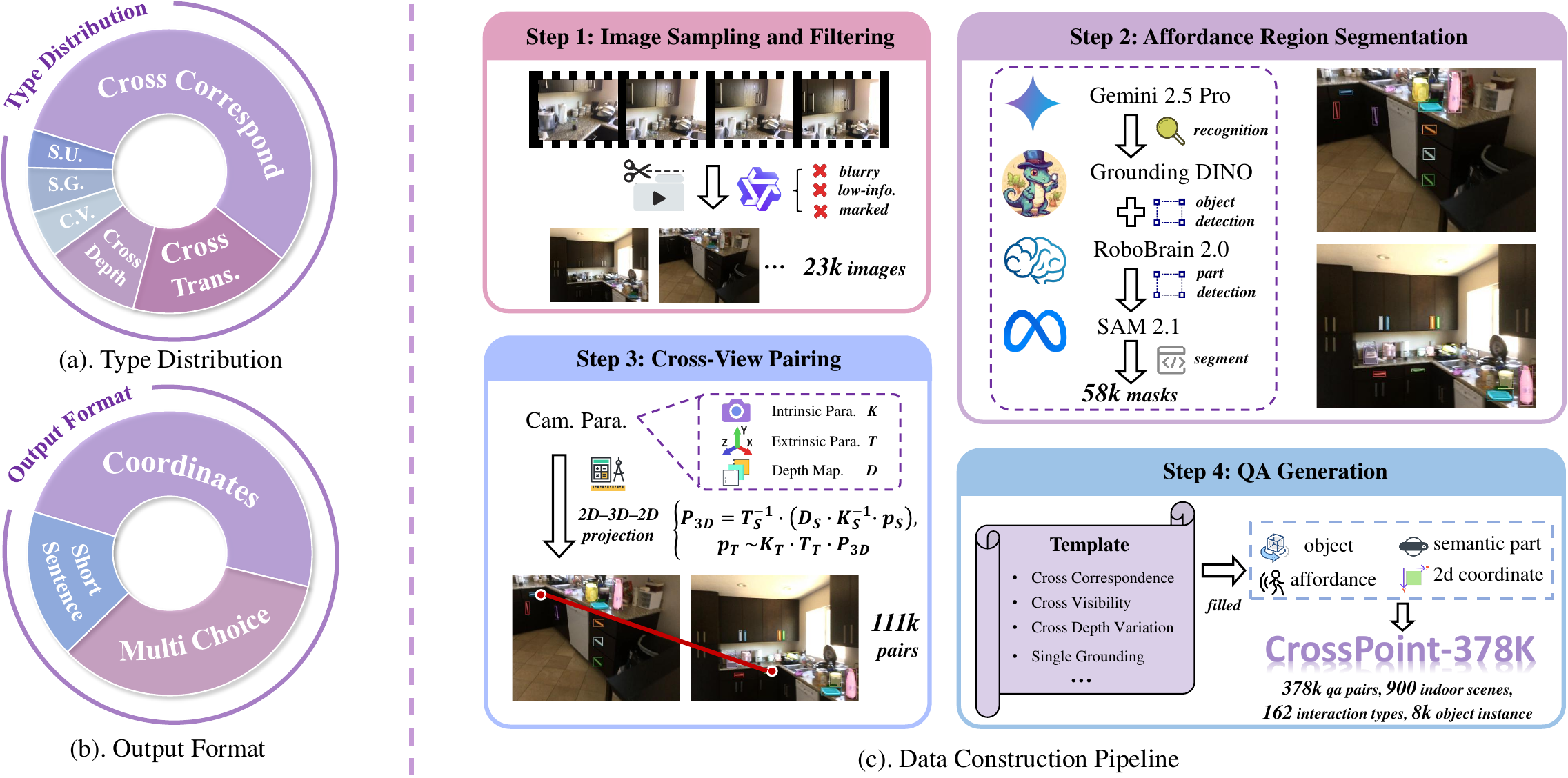}
    \caption{\textbf{Overview of the CrossPoint-378K.} CrossPoint-378K is a comprehensive dataset for Cross-View Point Correspondence, synthesized via a specially designed automated generation pipeline. S.U., S.G., and C.V. in (a) denote Single-View Spatial Understanding, Single-View Fine-Grained Grounding, and Cross-View Visibility Reasoning, respectively.}
    \label{fig:CrossPointOverview}
\end{figure*}

To evaluate CVPC in VLMs, we propose \textbf{CrossPoint-Bench}, the first benchmark specifically designed for this capability. This benchmark comprises \textbf{1k} QA pairs collected from real-world indoor scenes and focus on operable affordance regions. In practical scenarios of CVPC, the target points correspond to locations that require interaction, and the majority of interactions involve semantic parts, which provide more precise action cues than general objects. Accordingly, we categorize operable regions into two granularities: general objects (e.g., light switches) and semantic parts (e.g., teacup handles).

Following the task definition in Sec.~\ref{sec:definition}, our benchmark defines four task types. Within the \textit{Correspondence} category, we introduce an additional Judgement task alongside the original Pointing task. This transitional task enables evaluation of basic semantic correspondence before precise coordinate prediction, providing deeper insights into the model’s performance. The following sections summarize each task, and dataset statistics are presented in Fig.~\ref{fig:CrossPoint-Bench}.
\begin{itemize}
    \item \textbf{Fine-grained Grounding}: Input an image and an instruction; output the coordinates of the referred target.

    \item \textbf{Visibility Reasoning}: Input an image and a point; output whether the point is visible in another view.

    \item \textbf{Correspondence-Judgement}: Input an image and a point; select the correct correspondence from multiple candidates in another view.

    \item \textbf{Correspondence-Pointing}: Input an image and a point; predict the exact coordinates of the corresponding point in another view.
\end{itemize}

\subsection{CrossPoint-378K}

To enhance models’ capabilities in CVPC, we introduce \textbf{CrossPoint-378K}, a dataset focused on spatial pointing ability, as shown in Fig. ~\ref{fig:CrossPointOverview}. It emphasizes point-level spatial understanding and is constructed around the semantic, geometric, and cross-view relationships of spatial points. The goal is to advance model development toward unified fine-grained spatial perception and reasoning. The key characteristics of the dataset are as follows:

\begin{itemize}
    \item \textbf{Affordance-Driven:} The dataset prioritizes affordable regions (e.g., the handle of a teacup), as opposed to static or non-interactive regions. We emphasize the relevance to real-world interactive scenarios, enabling models to learn more physically meaningful spatial representations and gain the ability to understand and point to actionable targets in real environments.
    
    \item \textbf{Multidimensional:} The dataset covers both single-view and cross-view dimensions. Single-view capabilities form the foundation of spatial understanding, while cross-view capabilities represent its core and extension. In the single-view context, it encompasses fine-grained grounding and spatial understanding, guiding models to master point-level location and spatial structure from a single viewpoint. In the cross-view context, it further introduces visibility reasoning, depth variation, transformation understanding, and correspondence reasoning.
    
    \item \textbf{Rich Diversity:} The dataset contains approximately 378k visual question-answer pairs, spanning 900 indoor scenes, and covering 162 types of interaction actions and 8k object instances. The diverse scene distribution and rich forms of interaction ensure the dataset systematically enhances models' spatial pointing ability while also emphasizing affordance regions.
    
    \item \textbf{Scalability and Generative Ability:} The dataset exhibits strong scalability, enabling efficient generation of more samples through our designed automated data synthesis pipeline, which integrates 3D scene resources and multi-view video data.
\end{itemize}

We propose a highly automated pipeline to build the data. The entire process involves image sampling and filtering, affordance region segmentation, cross-view pairing, and QA generation. Details are provided in Appx.~\ref{sec:pipeline}.

\begin{itemize}
    \item \textbf{Image Sampling and Filtering:} We first downsample the high-frame-rate 3D videos~\cite{dai2017scannet,yeshwanth2023scannet++} with an average sampling ratio of 1:50 to reduce frame redundancy. Then, we apply Qwen2.5-VL-7B~\cite{Qwen2.5-VL} for automatic quality screening to remove blurry, low-information-density, or marked images and obtain 23k images.
    
    \item \textbf{Affordance Region Segmentation:} We employ a multi-stage fine-grained mask generation pipeline. Gemini-2.5-Pro~\cite{gemini25pro} is used to detect operable regions, Grounding-DINO~\cite{liu2024grounding} to obtain bounding boxes of fine-grained objects, and RoboBrain2.0~\cite{team2025robobrain} to extract bounding boxes of semantic parts. Finally, SAM2.1~\cite{ravi2025sam} segments these regions, yielding 58k fine-grained mask annotations.
    
    \item \textbf{Cross-View Pairing:} For each instance mask in an image, we randomly select a point and use the scene’s camera parameters to perform spatial mapping across different viewpoints. By conducting one-to-one cross-view point mapping for all images within each scene, we generate 111k cross-view correspondence pairs.
    
    \item \textbf{QA Generation:} We design multiple sets of question–answer templates, some containing placeholders. For single-image and cross-view tasks, these placeholders are filled with corresponding object categories, actions, or coordinate information, producing 378k high-quality QA samples consistent with task semantics. 
\end{itemize}

\subsection{Training Details}

To further enhance the model’s capability in establishing accurate geometric alignment and fine-grained correspondence across views, we develop a strong baseline model, \textbf{CroPond}, for evaluating the CVPC task.

We utilize Qwen2.5-VL (3B/7B)~\cite{Qwen2.5-VL} as the base model. This model consists of an image encoder, an MLP projector, and a language model. We apply supervised fine-tuning to train CroPond. During training, we combine the CrossPoint-378K with single-view spatial understanding data~\cite{zhou2025roborefer,SAT}, multi-view spatial understanding data~\cite{zhang2025from,zhang2025mllms} and instruction-tuned data~\cite{liu2024improved}. This multi-source joint training strategy not only significantly enhances the model's point-level correspondence abilities in cross-view scenarios but also ensures that the model retains the general multimodal knowledge acquired during pre-training.

\section{Experiment}
\begin{table*}[htbp]
\centering
\caption{\textbf{Performance of various models on CrossPoint-Bench.} Top-1 \& Top-2 accuracies are represented using \textbf{bold text}, and \underline{underlined}. As a comprehensive comparison, we report results for human evaluation, proprietary models, open-source VLMs, spatial / referring specialist models, and our proposed CroPond variants.}
\setlength{\tabcolsep}{5pt}
\renewcommand{\arraystretch}{1.1}
\scriptsize
\begin{tabularx}{\textwidth}{l|Y|Y Y Y|Y Y Y|Y Y Y|Y Y Y}
\toprule
\multirow{2}{*}{\textbf{Model}} & \multirow{2}{*}{\textbf{Score}} &
\multicolumn{3}{c|}{\textbf{Fine-grained Ground.}} &
\multicolumn{3}{c|}{\textbf{Visibility Reason.}} &
\multicolumn{3}{c|}{\textbf{Correspondence-Judge.}} &
\multicolumn{3}{c}{\textbf{Correspondence-Point.}} \\
 &  &
\textbf{object} & \textbf{part} & \textbf{sum} &
\textbf{object} & \textbf{part} & \textbf{sum} &
\textbf{object} & \textbf{part} & \textbf{sum} &
\textbf{object} & \textbf{part} & \textbf{sum} \\
\midrule
\rowcolor[HTML]{F2F2F2} \multicolumn{14}{l}{\textbf{Human Evaluation}}\\
Human & 91.75 & 83.57 & 76.37 & 79.51 & 96.15 & 89.66 & 92.73 & 98.49 & 96.63 & 97.44 & 95.87 & 91.18 & 93.63 \\
\midrule
\rowcolor[HTML]{F2F2F2} \multicolumn{14}{l}{\textbf{Proprietary Models}}\\
Claude-Sonnet-4 & 23.10 & 17.14 & 7.69 & 11.80 & 66.35 & 50.86 & 58.18 & 41.67 & 37.04 & 38.46 & 7.02 & 3.17 & 5.18 \\
Gemini-2.5-Pro & 37.10 & 50.00 & 19.78 & 32.92 & 68.27 & 66.38 & 67.27 & 75.00 & 53.70 & 60.26 & 20.25 & 12.22 & 16.41 \\
o3 & 32.60 & 11.43 & 6.59 & 8.70 & \underline{79.81} & \underline{71.55} & \underline{75.45} & 70.83 & 60.19 & 63.46 & 10.74 & 9.50 & 10.15 \\
\midrule
\rowcolor[HTML]{F2F2F2} \multicolumn{14}{l}{\textbf{Open-Source Vision-Language Models}}\\
Qwen2.5-VL-3B-Instruct & 25.40 & 67.14 & 41.76 & 52.80 & 50.96 & 43.97 & 47.27 & 33.33 & 33.71 & 33.97 & 2.89 & 2.26 & 2.59 \\
Qwen2.5-VL-7B-Instruct & 26.80 & 65.71 & 35.16 & 48.45 & 50.00 & 63.79 & 57.27 & 54.55 & 37.08 & 41.03 & 0.00 & 0.00 & 0.00 \\
Qwen2.5-VL-32B-Instruct & 29.10 & 32.86 & 14.29 & 22.36 & 56.73 & 58.62 & 57.73 & 47.92 & 40.74 & 42.95 & 19.83 & 5.88 & 13.17 \\
Qwen2.5-VL-72B-Instruct & 36.50 & 65.71 & 46.15 & 54.66 & 49.04 & 58.62 & 54.09 & 45.83 & 51.85 & 50.00 & 23.14 & 10.86 & 17.28 \\
Qwen3-VL-30B-A3B-Instruct & 28.10 & 55.71 & 26.37 & 39.13 & 66.35 & 52.59 & 59.09 & 50.00 & 41.67 & 44.23 & 4.55 & 3.62 & 4.10 \\
Qwen3-VL-30B-A3B-Thinking & 39.70 & 72.86 & 38.46 & 53.42 & 49.04 & 59.48 & 54.55 & 45.83 & 45.37 & 45.51 & 32.23 & 19.00 & 25.92 \\
Qwen3-VL-235B-A22B-Instruct & 35.20 & 70.00 & \underline{52.75} & \underline{60.25} & 59.62 & 56.90 & 58.18 & 79.17 & 54.63 & 62.18 & 7.85 & 4.98 & 6.48 \\
Qwen3-VL-235B-A22B-Thinking & 52.70 & \underline{74.29} & \textbf{58.24} & \textbf{65.22} & 65.38 & 59.48 & 62.27 & 77.08 & 61.11 & 66.03 & 50.83 & 26.70 & 39.31 \\
\midrule
\rowcolor[HTML]{F2F2F2} \multicolumn{14}{l}{\textbf{Spatial / Referring Specialist Models}}\\
RoboPoint-v1-Vicuna-13B & 14.60 & 7.14 & 7.69 & 7.45 & 33.65 & 41.38 & 37.73 & 37.50 & 25.93 & 29.49 & 0.83 & 1.36 & 1.08 \\
RoboRefer-8B-SFT & 21.00 & 62.86 & 32.97 & 45.96 & 37.50 & 37.07 & 37.27 & 22.92 & 31.48 & 28.85 & 1.65 & 2.26 & 1.94 \\
RoboBrain2.0-7B(no-thinking) & 25.40 & 64.29 & 41.76 & 51.55 & 46.15 & 55.17 & 50.91 & 39.58 & 33.33 & 35.26 & 0.41 & 1.36 & 0.86 \\
RoboBrain2.0-7B(thinking) & 26.00 & 72.86 & 40.66 & 54.66 & 46.15 & 52.59 & 49.55 & 29.17 & 41.67 & 37.82 & 0.41 & 1.36 & 0.86 \\
Molmo-7B-D & 25.80 & 62.86 & 36.26 & 47.83 & 49.04 & 51.72 & 50.45 & 37.50 & 35.19 & 35.90 & 4.13 & 1.81 & 3.02 \\
\midrule
\rowcolor[HTML]{F2F2F2} \multicolumn{14}{l}{\textbf{CroPond Variants}}\\
CroPond-3B & \underline{71.60} & \textbf{78.57} & 43.96 & 59.01 & \underline{79.81} & 65.52 & 72.27 & \underline{90.91} & \underline{73.03} & \underline{76.92} & \textbf{84.30} & \underline{62.44} & \underline{73.87} \\
CroPond-7B & \textbf{76.80} & 71.43 & 51.65 & \underline{60.25} & \textbf{81.73} & \textbf{76.72} & \textbf{79.09} & \textbf{93.94} & \textbf{87.64} & \textbf{87.18} & \underline{83.88} & \textbf{71.49} & \textbf{77.97} \\
\bottomrule
\end{tabularx}
\label{tab:crosspoint-bench}
\end{table*}

\textbf{Evaluation metrics.} On CrossPoint-Bench, we adopt two complementary metrics: multiple-choice tasks use average accuracy, while pointing tasks use in-mask hit rate, which measures whether the predicted 2D coordinate falls within the ground-truth mask.

\textbf{Compared baselines.} We evaluate three families of methods: proprietary VLMs, including Claude-Sonnet-4~\cite{Claude-4}, Gemini-2.5-Pro~\cite{gemini25pro}, and o3~\cite{gpto3-o4-mini}; open-source VLMs represented by the Qwen-VL series including Qwen2.5-VL~\cite{Qwen2.5-VL} and Qwen3-VL~\cite{Qwen3-VL}; and spatial or referring specialist models, including RoboPoint~\cite{yuan2025robopoint}, RoboRefer~\cite{zhou2025roborefer}, RoboBrain-2.0~\cite{team2025robobrain}, and Molmo~\cite{deitke2025molmo}.

\begin{figure}[t]  
  \centering
  \includegraphics[width=\linewidth]{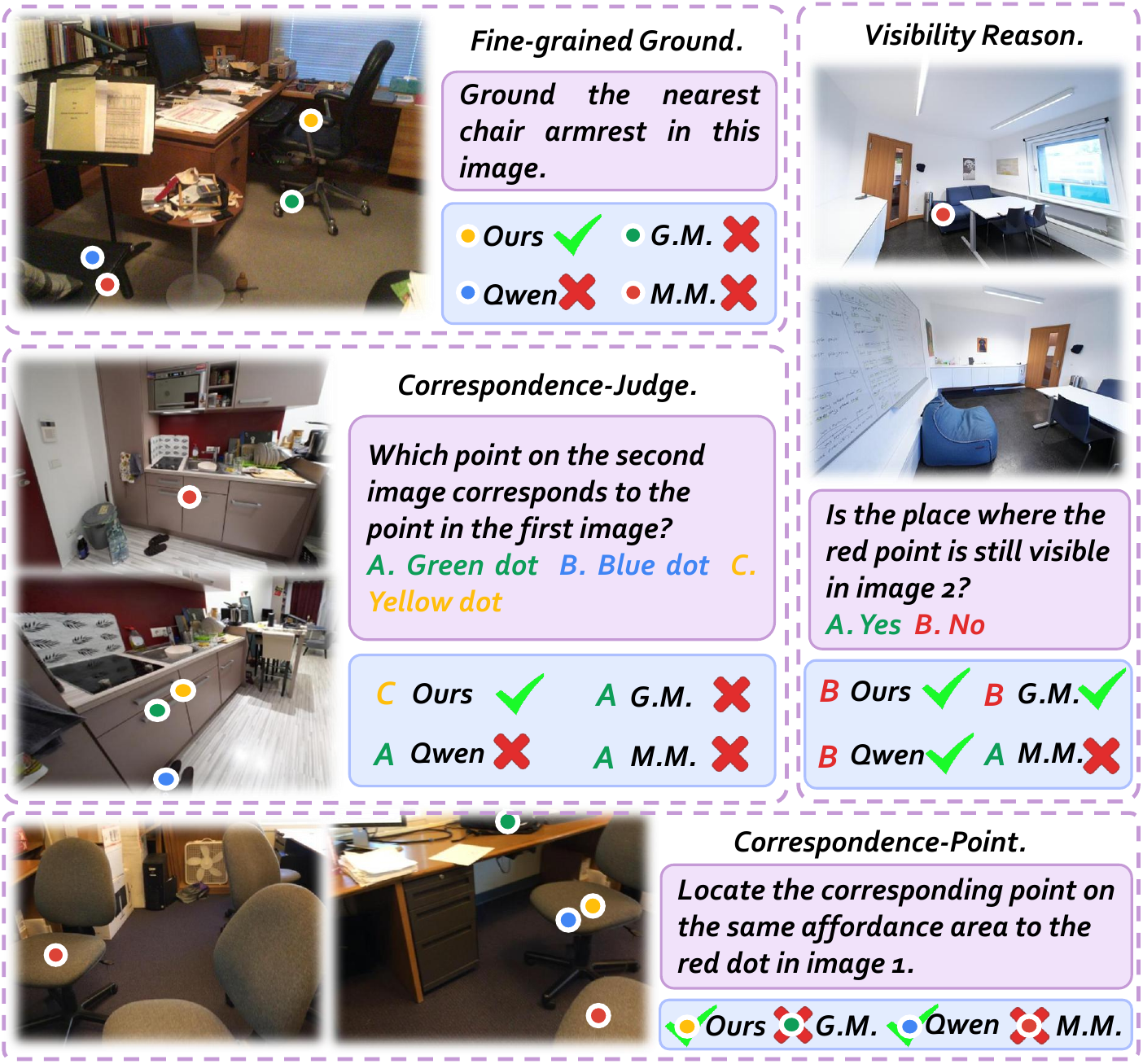} 
  \caption{\textbf{CrossPoint-Bench results.} Qwen, G.M., and M.M. donate Qwen3-VL-235B-A22B-Insturct~\cite{Qwen3-VL}, Gemini-2.5-Pro~\cite{gemini25pro}, and Molmo-7B-D~\cite{deitke2025molmo}. CroPond-7B excels across the four cases.}
  \label{fig:case}
\end{figure}

\subsection{Results on CrossPoint-Bench}
To assess the capabilities of general and spatially specialized VLMs in CVPC, we conduct a systematic evaluation on CrossPoint-Bench, with results summarized in Tab.~\ref{tab:crosspoint-bench}, and representative case types illustrated in Fig.~\ref{fig:case}. We report the following findings:

    \textit{\textbf{1) General VLMs remain substantially weaker than humans.}} Humans achieve an overall accuracy of 91.75\%, while the best-performing model, Qwen3-VL-235B-A22B-Thinking, reaches only 52.70\%. On the critical Correspondence-Pointing task, general VLMs average just 12.78\%, highlighting a substantial gap in their ability to perform precise spatial reasoning.

    \textit{\textbf{2) Performance degrades sharply from discrete to continuous tasks.}} General VLMs demonstrate an average accuracy of 49.8\% on Correspondence-Judgement tasks, yet this figure drops precipitously to 12.78\% on Correspondence-Pointing tasks that require continuous coordinate generation. This trend is consistent across models; Gemini-2.5-Pro, for example, moves from 60.26\% on Correspondence-Judgement to 16.41\% on Correspondence-Pointing. The average reduction of 37.02\% highlights the core gap from judgment to precise landing.

    \textit{\textbf{3) Spatially specialized models do not yet solve cross-view correspondence.}} These specialized models achieve only 22.56\% overall on average, not markedly better than general VLMs. Training that concentrates on single view or region level supervision transfers poorly to cross pose settings with free coordinate outputs.

    \textit{\textbf{4) Cross-view correspondence is sensitive to granularity.}} General VLMs score 46.48\% on general objects but only 36.47\% on semantic parts on average, showing a 10.01\% advantage for objects. Small parts like handles or buttons remain difficult, revealing weaknesses in capturing fine-grained boundaries and topology across viewpoints.

\begin{figure}[t]  
  \centering
  \includegraphics[width=\linewidth]{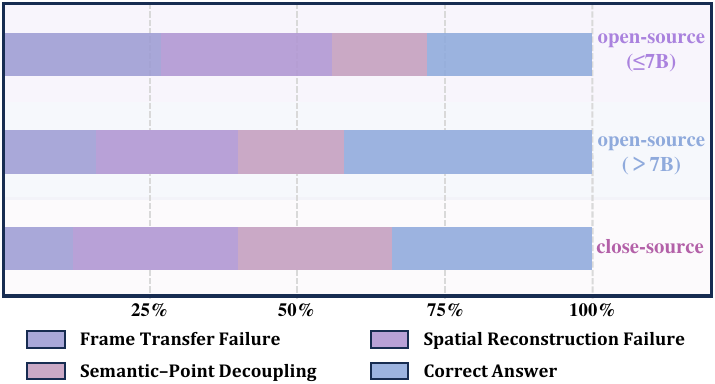} 
  \caption{\textbf{Error distribution for different models.} Spatial reconstruction failure dominates across all models, while frame transfer errors are mainly in open-source models.}
  \label{fig:error}
\end{figure}

\subsection{Error Analysis}
    To gain deeper insight into the challenges encountered by models in cross-view correspondence, we analyze the key errors observed during evaluation. These errors highlight several limitations of current VLMs, as illustrated in Fig.~\ref{fig:error}.
    
    Our analysis reveals three primary error patterns in cross-view correspondence: \textit{\textbf{1) Frame Transfer Failure.}} The model fails to transfer and align coordinate systems during cross-view reasoning. It continues to reason within the reference frame of the source view, resulting in incorrect mapping to the target-view coordinate space; \textit{\textbf{2) Spatial Reconstruction Failure.}} After observing multiple views, the model fails to internally reconstruct a consistent 3D spatial representation. It lacks the ability to integrate occlusion reasoning and relative spatial layout, leading to inconsistent or fragmented spatial understanding; \textit{\textbf{3) Semantic–Point Decoupling.}} A misalignment emerges between semantic representation and pixel-level grounding. The model can correctly infer the semantic concept of the target but struggles to precisely align it with the corresponding pixel location, or fails to interpret the fine-grained semantics implied by the specific point.

    To further investigate the root causes of these errors, we conducted exploratory experiments targeting these error types. These experiments aimed to determine whether the observed failures stem from deficiencies in the model's inherent reasoning capabilities. For instance, we explicitly supplied the semantic target and examined whether the model could subsequently point to the correct location, thereby testing whether such semantic guidance helps it rectify prior failures. The results offer valuable insights on enhancing VLMs for cross-view tasks. Detailed experiment are provided in Appx.~\ref{sec:exploratory}.

\subsection{Results on CroPond}

We evaluate CroPond models (3B and 7B) on CrossPoint-Bench and report three key findings.

    \textit{\textbf{1) Sufficient data and supervised fine-tuning substantially outperform strong language model baselines.}} CroPond-3B attains an overall accuracy of 71.6\%, exceeding the baselines Qwen3-VL-235B-A22B-Thinking and Gemini-2.5-Pro, and it is notably stronger on the most challenging Correspondence Pointing task. These results underscore the central importance of data scale and task alignment for point-level cross-view reasoning.
    
    \textit{\textbf{2) Clear scaling behavior emerges, with larger models yielding greater gains from supervised fine-tuning.}} CroPond-7B achieves an overall accuracy of 76.8\%, improving upon CroPond-3B across all evaluation subsets. The largest gains occur in Visibility Reasoning and Correspondence-Judgement, with improvements of 6.82\% and 10.26\%, respectively. These results suggest that larger model scale enables more robust spatial reasoning and more accurate mapping from semantics to coordinates.
    
    \textit{\textbf{3) Dual leadership in cross-view consistency and coordinate grounding.}} CroPond-7B substantially outperforms Qwen3-VL-235B-A22B-Thinking on Visibility Reasoning, Correspondence-Judgement, and Correspondence-Pointing, and it attains approximately 83\% of human performance on free coordinate Correspondence Pointing. Taken together, these findings demonstrate that supervised fine tuning on sufficient, well aligned data produces robust geometric and semantic alignment for cross-view point tasks.

\begin{figure}[t]  
  \centering
  \includegraphics[width=\linewidth]{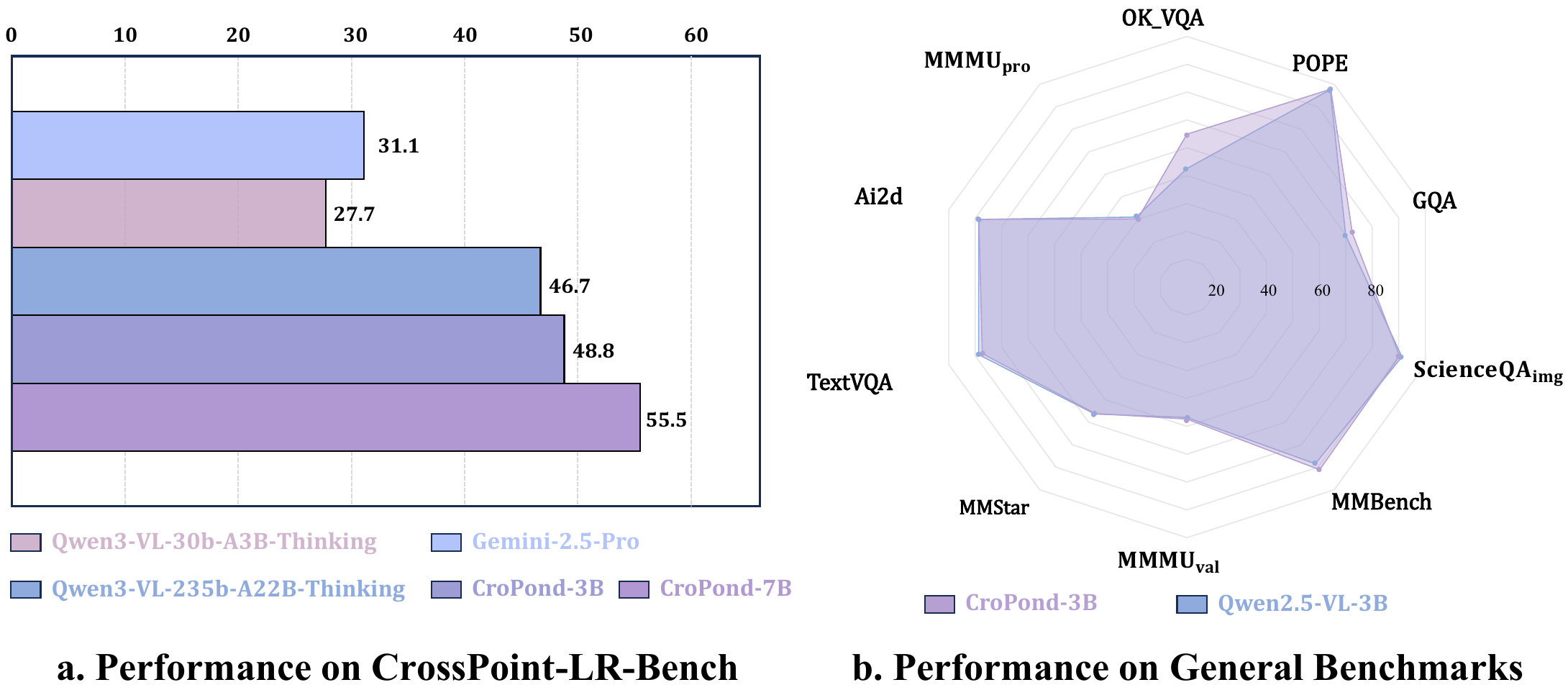} 
  \caption{\textbf{Performance on CrossPoint-LR-Bench and general benchmarks.} CroPond significantly outperforms baselines on long-range reasoning tasks and maintains comparable performance on general benchmarks.}
  \label{fig:LongAndGeneral}
\end{figure}

\subsection{Results on CrossPoint-LR-Bench}

    To evaluate more challenging long-range cross-view tasks, we construct \textbf{CrossPoint-LR-Bench}, a small-scale benchmark designed to assess continuous long-horizon reasoning capabilities. CrossPoint-LR-Bench comprises 90 samples featuring more complex spatial variations and out-of-domain scenes that differ significantly from the training distribution.  Each sample in CrossPoint-LR-Bench consists of a three-stage (refer to Sec.~\ref{sec:definition}), multi-turn dialogue task. Crucially, a trial is deemed successful only if all stages are completed correctly. Detailed procedures and evaluation metrics are provided in Appx.~\ref{sec:LR-Bench}. As shown in Fig.~\ref{fig:LongAndGeneral} (a), we find that \textit{\textbf{CroPond demonstrates robust extrapolation in long-range cross-view reasoning}}. Despite the significant performance drop in multi-stage tasks due to error propagation and added complexity, CroPond outperforms other models, with CroPond-7B achieving 55.5\% accuracy and showing strong generalization in these challenging tasks.

\begin{table*}[htbp]
\centering
\caption{\textbf{Performance on spatial understanding benchmarks.} Rel. and Dist. denote relation and distance; M.V. and V.C. denote Multi-view Reasoning and Visual Correspondence; P.M. and C.M. denote Position Matching and Camera Motion Infer. Results on SPAR-Bench are taken from the original paper~\cite{zhang2025from}. Top-1 \& Top-2 accuracies are represented using \textbf{bold text}, and \underline{underlined}.}
\setlength{\tabcolsep}{5pt}
\renewcommand{\arraystretch}{1.1}
\scriptsize
\begin{tabularx}{\textwidth}{l|YYY|YYYY|YY|Y}
\toprule
\multirow{2}{*}{\textbf{Model}} &
\multicolumn{3}{c|}{\textbf{CV-Bench~\cite{CV-Bench}}} &
\multicolumn{4}{c|}{\textbf{BLINK\textsubscript{val}~\cite{fu2024blink}}} &
\multicolumn{2}{c|}{\textbf{SPAR-Bench~\cite{zhang2025from}}} &
\multirow{2}{*}{\textbf{SAT~\cite{SAT}}}\\ 
& \textbf{Rel.} & \textbf{Dist.} & \textbf{Depth}
& \textbf{Depth} & \textbf{Rel.} & \textbf{M.V.} & \textbf{V.C.}
& \textbf{P.M.} & \textbf{C.M.} & \\

\midrule
\rowcolor[HTML]{F2F2F2}
\multicolumn{11}{l}{\textbf{Proprietary Models}}\\
GPT-4o                          & 84.62 & 83.33 & 86.50 & 74.19 & 80.42 & 50.38 & 89.53 & 30.00 & 16.00 & 53.67 \\
Gemini-2.5-Pro                  & \underline{93.54} & \textbf{90.67} & 91.00 & \underline{81.45} & \textbf{90.21} & 38.35 & 91.86 &  -  &  -  & \underline{79.00}  \\
Claude-3.7-Sonnet               & 74.15 & 84.17 & 85.83 & 70.16 & 79.02 & 46.62 & 90.70 & 16.00 &  6.00 & 60.00  \\
\midrule
\rowcolor[HTML]{F2F2F2}
\multicolumn{11}{l}{\textbf{Open-Source Vision-Language Models}}\\
Qwen2.5-VL-3B-Instruct          & 71.23 & 52.50 & 77.17 & 62.90 & 79.72 & 44.36 & 37.79 & 55.47 & 41.50 & 56.67  \\
Qwen2.5-VL-7B-Instruct          & 87.85 & 78.17 & 74.33 & 70.16 & \underline{88.11} & 43.61 & 72.09 & 53.94 & 38.50 & 56.00  \\
Qwen2.5-VL-32B-Instruct         & 89.08 & 79.83 & 87.17 & 71.77 & 74.13 & 43.61 & 73.26 &  -  &  -  & 65.00  \\
Qwen2.5-VL-72B-Instruct         & 92.62 & 85.00 & 89.33 & 77.42 & 86.71 & 47.37 & 83.72 & 40.00 & 16.00 & 69.00  \\
Qwen3-VL-30B-A3B-Instruct       & 92.59 & 80.00 & 90.83 & 73.39 & 83.92 & 40.60 & 77.33 &  -  &  -   & 69.39  \\
Qwen3-VL-235B-A22B-Instruct     & 93.00 & \underline{90.48} & \underline{92.83} & 78.69 & 85.19 & 40.15 & \underline{94.08} &  -  &  -  & \textbf{80.00}  \\
\midrule
\rowcolor[HTML]{F2F2F2}
\multicolumn{11}{l}{\textbf{CroPond Variants}}\\
CroPond-3B                      & 93.08 & 86.50 & 92.33 & \textbf{83.06} & 84.02 & \underline{88.72} & 81.98 & \underline{75.57} & \underline{71.00}  & 78.33  \\
CroPond-7B                      & \textbf{94.00} & 88.33 & \textbf{93.83} & \textbf{83.06} & 74.83 & \textbf{98.50} & \textbf{97.67} & \textbf{79.39} & \textbf{80.25} & 73.00  \\
\bottomrule
\end{tabularx}
\label{tab:spatialbench}
\end{table*}

\subsection{Results on Spatial and General Benchmarks}
To evaluate CroPond’s transferability and robustness in spatial understanding, including cross-view reasoning, we compare it with strong baselines on spatial benchmarks. We further analyze its general capability by comparing CroPond-3B with its base model Qwen2.5-VL-3B on general vision-language benchmarks. Check Appx.~\ref{sec:Spatial_Bench} and Appx.~\ref{sec:General_Bench} for more evaluation details. We find the following conclusions: \textit{\textbf{1) Spatial understanding, particularly in cross-view related tasks, surpasses strong baselines.}} In Tab.~\ref{tab:spatialbench}, CroPond attains leading performance across these tasks, indicating strong generalization and robust spatial reasoning. In particular, CroPond-7B improves over Gemini-2.5-Pro by approximately 4.27\% on the averaged score. \textit{\textbf{2) CroPond achieves robust transfer gains in general spatial understanding.}} In Fig.~\ref{fig:LongAndGeneral} (b), CroPond-3B performs comparably to or slightly better than other models on a range of general tasks. These findings suggest that incorporating spatially-aware supervision, through the use of CrossPoint-378K and other multi-source visual instruction datasets, can significantly improve the model's spatial reasoning and cross-view correspondence ability without sacrificing its overall visual-language proficiency.    

\section{Real-world Applications}
\label{sec:applications}

\begin{figure}[t]  
  \centering
  \includegraphics[width=\linewidth]{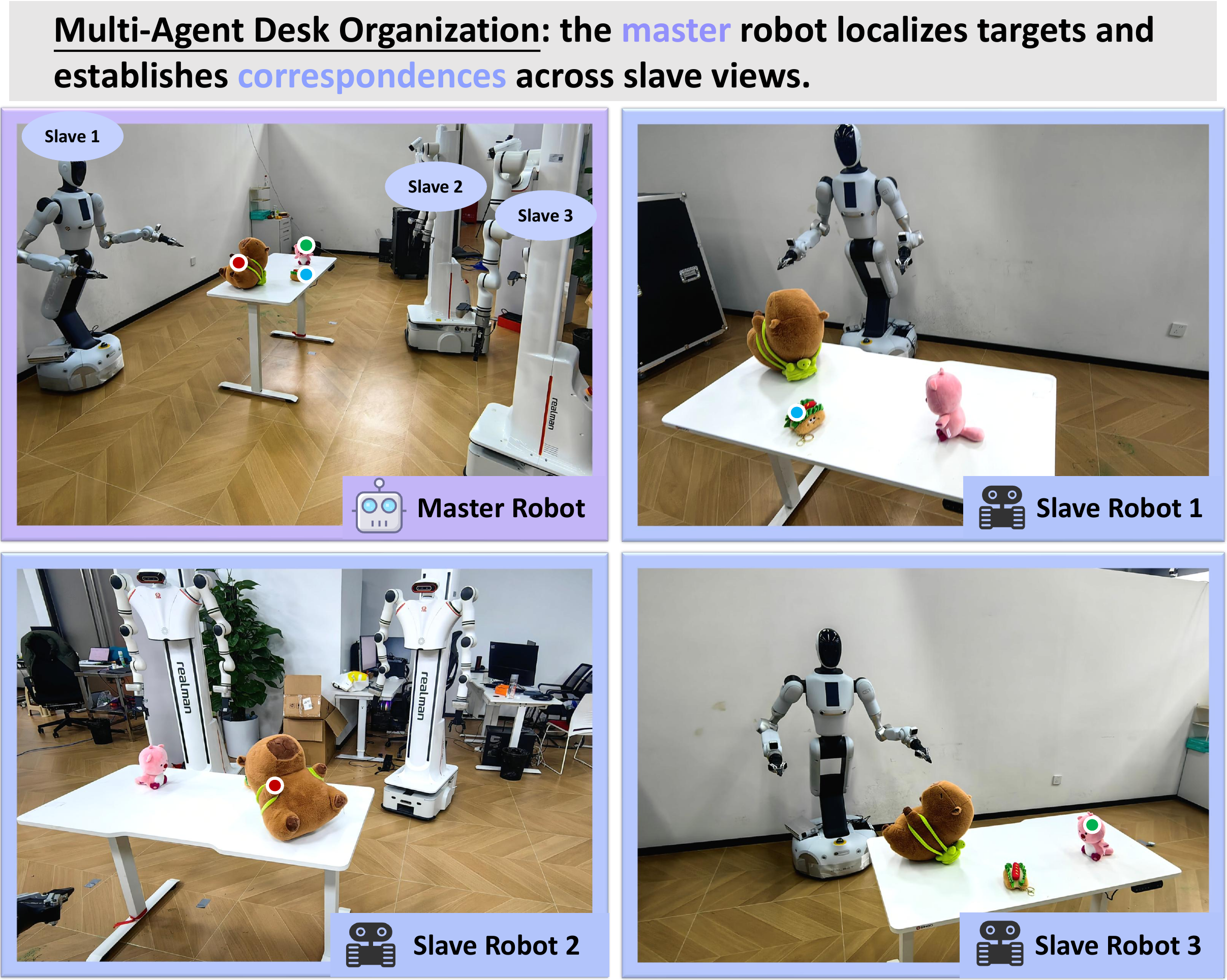} 
  \caption{\textbf{Multi-Agent Collaboration.} The points in the slave robots' views denote the target predicted by our model.}
  \label{fig:applications}
\end{figure}

To demonstrate CroPond’s practical performance, we evaluate it in a multi-agent desk organization scenario, where the task involves organizing three toys on a desk. A master agent with a global view, controlled by CroPond, coordinates three slave agents positioned around the desk. Upon receiving a user command, the master agent identifies the target toys and establishes their correspondences across the slave agents' views. Each slave agent retrieves the nearest toy based on its field of view, with the master sending point-based coordinates for guidance. This approach improves task efficiency by minimizing redundant movements, as each robot focuses on its closest toy. CroPond’s coordination across agents ensures that toys are correctly identified from different perspectives, contributing to more reliable multi-agent task execution. As shown in Fig.~\ref{fig:applications}, this setup demonstrates CroPond’s potential in real-world multi-agent scenarios, enhancing coordination in task completion. This capability further enables downstream applications in multi-robot navigation and coordinated manipulation, where consistent cross-view understanding is essential. By providing reliable correspondence cues, CroPond supports more precise motion planning and more stable object-handling behaviors in multi-agent systems.

\section{Conclusion}
\label{sec:conclusion}

We systematically define the Cross-View Point Correspondence (CVPC) task, a capability critical for VLMs and embodied AI. To study this problem, we propose CrossPoint-Bench, the first comprehensive hierarchical benchmark for CVPC. Our evaluation reveals critical limitations in current models’ geometric consistency, particularly in transitioning from coarse region selection to precise coordinate prediction. To address this, we construct CrossPoint-378k, a large-scale dataset spanning 900 real-world indoor scenes, 162 interaction types, and 8k object instances, and train CroPond on this dataset. Experiments demonstrate that CroPond significantly outperforms existing models on CrossPoint-Bench and exhibits strong generalization across other spatia and general VLMs benchmarks.

\textbf{Limitation and Outlook.}  CroPond is currently trained with a simple SFT scheme. Future work will explore advanced strategies, such as reinforcement learning, to enhance spatial reasoning. We further aim to extend CVPC beyond perception by integrating it with multi-agent task planning for collaborative robotics. 

\section*{Acknowledgments}
{\raggedright
This work was supported in part by the National Natural Science Foundation of China under Grant 72434005, Grant 72225011 and Grant 72293575.
\par}
{
    \small
    \bibliographystyle{ieeenat_fullname}
    \bibliography{arxiv_cvpr}

\begin{thebibliography}{95}
\providecommand{\natexlab}[1]{#1}
\providecommand{\url}[1]{\texttt{#1}}
\expandafter\ifx\csname urlstyle\endcsname\relax
  \providecommand{\doi}[1]{doi: #1}\else
  \providecommand{\doi}{doi: \begingroup \urlstyle{rm}\Url}\fi

\bibitem[An et~al.(2025{\natexlab{a}})An, Kim, Park, Jung, Han, Hong, and Kim]{an2025cross}
Honggyu An, Jin~Hyeon Kim, Seonghoon Park, Jaewoo Jung, Jisang Han, Sunghwan Hong, and Seungryong Kim.
\newblock Cross-view completion models are zero-shot correspondence estimators.
\newblock In \emph{Proceedings of the Computer Vision and Pattern Recognition Conference}, pages 1103--1115, 2025{\natexlab{a}}.

\bibitem[An et~al.(2025{\natexlab{b}})An, Xie, Yang, Zhang, Zhao, Cheng, Wang, Xu, Chen, Wu, et~al.]{an2025llava}
Xiang An, Yin Xie, Kaicheng Yang, Wenkang Zhang, Xiuwei Zhao, Zheng Cheng, Yirui Wang, Songcen Xu, Changrui Chen, Chunsheng Wu, et~al.
\newblock Llava-onevision-1.5: Fully open framework for democratized multimodal training.
\newblock \emph{arXiv preprint arXiv:2509.23661}, 2025{\natexlab{b}}.

\bibitem[{Anthropic}(2025)]{Claude-4}
{Anthropic}.
\newblock Introducing claude 4.
\newblock \url{https://www.anthropic.com/news/claude-4}, 2025.
\newblock Accessed: 2025-05-23.

\bibitem[Asim et~al.(2025)Asim, Wewer, Wimmer, Schiele, and Lenssen]{asim2025met3r}
Mohammad Asim, Christopher Wewer, Thomas Wimmer, Bernt Schiele, and Jan~Eric Lenssen.
\newblock Met3r: Measuring multi-view consistency in generated images.
\newblock In \emph{Proceedings of the Computer Vision and Pattern Recognition Conference}, pages 6034--6044, 2025.

\bibitem[Bai et~al.(2025{\natexlab{a}})Bai, Chen, Liu, Wang, Ge, Song, Dang, Wang, Wang, Tang, et~al.]{Qwen2.5-VL}
Shuai Bai, Keqin Chen, Xuejing Liu, Jialin Wang, Wenbin Ge, Sibo Song, Kai Dang, Peng Wang, Shijie Wang, Jun Tang, et~al.
\newblock Qwen2. 5-vl technical report.
\newblock \emph{arXiv preprint arXiv:2502.13923}, 2025{\natexlab{a}}.

\bibitem[Bai et~al.(2025{\natexlab{b}})Bai, Song, Chen, Ji, Zhong, Yang, Zhao, Zhou, Zhao, Li, et~al.]{bai2025towards}
Shuanghao Bai, Wenxuan Song, Jiayi Chen, Yuheng Ji, Zhide Zhong, Jin Yang, Han Zhao, Wanqi Zhou, Wei Zhao, Zhe Li, et~al.
\newblock Towards a unified understanding of robot manipulation: A comprehensive survey.
\newblock \emph{arXiv preprint arXiv:2510.10903}, 2025{\natexlab{b}}.

\bibitem[Baruch et~al.()Baruch, Chen, Dehghan, Feigin, Fu, Gebauer, Kurz, Dimry, Joffe, Schwartz, et~al.]{baruch1arkitscenes}
Gilad Baruch, Zhuoyuan Chen, Afshin Dehghan, Yuri Feigin, Peter Fu, Thomas Gebauer, Daniel Kurz, Tal Dimry, Brandon Joffe, Arik Schwartz, et~al.
\newblock Arkitscenes: A diverse real-world dataset for 3d indoor scene understanding using mobile rgb-d data.
\newblock In \emph{Thirty-fifth Conference on Neural Information Processing Systems Datasets and Benchmarks Track (Round 1)}.

\bibitem[Cabon et~al.(2025)Cabon, Stoffl, Antsfeld, Csurka, Chidlovskii, Revaud, and Leroy]{cabon2025must3r}
Yohann Cabon, Lucas Stoffl, Leonid Antsfeld, Gabriela Csurka, Boris Chidlovskii, Jerome Revaud, and Vincent Leroy.
\newblock Must3r: Multi-view network for stereo 3d reconstruction.
\newblock In \emph{Proceedings of the Computer Vision and Pattern Recognition Conference}, pages 1050--1060, 2025.

\bibitem[Cai et~al.(2024)Cai, Song, Guan, Chen, Li, Luo, Yi, and Kot]{cai2024benchlmm}
Rizhao Cai, Zirui Song, Dayan Guan, Zhenhao Chen, Yaohang Li, Xing Luo, Chenyu Yi, and Alex Kot.
\newblock Benchlmm: Benchmarking cross-style visual capability of large multimodal models.
\newblock In \emph{European Conference on Computer Vision}, pages 340--358. Springer, 2024.

\bibitem[Cao et~al.(2024)Cao, Yu, Wang, Xue, and Fu]{cao2024mvinpainter}
Chenjie Cao, Chaohui Yu, Fan Wang, Xiangyang Xue, and Yanwei Fu.
\newblock Mvinpainter: learning multi-view consistent inpainting to bridge 2d and 3d editing.
\newblock In \emph{Proceedings of the 38th International Conference on Neural Information Processing Systems}, pages 99299--99332, 2024.

\bibitem[Chen et~al.(2024{\natexlab{a}})Chen, Xu, Kirmani, Ichter, Sadigh, Guibas, and Xia]{chen2024spatialvlm}
Boyuan Chen, Zhuo Xu, Sean Kirmani, Brain Ichter, Dorsa Sadigh, Leonidas Guibas, and Fei Xia.
\newblock Spatialvlm: Endowing vision-language models with spatial reasoning capabilities.
\newblock In \emph{Proceedings of the IEEE/CVF Conference on Computer Vision and Pattern Recognition}, pages 14455--14465, 2024{\natexlab{a}}.

\bibitem[Chen et~al.(2025{\natexlab{a}})Chen, Xie, Ma, Sanketi, and Goldberg]{chen2025robo2vlm}
Kaiyuan Chen, Shuangyu Xie, Zehan Ma, Pannag~R Sanketi, and Ken Goldberg.
\newblock Robo2vlm: Visual question answering from large-scale in-the-wild robot manipulation datasets.
\newblock \emph{arXiv preprint arXiv:2505.15517}, 2025{\natexlab{a}}.

\bibitem[Chen et~al.(2024{\natexlab{b}})Chen, Li, Dong, Zhang, Zang, Chen, Duan, Wang, Qiao, Lin, et~al.]{chen2024we}
Lin Chen, Jinsong Li, Xiaoyi Dong, Pan Zhang, Yuhang Zang, Zehui Chen, Haodong Duan, Jiaqi Wang, Yu Qiao, Dahua Lin, et~al.
\newblock Are we on the right way for evaluating large vision-language models?
\newblock \emph{Advances in Neural Information Processing Systems}, 37:\penalty0 27056--27087, 2024{\natexlab{b}}.

\bibitem[Chen et~al.(2025{\natexlab{b}})Chen, Chi, Ji, Ye, Liu, Jia, Yu, and Cheng]{chen2025survey}
Weinan Chen, Wenzheng Chi, Sehua Ji, Hanjing Ye, Jie Liu, Yunjie Jia, Jiajie Yu, and Jiyu Cheng.
\newblock A survey of autonomous robots and multi-robot navigation: Perception, planning and collaboration.
\newblock \emph{Biomimetic Intelligence and Robotics}, 5\penalty0 (2):\penalty0 100203, 2025{\natexlab{b}}.

\bibitem[Cheng et~al.(2024)Cheng, Yin, Fu, Guo, Yang, Kautz, Wang, and Liu]{cheng2024spatialrgpt}
An-Chieh Cheng, Hongxu Yin, Yang Fu, Qiushan Guo, Ruihan Yang, Jan Kautz, Xiaolong Wang, and Sifei Liu.
\newblock Spatialrgpt: Grounded spatial reasoning in vision-language models.
\newblock \emph{Advances in Neural Information Processing Systems}, 37:\penalty0 135062--135093, 2024.

\bibitem[Dai et~al.(2017)Dai, Chang, Savva, Halber, Funkhouser, and Nie{\ss}ner]{dai2017scannet}
Angela Dai, Angel~X Chang, Manolis Savva, Maciej Halber, Thomas Funkhouser, and Matthias Nie{\ss}ner.
\newblock Scannet: Richly-annotated 3d reconstructions of indoor scenes.
\newblock In \emph{Proceedings of the IEEE conference on computer vision and pattern recognition}, pages 5828--5839, 2017.

\bibitem[Deitke et~al.(2025)Deitke, Clark, Lee, Tripathi, Yang, Park, Salehi, Muennighoff, Lo, Soldaini, et~al.]{deitke2025molmo}
Matt Deitke, Christopher Clark, Sangho Lee, Rohun Tripathi, Yue Yang, Jae~Sung Park, Mohammadreza Salehi, Niklas Muennighoff, Kyle Lo, Luca Soldaini, et~al.
\newblock Molmo and pixmo: Open weights and open data for state-of-the-art vision-language models.
\newblock In \emph{Proceedings of the Computer Vision and Pattern Recognition Conference}, pages 91--104, 2025.

\bibitem[Dong et~al.(2025)Dong, Chen, Lv, Yu, Zhang, Zhang, Zhu, Tian, Li, Moffatt, et~al.]{dong2025digital}
Zhao Dong, Ka Chen, Zhaoyang Lv, Hong-Xing Yu, Yunzhi Zhang, Cheng Zhang, Yufeng Zhu, Stephen Tian, Zhengqin Li, Geordie Moffatt, et~al.
\newblock Digital twin catalog: A large-scale photorealistic 3d object digital twin dataset.
\newblock In \emph{Proceedings of the Computer Vision and Pattern Recognition Conference}, pages 753--763, 2025.

\bibitem[Fan et~al.(2024)Fan, Zhang, Cong, Wang, Li, Wen, Zhou, Kadambi, Wang, Xu, et~al.]{fan2024large}
Zhiwen Fan, Jian Zhang, Wenyan Cong, Peihao Wang, Renjie Li, Kairun Wen, Shijie Zhou, Achuta Kadambi, Zhangyang Wang, Danfei Xu, et~al.
\newblock Large spatial model: End-to-end unposed images to semantic 3d.
\newblock \emph{Advances in neural information processing systems}, 37:\penalty0 40212--40229, 2024.

\bibitem[Fu et~al.(2025)Fu, Chen, Shen, Qin, Zhang, Lin, Yang, Zheng, Li, Sun, et~al.]{fu2025mme}
Chaoyou Fu, Peixian Chen, Yunhang Shen, Yulei Qin, Mengdan Zhang, Xu Lin, Jinrui Yang, Xiawu Zheng, Ke Li, Xing Sun, et~al.
\newblock Mme: A comprehensive evaluation benchmark for multimodal large language models.
\newblock In \emph{The Thirty-ninth Annual Conference on Neural Information Processing Systems Datasets and Benchmarks Track}, 2025.

\bibitem[Fu et~al.(2024)Fu, Hu, Li, Feng, Wang, Lin, Roth, Smith, Ma, and Krishna]{fu2024blink}
Xingyu Fu, Yushi Hu, Bangzheng Li, Yu Feng, Haoyu Wang, Xudong Lin, Dan Roth, Noah~A Smith, Wei-Chiu Ma, and Ranjay Krishna.
\newblock Blink: Multimodal large language models can see but not perceive.
\newblock In \emph{European Conference on Computer Vision}, pages 148--166. Springer, 2024.

\bibitem[Google(2025)]{gemini25pro}
Google.
\newblock Gemini 2.5 pro preview: even better coding performance.
\newblock \url{https://developers.googleblog.com/en/gemini-2-5-pro-io-improved-coding-performance/}, 2025.
\newblock Accessed: 2025-05-06.

\bibitem[Goto et~al.(2025)Goto, Hirose, Ukai, Kurita, and Inoue]{goto2025referring}
Kanoko Goto, Takumi Hirose, Mahiro Ukai, Shuhei Kurita, and Nakamasa Inoue.
\newblock Referring expression comprehension for small objects.
\newblock In \emph{Proceedings of the IEEE/CVF International Conference on Computer Vision}, pages 21231--21242, 2025.

\bibitem[Grauman et~al.(2024)Grauman, Westbury, Torresani, Kitani, Malik, Afouras, Ashutosh, Baiyya, Bansal, Boote, et~al.]{grauman2024ego}
Kristen Grauman, Andrew Westbury, Lorenzo Torresani, Kris Kitani, Jitendra Malik, Triantafyllos Afouras, Kumar Ashutosh, Vijay Baiyya, Siddhant Bansal, Bikram Boote, et~al.
\newblock Ego-exo4d: Understanding skilled human activity from first-and third-person perspectives.
\newblock In \emph{Proceedings of the IEEE/CVF Conference on Computer Vision and Pattern Recognition}, pages 19383--19400, 2024.

\bibitem[Gupta(2025)]{gupta2025embodied}
Satyandra~K Gupta.
\newblock Embodied ai for smart robotic cells in manufacturing applications.
\newblock In \emph{Proceedings of the AAAI Conference on Artificial Intelligence}, pages 28630--28636, 2025.

\bibitem[He et~al.(2024)He, Cascante-Bonilla, Yang, Berg, and Ordonez]{he2024improved}
Ruozhen He, Paola Cascante-Bonilla, Ziyan Yang, Alexander~C Berg, and Vicente Ordonez.
\newblock Improved visual grounding through self-consistent explanations.
\newblock In \emph{Proceedings of the IEEE/CVF Conference on Computer Vision and Pattern Recognition}, pages 13095--13105, 2024.

\bibitem[Hollidt et~al.(2024)Hollidt, Streli, Jiang, Haghighi, Qian, Liu, and Holz]{hollidt2024egosim}
Dominik Hollidt, Paul Streli, Jiaxi Jiang, Yasaman Haghighi, Changlin Qian, Xintong Liu, and Christian Holz.
\newblock Egosim: An egocentric multi-view simulator and real dataset for body-worn cameras during motion and activity.
\newblock \emph{Advances in Neural Information Processing Systems}, 37:\penalty0 106607--106627, 2024.

\bibitem[Hong et~al.(2024)Hong, Liu, Li, Li, and He]{hong2024multi}
Shixin Hong, Yu Liu, Zhi Li, Shaohui Li, and You He.
\newblock Multi-agent collaborative perception via motion-aware robust communication network.
\newblock In \emph{Proceedings of the IEEE/CVF Conference on Computer Vision and Pattern Recognition}, pages 15301--15310, 2024.

\bibitem[Hudson and Manning(2019)]{hudson2019gqa}
Drew~A Hudson and Christopher~D Manning.
\newblock Gqa: A new dataset for real-world visual reasoning and compositional question answering.
\newblock In \emph{Proceedings of the IEEE/CVF conference on computer vision and pattern recognition}, pages 6700--6709, 2019.

\bibitem[Ji et~al.(2025{\natexlab{a}})Ji, Tan, Chi, Xu, Zhao, Zhou, Lyu, Wang, Wang, Zhang, et~al.]{ji2025mathsticks}
Yuheng Ji, Huajie Tan, Cheng Chi, Yijie Xu, Yuting Zhao, Enshen Zhou, Huaihai Lyu, Pengwei Wang, Zhongyuan Wang, Shanghang Zhang, et~al.
\newblock Mathsticks: A benchmark for visual symbolic compositional reasoning with matchstick puzzles.
\newblock \emph{arXiv preprint arXiv:2510.00483}, 2025{\natexlab{a}}.

\bibitem[Ji et~al.(2025{\natexlab{b}})Ji, Tan, Shi, Hao, Zhang, Zhang, Wang, Zhao, Mu, An, et~al.]{ji2025robobrain}
Yuheng Ji, Huajie Tan, Jiayu Shi, Xiaoshuai Hao, Yuan Zhang, Hengyuan Zhang, Pengwei Wang, Mengdi Zhao, Yao Mu, Pengju An, et~al.
\newblock Robobrain: A unified brain model for robotic manipulation from abstract to concrete.
\newblock In \emph{Proceedings of the Computer Vision and Pattern Recognition Conference}, pages 1724--1734, 2025{\natexlab{b}}.

\bibitem[Ji et~al.(2025{\natexlab{c}})Ji, Wang, Liu, Hao, Liu, Zhao, Lyu, and Zheng]{ji2025visualtrans}
Yuheng Ji, Yipu Wang, Yuyang Liu, Xiaoshuai Hao, Yue Liu, Yuting Zhao, Huaihai Lyu, and Xiaolong Zheng.
\newblock Visualtrans: A benchmark for real-world visual transformation reasoning.
\newblock \emph{arXiv preprint arXiv:2508.04043}, 2025{\natexlab{c}}.

\bibitem[Kang et~al.(2025)Kang, Kim, Kim, and Hwang]{kang2025your}
Seil Kang, Jinyeong Kim, Junhyeok Kim, and Seong~Jae Hwang.
\newblock Your large vision-language model only needs a few attention heads for visual grounding.
\newblock In \emph{Proceedings of the Computer Vision and Pattern Recognition Conference}, pages 9339--9350, 2025.

\bibitem[Kembhavi et~al.(2016)Kembhavi, Salvato, Kolve, Seo, Hajishirzi, and Farhadi]{kembhavi2016diagram}
Aniruddha Kembhavi, Mike Salvato, Eric Kolve, Minjoon Seo, Hannaneh Hajishirzi, and Ali Farhadi.
\newblock A diagram is worth a dozen images.
\newblock In \emph{European conference on computer vision}, pages 235--251. Springer, 2016.

\bibitem[Kuang et~al.()Kuang, Li, and Iba]{kuang2024towards}
Nikki~Lijing Kuang, Songpo Li, and Soshi Iba.
\newblock Towards robust estimation of human intention hierarchy in robot teleoperation.
\newblock In \emph{NeurIPS 2024 Workshop on Behavioral Machine Learning}.

\bibitem[Lee et~al.(2025{\natexlab{a}})Lee, Xu, Richard, Wei, Saito, Bai, Wang, Sung, Kim, and Saragih]{lee2025rewind}
Jihyun Lee, Weipeng Xu, Alexander Richard, Shih-En Wei, Shunsuke Saito, Shaojie Bai, Te-Li Wang, Minhyuk Sung, Tae-Kyun Kim, and Jason Saragih.
\newblock Rewind: Real-time egocentric whole-body motion diffusion with exemplar-based identity conditioning.
\newblock In \emph{Proceedings of the Computer Vision and Pattern Recognition Conference}, pages 7095--7104, 2025{\natexlab{a}}.

\bibitem[Lee et~al.(2025{\natexlab{b}})Lee, Park, and Kim]{lee2025dynscene}
Sangmin Lee, Sungyong Park, and Heewon Kim.
\newblock Dynscene: Scalable generation of dynamic robotic manipulation scenes for embodied ai.
\newblock In \emph{Proceedings of the Computer Vision and Pattern Recognition Conference}, pages 12166--12175, 2025{\natexlab{b}}.

\bibitem[Li et~al.(2024{\natexlab{a}})Li, Yan, Yu, An, Wang, and Chen]{li2024comprehensive}
Tong Li, Yuhang Yan, Chengshun Yu, Jing An, Yifan Wang, and Gang Chen.
\newblock A comprehensive review of robot intelligent grasping based on tactile perception.
\newblock \emph{Robotics and Computer-Integrated Manufacturing}, 90:\penalty0 102792, 2024{\natexlab{a}}.

\bibitem[Li et~al.(2025)Li, Zheng, Xu, Gan, Lu, Li, Ni, Tian, Huang, Gao, et~al.]{li2025multi}
Wenqiao Li, Bozhong Zheng, Xiaohao Xu, Jinye Gan, Fading Lu, Xiang Li, Na Ni, Zheng Tian, Xiaonan Huang, Shenghua Gao, et~al.
\newblock Multi-sensor object anomaly detection: Unifying appearance, geometry, and internal properties.
\newblock In \emph{Proceedings of the Computer Vision and Pattern Recognition Conference}, pages 9984--9993, 2025.

\bibitem[Li et~al.(2024{\natexlab{b}})Li, Zhang, Geng, Geng, Long, Shen, Zhang, Liu, and Dong]{li2024manipllm}
Xiaoqi Li, Mingxu Zhang, Yiran Geng, Haoran Geng, Yuxing Long, Yan Shen, Renrui Zhang, Jiaming Liu, and Hao Dong.
\newblock Manipllm: Embodied multimodal large language model for object-centric robotic manipulation.
\newblock In \emph{Proceedings of the IEEE/CVF Conference on Computer Vision and Pattern Recognition}, pages 18061--18070, 2024{\natexlab{b}}.

\bibitem[Li et~al.(2023)Li, Du, Zhou, Wang, Zhao, and Wen]{li2023evaluating}
Yifan Li, Yifan Du, Kun Zhou, Jinpeng Wang, Wayne~Xin Zhao, and Ji-Rong Wen.
\newblock Evaluating object hallucination in large vision-language models.
\newblock In \emph{Proceedings of the 2023 Conference on Empirical Methods in Natural Language Processing}, pages 292--305, 2023.

\bibitem[Liang et~al.(2024)Liang, Ellis, and Henriques]{liang2024rapid}
Yichao Liang, Kevin Ellis, and Joao Henriques.
\newblock Rapid motor adaptation for robotic manipulator arms.
\newblock In \emph{Proceedings of the IEEE/CVF Conference on Computer Vision and Pattern Recognition}, pages 16404--16413, 2024.

\bibitem[Liu et~al.(2025)Liu, Teng, Xue, Wang, Zhu, Wang, and Wu]{liu2025mmcooper}
Bingyi Liu, Jian Teng, Hongfei Xue, Enshu Wang, Chuanhui Zhu, Pu Wang, and Libing Wu.
\newblock mmcooper: A multi-agent multi-stage communication-efficient and collaboration-robust cooperative perception framework.
\newblock \emph{arXiv preprint arXiv:2501.12263}, 2025.

\bibitem[Liu et~al.(2024{\natexlab{a}})Liu, Li, Li, and Lee]{liu2024improved}
Haotian Liu, Chunyuan Li, Yuheng Li, and Yong~Jae Lee.
\newblock Improved baselines with visual instruction tuning.
\newblock In \emph{Proceedings of the IEEE/CVF conference on computer vision and pattern recognition}, pages 26296--26306, 2024{\natexlab{a}}.

\bibitem[Liu et~al.(2024{\natexlab{b}})Liu, Zeng, Ren, Li, Zhang, Yang, Jiang, Li, Yang, Su, et~al.]{liu2024grounding}
Shilong Liu, Zhaoyang Zeng, Tianhe Ren, Feng Li, Hao Zhang, Jie Yang, Qing Jiang, Chunyuan Li, Jianwei Yang, Hang Su, et~al.
\newblock Grounding dino: Marrying dino with grounded pre-training for open-set object detection.
\newblock In \emph{European conference on computer vision}, pages 38--55. Springer, 2024{\natexlab{b}}.

\bibitem[Liu et~al.(2024{\natexlab{c}})Liu, Duan, Zhang, Li, Zhang, Zhao, Yuan, Wang, He, Liu, et~al.]{liu2024mmbench}
Yuan Liu, Haodong Duan, Yuanhan Zhang, Bo Li, Songyang Zhang, Wangbo Zhao, Yike Yuan, Jiaqi Wang, Conghui He, Ziwei Liu, et~al.
\newblock Mmbench: Is your multi-modal model an all-around player?
\newblock In \emph{European conference on computer vision}, pages 216--233. Springer, 2024{\natexlab{c}}.

\bibitem[Lu et~al.(2022)Lu, Mishra, Xia, Qiu, Chang, Zhu, Tafjord, Clark, and Kalyan]{lu2022learn}
Pan Lu, Swaroop Mishra, Tanglin Xia, Liang Qiu, Kai-Wei Chang, Song-Chun Zhu, Oyvind Tafjord, Peter Clark, and Ashwin Kalyan.
\newblock Learn to explain: Multimodal reasoning via thought chains for science question answering.
\newblock \emph{Advances in Neural Information Processing Systems}, 35:\penalty0 2507--2521, 2022.

\bibitem[Luo et~al.(2024)Luo, Cao, Khirodkar, Winkler, Kitani, and Xu]{luo2024real}
Zhengyi Luo, Jinkun Cao, Rawal Khirodkar, Alexander Winkler, Kris Kitani, and Weipeng Xu.
\newblock Real-time simulated avatar from head-mounted sensors.
\newblock In \emph{Proceedings of the IEEE/CVF Conference on Computer Vision and Pattern Recognition}, pages 571--581, 2024.

\bibitem[Lyu et~al.(2025)Lyu, Chen, Ji, and Xu]{lyu2025egoprompt}
Huaihai Lyu, Chaofan Chen, Yuheng Ji, and Changsheng Xu.
\newblock Egoprompt: Prompt pool learning for egocentric action recognition.
\newblock \emph{arXiv preprint arXiv:2508.03266}, 2025.

\bibitem[Majumder et~al.(2025)Majumder, Nagarajan, Al-Halah, Pradhan, and Grauman]{majumder2025viewpoint}
Sagnik Majumder, Tushar Nagarajan, Ziad Al-Halah, Reina Pradhan, and Kristen Grauman.
\newblock Which viewpoint shows it best? language for weakly supervising view selection in multi-view instructional videos.
\newblock In \emph{Proceedings of the Computer Vision and Pattern Recognition Conference}, pages 29016--29028, 2025.

\bibitem[Marino et~al.(2019)Marino, Rastegari, Farhadi, and Mottaghi]{marino2019ok}
Kenneth Marino, Mohammad Rastegari, Ali Farhadi, and Roozbeh Mottaghi.
\newblock Ok-vqa: A visual question answering benchmark requiring external knowledge.
\newblock In \emph{Proceedings of the IEEE/cvf conference on computer vision and pattern recognition}, pages 3195--3204, 2019.

\bibitem[{OpenAI}(2025)]{gpto3-o4-mini}
{OpenAI}.
\newblock Openai o3 and o4-mini system card.
\newblock \url{https://openai.com/index/introducing-o3-and-o4-mini/}, 2025.
\newblock Accessed: 2025-04-16.

\bibitem[Park et~al.(2025)Park, Tang, Das, Appalaraju, Singh, Manmatha, and Ghadar]{park2025r}
Joonhyung Park, Peng Tang, Sagnik Das, Srikar Appalaraju, Kunwar~Yashraj Singh, R Manmatha, and Shabnam Ghadar.
\newblock R-vlm: Region-aware vision language model for precise gui grounding.
\newblock In \emph{Findings of the Association for Computational Linguistics: ACL 2025}, pages 9669--9685, 2025.

\bibitem[Ravi et~al.(2025)Ravi, Gabeur, Hu, Hu, Ryali, Ma, Khedr, R{\"a}dle, Rolland, Gustafson, Mintun, Pan, Alwala, Carion, Wu, Girshick, Dollar, and Feichtenhofer]{ravi2025sam}
Nikhila Ravi, Valentin Gabeur, Yuan-Ting Hu, Ronghang Hu, Chaitanya Ryali, Tengyu Ma, Haitham Khedr, Roman R{\"a}dle, Chloe Rolland, Laura Gustafson, Eric Mintun, Junting Pan, Kalyan~Vasudev Alwala, Nicolas Carion, Chao-Yuan Wu, Ross Girshick, Piotr Dollar, and Christoph Feichtenhofer.
\newblock {SAM} 2: Segment anything in images and videos.
\newblock In \emph{The Thirteenth International Conference on Learning Representations}, 2025.

\bibitem[Ray et~al.(2025)Ray, Duan, II, Tan, Bashkirova, Hendrix, Ehsani, Kembhavi, Plummer, Krishna, Zeng, and Saenko]{SAT}
Arijit Ray, Jiafei Duan, Ellis L~Brown II, Reuben Tan, Dina Bashkirova, Rose Hendrix, Kiana Ehsani, Aniruddha Kembhavi, Bryan~A. Plummer, Ranjay Krishna, Kuo-Hao Zeng, and Kate Saenko.
\newblock {SAT}: Dynamic spatial aptitude training for multimodal language models.
\newblock In \emph{Second Conference on Language Modeling}, 2025.

\bibitem[Schops et~al.(2017)Schops, Schonberger, Galliani, Sattler, Schindler, Pollefeys, and Geiger]{schops2017multi}
Thomas Schops, Johannes~L Schonberger, Silvano Galliani, Torsten Sattler, Konrad Schindler, Marc Pollefeys, and Andreas Geiger.
\newblock A multi-view stereo benchmark with high-resolution images and multi-camera videos.
\newblock In \emph{Proceedings of the IEEE conference on computer vision and pattern recognition}, pages 3260--3269, 2017.

\bibitem[Shepard and Metzler(1971)]{shepard1971mental}
Roger~N Shepard and Jacqueline Metzler.
\newblock Mental rotation of three-dimensional objects.
\newblock \emph{Science}, 171\penalty0 (3972):\penalty0 701--703, 1971.

\bibitem[Singh et~al.(2019)Singh, Natarjan, Shah, Jiang, Chen, Parikh, and Rohrbach]{singh2019towards}
Amanpreet Singh, Vivek Natarjan, Meet Shah, Yu Jiang, Xinlei Chen, Devi Parikh, and Marcus Rohrbach.
\newblock Towards vqa models that can read.
\newblock In \emph{CVPR}, 2019.

\bibitem[Song et~al.(2025{\natexlab{a}})Song, Ouyang, Fang, Na, Shi, Chen, Yujie, Zhang, Jiang, Fang, et~al.]{song2025hazards}
Zirui Song, Guangxian Ouyang, Meng Fang, Hongbin Na, Zijing Shi, Zhenhao Chen, Fu Yujie, Zeyu Zhang, Shiyu Jiang, Miao Fang, et~al.
\newblock Hazards in daily life? enabling robots to proactively detect and resolve anomalies.
\newblock In \emph{Proceedings of the 2025 Conference of the Nations of the Americas Chapter of the Association for Computational Linguistics: Human Language Technologies (Volume 1: Long Papers)}, pages 7399--7415, 2025{\natexlab{a}}.

\bibitem[Song et~al.(2025{\natexlab{b}})Song, Ouyang, Li, Ji, Wang, Xu, Zhang, Zhang, Jiang, Chen, et~al.]{song2025maniplvm}
Zirui Song, Guangxian Ouyang, Mingzhe Li, Yuheng Ji, Chenxi Wang, Zixiang Xu, Zeyu Zhang, Xiaoqing Zhang, Qian Jiang, Zhenhao Chen, et~al.
\newblock Maniplvm-r1: Reinforcement learning for reasoning in embodied manipulation with large vision-language models.
\newblock \emph{arXiv preprint arXiv:2505.16517}, 2025{\natexlab{b}}.

\bibitem[Song et~al.(2025{\natexlab{c}})Song, Yang, Huang, Tonglet, Zhang, Cheng, Fang, Gurevych, and Chen]{song2025geolocation}
Zirui Song, Jingpu Yang, Yuan Huang, Jonathan Tonglet, Zeyu Zhang, Tao Cheng, Meng Fang, Iryna Gurevych, and Xiuying Chen.
\newblock Geolocation with real human gameplay data: A large-scale dataset and human-like reasoning framework.
\newblock \emph{arXiv preprint arXiv:2502.13759}, 2025{\natexlab{c}}.

\bibitem[Tan et~al.()Tan, Ji, Hao, Chen, Wang, Wang, and Zhang]{tanreason}
Huajie Tan, Yuheng Ji, Xiaoshuai Hao, Xiansheng Chen, Pengwei Wang, Zhongyuan Wang, and Shanghang Zhang.
\newblock Reason-rft: Reinforcement fine-tuning for visual reasoning of vision language models.
\newblock In \emph{The Thirty-ninth Annual Conference on Neural Information Processing Systems}.

\bibitem[Tan et~al.(2025{\natexlab{a}})Tan, Chi, Chen, Ji, Zhao, Hao, Lyu, Cao, Zhao, Lyu, et~al.]{tan2025roboosnext}
Huajie Tan, Cheng Chi, Xiansheng Chen, Yuheng Ji, Zhongxia Zhao, Xiaoshuai Hao, Yaoxu Lyu, Mingyu Cao, Junkai Zhao, Huaihai Lyu, et~al.
\newblock Roboos-next: A unified memory-based framework for lifelong, scalable, and robust multi-robot collaboration.
\newblock \emph{arXiv preprint arXiv:2510.26536}, 2025{\natexlab{a}}.

\bibitem[Tan et~al.(2025{\natexlab{b}})Tan, Hao, Chi, Lin, Lyu, Cao, Liang, Chen, Lyu, Peng, et~al.]{tan2025roboos}
Huajie Tan, Xiaoshuai Hao, Cheng Chi, Minglan Lin, Yaoxu Lyu, Mingyu Cao, Dong Liang, Zhuo Chen, Mengsi Lyu, Cheng Peng, et~al.
\newblock Roboos: A hierarchical embodied framework for cross-embodiment and multi-agent collaboration.
\newblock \emph{arXiv preprint arXiv:2505.03673}, 2025{\natexlab{b}}.

\bibitem[Team et~al.(2025{\natexlab{a}})Team, Cao, Tan, Ji, Chen, Lin, Li, Cao, Wang, Zhou, et~al.]{team2025robobrain}
BAAI~RoboBrain Team, Mingyu Cao, Huajie Tan, Yuheng Ji, Xiansheng Chen, Minglan Lin, Zhiyu Li, Zhou Cao, Pengwei Wang, Enshen Zhou, et~al.
\newblock Robobrain 2.0 technical report.
\newblock \emph{arXiv preprint arXiv:2507.02029}, 2025{\natexlab{a}}.

\bibitem[Team et~al.(2025{\natexlab{b}})Team, Abeyruwan, Ainslie, Alayrac, Arenas, Armstrong, Balakrishna, Baruch, Bauza, Blokzijl, et~al.]{GeminiRobotics}
Gemini~Robotics Team, Saminda Abeyruwan, Joshua Ainslie, Jean-Baptiste Alayrac, Montserrat~Gonzalez Arenas, Travis Armstrong, Ashwin Balakrishna, Robert Baruch, Maria Bauza, Michiel Blokzijl, et~al.
\newblock Gemini robotics: Bringing ai into the physical world.
\newblock \emph{arXiv preprint arXiv:2503.20020}, 2025{\natexlab{b}}.

\bibitem[Tong et~al.(2024)Tong, Brown, Wu, Woo, IYER, Akula, Yang, Yang, Middepogu, Wang, et~al.]{CV-Bench}
Peter Tong, Ellis Brown, Penghao Wu, Sanghyun Woo, Adithya Jairam~Vedagiri IYER, Sai~Charitha Akula, Shusheng Yang, Jihan Yang, Manoj Middepogu, Ziteng Wang, et~al.
\newblock Cambrian-1: A fully open, vision-centric exploration of multimodal llms.
\newblock \emph{Advances in Neural Information Processing Systems}, 37:\penalty0 87310--87356, 2024.

\bibitem[Wang et~al.(2025{\natexlab{a}})Wang, Chen, Karaev, Vedaldi, Rupprecht, and Novotny]{wang2025vggt}
Jianyuan Wang, Minghao Chen, Nikita Karaev, Andrea Vedaldi, Christian Rupprecht, and David Novotny.
\newblock Vggt: Visual geometry grounded transformer.
\newblock In \emph{Proceedings of the Computer Vision and Pattern Recognition Conference}, pages 5294--5306, 2025{\natexlab{a}}.

\bibitem[Wang et~al.(2024{\natexlab{a}})Wang, Chen, Zhao, and He]{wang2024scaling}
Lirui Wang, Xinlei Chen, Jialiang Zhao, and Kaiming He.
\newblock Scaling proprioceptive-visual learning with heterogeneous pre-trained transformers.
\newblock \emph{Advances in neural information processing systems}, 37:\penalty0 124420--124450, 2024{\natexlab{a}}.

\bibitem[Wang and Spelke(2002)]{wang2002human}
Ranxiao~Frances Wang and Elizabeth~S Spelke.
\newblock Human spatial representation: Insights from animals.
\newblock \emph{Trends in cognitive sciences}, 6\penalty0 (9):\penalty0 376--382, 2002.

\bibitem[Wang et~al.(2024{\natexlab{b}})Wang, Leroy, Cabon, Chidlovskii, and Revaud]{wang2024dust3r}
Shuzhe Wang, Vincent Leroy, Yohann Cabon, Boris Chidlovskii, and Jerome Revaud.
\newblock Dust3r: Geometric 3d vision made easy.
\newblock In \emph{Proceedings of the IEEE/CVF Conference on Computer Vision and Pattern Recognition}, pages 20697--20709, 2024{\natexlab{b}}.

\bibitem[Wang et~al.(2025{\natexlab{b}})Wang, Ma, Hou, Ding, Li, Chen, Chen, Peng, and Shen]{wang2025facebench}
Xiaoqin Wang, Xusen Ma, Xianxu Hou, Meidan Ding, Yudong Li, Junliang Chen, Wenting Chen, Xiaoyang Peng, and Linlin Shen.
\newblock Facebench: A multi-view multi-level facial attribute vqa dataset for benchmarking face perception mllms.
\newblock In \emph{Proceedings of the Computer Vision and Pattern Recognition Conference}, pages 9154--9164, 2025{\natexlab{b}}.

\bibitem[Wu et~al.(2024)Wu, Zhou, Kazanzides, Munawar, and Liu]{wu2024surgicai}
Jin Wu, Haoying Zhou, Peter Kazanzides, Adnan Munawar, and Anqi Liu.
\newblock Surgicai: A hierarchical platform for fine-grained surgical policy learning and benchmarking.
\newblock \emph{Advances in Neural Information Processing Systems}, 37:\penalty0 63771--63789, 2024.

\bibitem[Xia et~al.(2025)Xia, Lin, Xiang, Huang, Chen, Dong, Wang, and Wen]{xia2025learning}
Qiming Xia, Wenkai Lin, Haoen Xiang, Xun Huang, Siheng Chen, Zhen Dong, Cheng Wang, and Chenglu Wen.
\newblock Learning to detect objects from multi-agent lidar scans without manual labels.
\newblock In \emph{Proceedings of the Computer Vision and Pattern Recognition Conference}, pages 1418--1428, 2025.

\bibitem[Xu et~al.(2024)Xu, Chen, Tu, and Yang]{xu2024v2x}
Runsheng Xu, Chia-Ju Chen, Zhengzhong Tu, and Ming-Hsuan Yang.
\newblock V2x-vitv2: Improved vision transformers for vehicle-to-everything cooperative perception.
\newblock \emph{IEEE transactions on pattern analysis and machine intelligence}, 2024.

\bibitem[Xu et~al.(2025)Xu, Zhu, and Yang]{xu2025mc}
Yunqiu Xu, Linchao Zhu, and Yi Yang.
\newblock Mc-bench: A benchmark for multi-context visual grounding in the era of mllms.
\newblock In \emph{Proceedings of the IEEE/CVF International Conference on Computer Vision}, pages 17675--17687, 2025.

\bibitem[Yang et~al.(2025{\natexlab{a}})Yang, Li, Yang, Zhang, Hui, Zheng, Yu, Gao, Huang, Lv, et~al.]{Qwen3-VL}
An Yang, Anfeng Li, Baosong Yang, Beichen Zhang, Binyuan Hui, Bo Zheng, Bowen Yu, Chang Gao, Chengen Huang, Chenxu Lv, et~al.
\newblock Qwen3 technical report.
\newblock \emph{arXiv preprint arXiv:2505.09388}, 2025{\natexlab{a}}.

\bibitem[Yang et~al.(2025{\natexlab{b}})Yang, Yang, Gupta, Han, Fei-Fei, and Xie]{yang2025thinking}
Jihan Yang, Shusheng Yang, Anjali~W Gupta, Rilyn Han, Li Fei-Fei, and Saining Xie.
\newblock Thinking in space: How multimodal large language models see, remember, and recall spaces.
\newblock In \emph{Proceedings of the Computer Vision and Pattern Recognition Conference}, pages 10632--10643, 2025{\natexlab{b}}.

\bibitem[Yang et~al.(2025{\natexlab{c}})Yang, Xu, Xie, Yang, Li, Lin, Zhu, Chen, Duan, Yue, et~al.]{yang2025mmsi}
Sihan Yang, Runsen Xu, Yiman Xie, Sizhe Yang, Mo Li, Jingli Lin, Chenming Zhu, Xiaochen Chen, Haodong Duan, Xiangyu Yue, et~al.
\newblock Mmsi-bench: A benchmark for multi-image spatial intelligence.
\newblock \emph{arXiv preprint arXiv:2505.23764}, 2025{\natexlab{c}}.

\bibitem[Yang et~al.(2024)Yang, Liu, Lin, Hancke, and Lau]{yang2024boosting}
Zaiquan Yang, Yuhao Liu, Jiaying Lin, Gerhard Hancke, and Rynson Lau.
\newblock Boosting weakly supervised referring image segmentation via progressive comprehension.
\newblock \emph{Advances in Neural Information Processing Systems}, 37:\penalty0 93213--93239, 2024.

\bibitem[Yeh et~al.(2025)Yeh, Wang, Tong, Cheng, Wang, Chu, Zhai, Chen, Gao, and Ma]{yeh2025seeing-All-Angles}
Chun-Hsiao Yeh, Chenyu Wang, Shengbang Tong, Ta-Ying Cheng, Ruoyu Wang, Tianzhe Chu, Yuexiang Zhai, Yubei Chen, Shenghua Gao, and Yi Ma.
\newblock Seeing from another perspective: Evaluating multi-view understanding in mllms.
\newblock \emph{arXiv preprint arXiv:2504.15280}, 2025.

\bibitem[Yeshwanth et~al.(2023)Yeshwanth, Liu, Nie{\ss}ner, and Dai]{yeshwanth2023scannet++}
Chandan Yeshwanth, Yueh-Cheng Liu, Matthias Nie{\ss}ner, and Angela Dai.
\newblock Scannet++: A high-fidelity dataset of 3d indoor scenes.
\newblock In \emph{Proceedings of the IEEE/CVF International Conference on Computer Vision}, pages 12--22, 2023.

\bibitem[Yuan et~al.(2025)Yuan, Duan, Blukis, Pumacay, Krishna, Murali, Mousavian, and Fox]{yuan2025robopoint}
Wentao Yuan, Jiafei Duan, Valts Blukis, Wilbert Pumacay, Ranjay Krishna, Adithyavairavan Murali, Arsalan Mousavian, and Dieter Fox.
\newblock Robopoint: A vision-language model for spatial affordance prediction in robotics.
\newblock In \emph{Conference on Robot Learning}, pages 4005--4020. PMLR, 2025.

\bibitem[Yue et~al.(2024)Yue, Ni, Zhang, Zheng, Liu, Zhang, Stevens, Jiang, Ren, Sun, et~al.]{yue2024mmmu}
Xiang Yue, Yuansheng Ni, Kai Zhang, Tianyu Zheng, Ruoqi Liu, Ge Zhang, Samuel Stevens, Dongfu Jiang, Weiming Ren, Yuxuan Sun, et~al.
\newblock Mmmu: A massive multi-discipline multimodal understanding and reasoning benchmark for expert agi.
\newblock In \emph{Proceedings of the IEEE/CVF Conference on Computer Vision and Pattern Recognition}, pages 9556--9567, 2024.

\bibitem[Yue et~al.(2025)Yue, Zheng, Ni, Wang, Zhang, Tong, Sun, Yu, Zhang, Sun, et~al.]{yue2025mmmu}
Xiang Yue, Tianyu Zheng, Yuansheng Ni, Yubo Wang, Kai Zhang, Shengbang Tong, Yuxuan Sun, Botao Yu, Ge Zhang, Huan Sun, et~al.
\newblock Mmmu-pro: A more robust multi-discipline multimodal understanding benchmark.
\newblock In \emph{Proceedings of the 63rd Annual Meeting of the Association for Computational Linguistics (Volume 1: Long Papers)}, pages 15134--15186, 2025.

\bibitem[Zhang et~al.(2024{\natexlab{a}})Zhang, Chen, Wang, Liu, Wang, and Qiao]{zhang20244diffusion}
Haiyu Zhang, Xinyuan Chen, Yaohui Wang, Xihui Liu, Yunhong Wang, and Yu Qiao.
\newblock 4diffusion: Multi-view video diffusion model for 4d generation.
\newblock \emph{Advances in Neural Information Processing Systems}, 37:\penalty0 15272--15295, 2024{\natexlab{a}}.

\bibitem[Zhang et~al.(2025{\natexlab{a}})Zhang, Chen, Xu, Huang, Mei, Chen, Zhou, Yuan, Cai, Huang, Quan, Xu, and Zhang]{zhang2025from}
Jiahui Zhang, Yurui Chen, Yueming Xu, Ze Huang, Jilin Mei, Junhui Chen, Yanpeng Zhou, Yu-Jie Yuan, Xinyue Cai, Guowei Huang, Xingyue Quan, Hang Xu, and Li Zhang.
\newblock From flatland to space: Teaching vision-language models to perceive and reason in 3d.
\newblock In \emph{The Thirty-ninth Annual Conference on Neural Information Processing Systems Datasets and Benchmarks Track}, 2025{\natexlab{a}}.

\bibitem[Zhang et~al.(2025{\natexlab{b}})Zhang, Huang, Xu, Huang, Zhi, Ren, Xu, and Zhang]{zhang2025mllms}
Wanyue Zhang, Yibin Huang, Yangbin Xu, JingJing Huang, Helu Zhi, Shuo Ren, Wang Xu, and Jiajun Zhang.
\newblock Why do mllms struggle with spatial understanding? a systematic analysis from data to architecture.
\newblock \emph{arXiv preprint arXiv:2509.02359}, 2025{\natexlab{b}}.

\bibitem[Zhang et~al.(2025{\natexlab{c}})Zhang, Ng, Ma, Wang, Zhao, Koenecke, Li, and Wanglu]{zhang-etal-2025-sphere}
Wenyu Zhang, Wei~En Ng, Lixin Ma, Yuwen Wang, Junqi Zhao, Allison Koenecke, Boyang Li, and Wanglu Wanglu.
\newblock {SPHERE}: Unveiling spatial blind spots in vision-language models through hierarchical evaluation.
\newblock In \emph{Proceedings of the 63rd Annual Meeting of the Association for Computational Linguistics (Volume 1: Long Papers)}, pages 11591--11609, Vienna, Austria, 2025{\natexlab{c}}. Association for Computational Linguistics.

\bibitem[Zhang et~al.(2024{\natexlab{b}})Zhang, Ma, Gao, Shakiah, Gao, and Chai]{zhang2024groundhog}
Yichi Zhang, Ziqiao Ma, Xiaofeng Gao, Suhaila Shakiah, Qiaozi Gao, and Joyce Chai.
\newblock Groundhog: Grounding large language models to holistic segmentation.
\newblock In \emph{Proceedings of the IEEE/CVF conference on computer vision and pattern recognition}, pages 14227--14238, 2024{\natexlab{b}}.

\bibitem[Zhao et~al.(2025)Zhao, Lu, Kim, Fu, Zhang, Wu, Li, Ma, Han, Finn, et~al.]{zhao2025cot}
Qingqing Zhao, Yao Lu, Moo~Jin Kim, Zipeng Fu, Zhuoyang Zhang, Yecheng Wu, Zhaoshuo Li, Qianli Ma, Song Han, Chelsea Finn, et~al.
\newblock Cot-vla: Visual chain-of-thought reasoning for vision-language-action models.
\newblock In \emph{Proceedings of the Computer Vision and Pattern Recognition Conference}, pages 1702--1713, 2025.

\bibitem[Zheng et~al.(2020)Zheng, Wei, and Yang]{zheng2020university}
Zhedong Zheng, Yunchao Wei, and Yi Yang.
\newblock University-1652: A multi-view multi-source benchmark for drone-based geo-localization.
\newblock In \emph{Proceedings of the 28th ACM international conference on Multimedia}, pages 1395--1403, 2020.

\bibitem[Zhong et~al.(2025)Zhong, Wang, Xu, Li, Nie, and Yu]{zhong2025cooptrack}
Jiaru Zhong, Jiahao Wang, Jiahui Xu, Xiaofan Li, Zaiqing Nie, and Haibao Yu.
\newblock Cooptrack: Exploring end-to-end learning for efficient cooperative sequential perception.
\newblock In \emph{Proceedings of the IEEE/CVF International Conference on Computer Vision}, pages 26954--26965, 2025.

\bibitem[Zhou et~al.(2025{\natexlab{a}})Zhou, An, Chi, Han, Rong, Zhang, Wang, Wang, Huang, Sheng, et~al.]{zhou2025roborefer}
Enshen Zhou, Jingkun An, Cheng Chi, Yi Han, Shanyu Rong, Chi Zhang, Pengwei Wang, Zhongyuan Wang, Tiejun Huang, Lu Sheng, et~al.
\newblock Roborefer: Towards spatial referring with reasoning in vision-language models for robotics.
\newblock \emph{arXiv preprint arXiv:2506.04308}, 2025{\natexlab{a}}.

\bibitem[Zhou et~al.(2025{\natexlab{b}})Zhou, Wang, Liang, Bai, and Zhang]{zhou2025cross}
Ziyang Zhou, Pinghui Wang, Zi Liang, Haitao Bai, and Ruofei Zhang.
\newblock Cross-modal 3d representation with multi-view images and point clouds.
\newblock In \emph{Proceedings of the Computer Vision and Pattern Recognition Conference}, pages 3728--3739, 2025{\natexlab{b}}.

\end{thebibliography}
}

\clearpage
\appendix

\section*{Appendix}
This supplementary material provides details of CrossPoint-Bench, CrossPoint-378K and CroPond. It complements the main paper by offering implementation details and
additional resources to support reproducibility and further
research. The document is organized as follows:
\begin{itemize}
    \item \textbf{Sec.A} describes the implementation details of the dataset and benchmark, including their data generation pipeline and construction procedures. 

    \item \textbf{Sec.B} provides implementation details of CroPond, including its architecture, training data, and training setup. 
    
    \item \textbf{Sec.C} presents evaluation settings and experimental details that complement the main paper. 

    \item \textbf{Sec.D} includes exploratory analyses of error patterns.

    \item \textbf{Sec.E} provides additional visualizations of prompts, CrossPoint-Bench, and CrossPoint-378K.

    \item \textbf{Sec.F} summarizes future directions, focusing on training strategies and extensions toward multi-agent planning. 
\end{itemize}

\begin{center}
  \includegraphics[width=\linewidth]{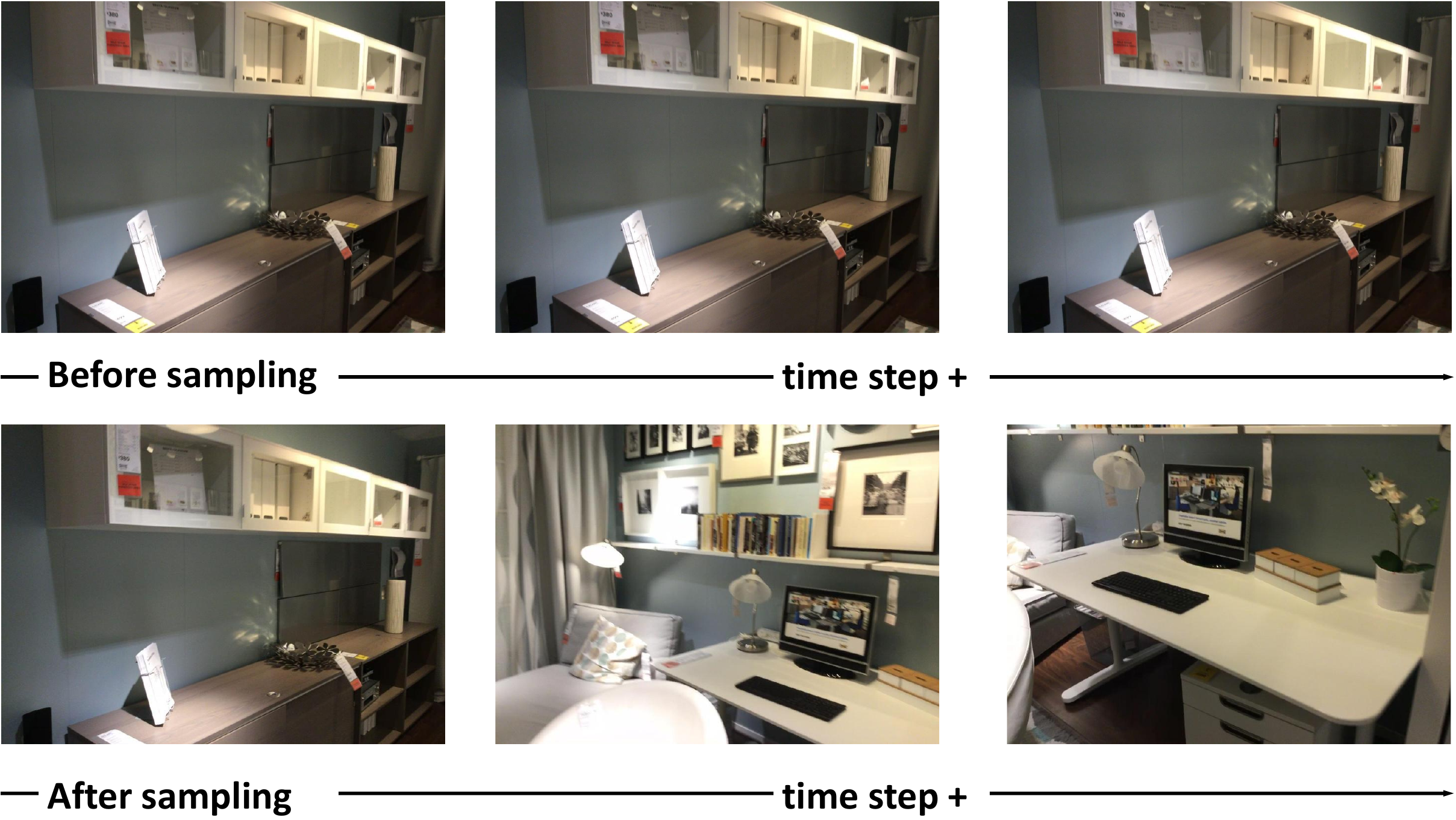}
  \captionof{figure}{\textbf{Comparison of frame sequences before and after sampling.} Top: three consecutive frames from the original video. Bottom: corresponding frames after sampling, exhibiting increased temporal variation between adjacent frames.}
  \label{fig:sample}
\end{center}

\section{Implementation Details of Dataset and Benchmark}
\label{sec:data_generation}
\subsection{Data Source Description }
We use ScanNet and ScanNet++ as the data sources for CrossPoint-Bench and CrossPoint-378K, and employ ETH3D and ARKitScenes as the data sources for CrossPoint-LR-Bench.
\begin{itemize}
    \item \textbf{ScanNet} \cite{dai2017scannet}: A large-scale indoor RGB-D dataset containing 1.5k scenes, 2.5M frames, and 3D reconstructions with semantic annotations. It is widely used for indoor perception tasks such as 3D reconstruction, semantic parsing, and object detection.

    \item \textbf{ScanNet++} \cite{yeshwanth2023scannet++}: A high-fidelity extension of ScanNet built with multi-camera, high-resolution capture and photogrammetric reconstruction. It provides improved geometric accuracy and photometric consistency, enabling reliable benchmarks in depth estimation, semantic parsing, and multi-view reasoning.

    \item \textbf{ETH3D} \cite{schops2017multi}: A high-precision benchmark for multi-view stereo and visual odometry that covers diverse indoor and outdoor scenes with accurate camera poses and dense geometric annotations. It is commonly used for evaluating stereo matching and high-quality 3D reconstruction.

    \item \textbf{ARKitScenes} \cite{baruch1arkitscenes}: A large-scale mobile RGB-D dataset collected with Apple ARKit devices, consisting of over 5k indoor scans with scale-aligned RGB-D data. It supports a wide range of mobile perception tasks, including object detection, semantic segmentation, and scene-level 3D reconstruction.
\end{itemize}

\begin{figure}[t]
    \centering
    \includegraphics[width=\linewidth]{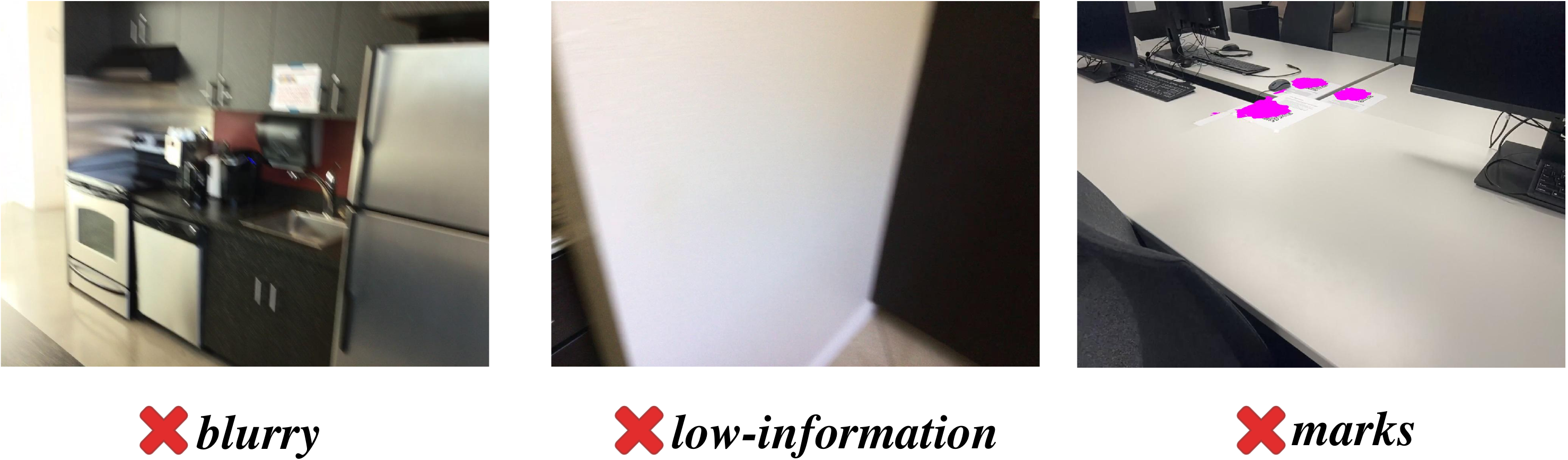}
    \caption{Examples of three types of undesirable images.}
    \label{fig:filter}
\end{figure}

\subsection{Data Generation Pipeline}
\label{sec:pipeline}

\begin{figure}[t]
    \centering
    \includegraphics[width=\linewidth]{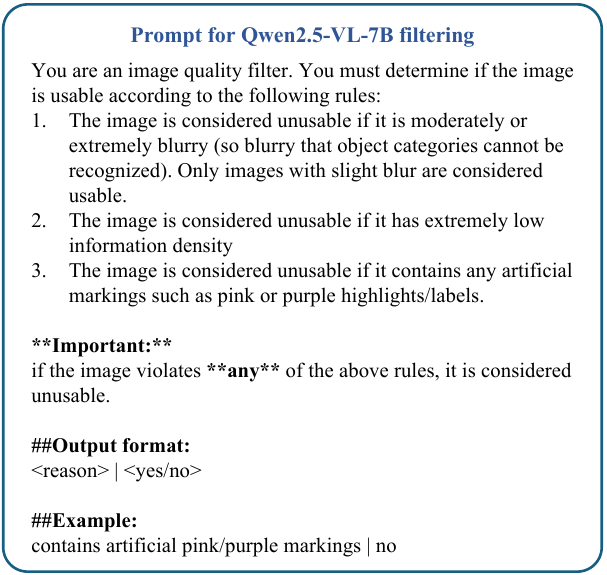}
    \caption{Prompt for Qwen2.5-VL-7B filtering.}
    \label{fig:prompt_filter}
\end{figure}

\begin{itemize}

    \item \textbf{Image Sampling and Filtering:} 
    We first downsample the high-frame-rate 3D videos~\cite{dai2017scannet,yeshwanth2023scannet++} to reduce temporal redundancy. These videos contain consecutive frames that are often nearly identical, with only minimal visual variation, which would otherwise introduce a large number of redundant samples and provide limited informational gain during training. To address this issue, we adopt a frame-sampling strategy that adjusts the sampling rate based on redundancy, using an average ratio of 1/50. The effect before and after sampling is shown in Fig.~\ref{fig:sample}.

    After sampling, the images undergo an additional filtering stage to further improve dataset quality. As illustrated in Fig.~\ref{fig:filter}, we filter out three types of undesirable images: 1) blurry images in which the scene or key objects cannot be clearly identified; 2) low-information images that contain little meaningful visual content; and 3) images containing many visual marks that may interfere with model training. This filtering is performed using Qwen2.5-VL-7B~\cite{Qwen2.5-VL}, with the prompt shown in Fig.~\ref{fig:prompt_filter}, resulting in a curated set of 23k high-quality images.
    
\end{itemize}

\begin{figure}[t]
    \centering
    \includegraphics[width=\linewidth]{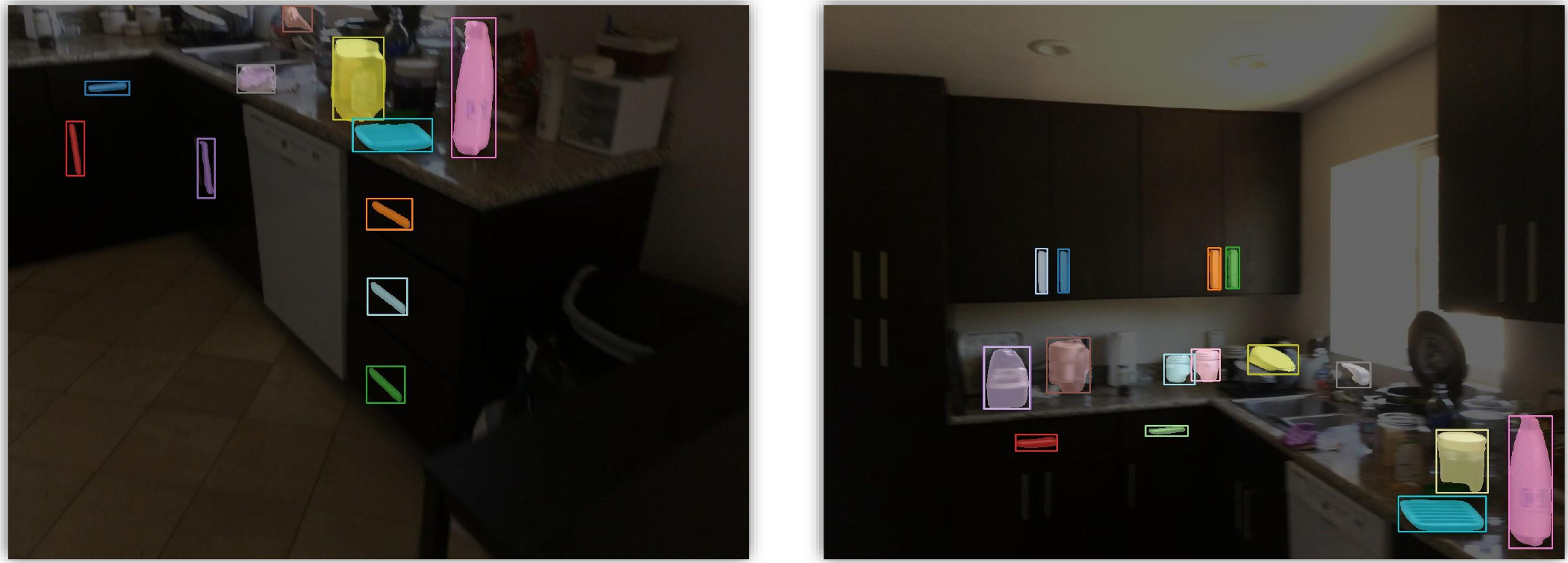}
    \caption{Examples of masks for affordance regions, including general objects (e.g., bottles) and semantic parts (e.g., drawer handles).}
    \label{fig:mask}
\end{figure}

\begin{itemize}
    \item \textbf{Affordance Region Segmentation:} To obtain fine-grained annotations of both general objects and semantic parts, we employ a multi-stage process for mask generation. We first utilize Gemini-2.5-Pro~\cite{gemini25pro} to identify operable regions, with the prompting strategy illustrated in Fig.~\ref{fig:prompt_recognize}. Subsequently, Grounding-DINO~\cite{liu2024grounding} is used to generate bounding boxes for fine-grained objects; since this model does not support part-level detection, RoboBrain2.0~\cite{team2025robobrain} is incorporated to extract bounding boxes for semantic parts. To further ensure the accuracy of the bounding boxes and the quality of the generated data, annotators manually verified the correctness of all bounding boxes. These bounding boxes are then provided as auxiliary cues to SAM2.1~\cite{ravi2025sam}, which segments the corresponding regions and yields 58k fine-grained mask annotations. Representative masks, including both general objects and semantic parts, are shown in Fig.~\ref{fig:mask}.
\end{itemize}

\begin{itemize}
    \item \textbf{Cross-View Pairing:} For each instance mask in an image, we randomly sample a point and compute its cross-view projection using the geometric information provided by the corresponding dataset.
    
    For ScanNet, the mapping is performed using RGB--D frames: the sampled pixel is first associated with its depth value and back-projected into the source camera coordinate system. The point is then transformed into the coordinate frame of every other view using the camera poses, and finally reprojected onto the target images using their intrinsic parameters.
    
    For ScanNet++, the mapping relies on the SfM reconstruction produced by COLMAP. Each pixel is linked to a unique 3D world point through the feature tracks recorded in the reconstruction files. The 3D world point is then projected into all other images by applying the corresponding camera extrinsic and intrinsic matrices, yielding its 2D correspondence across views.

    By applying these two mapping strategies to all frames within each scene, we construct a total of 111k one-to-one cross-view correspondence pairs.
\end{itemize}

\begin{table*}[t]
\centering
\caption{\textbf{Templates used to generate question-answer pair.} \{\} are placeholders in the templates to be replaced with actual content when generating questions.}

\renewcommand{\arraystretch}{1.2}
\begin{tabular}{>{\centering\arraybackslash}m{0.25\linewidth} >{\raggedright\arraybackslash}m{0.7\linewidth}} 
\hline
\textbf{Task Type} & \multicolumn{1}{c}{\textbf{Template Examples}} \\
\hline

\textbf{Single Fine Grounding} &
[Q]: Where would you \textit{\{action\}} the \textit{\{object\}}? [A]: \textit{\{(x, y)\}} \newline
[Q]: Where is the \textit{\{object\}} located? [A]: \textit{\{(x, y)\}} \newline
[Q]: Could you provide the coordinates of the \textit{\{object\}}? [A]: \textit{\{(x, y)\}} \newline
[Q]: Find where to \textit{\{action\}} the \textit{\{object\}}. [A]: \textit{\{(x, y)\}} \\
\hline

\textbf{Single Spatial Understanding} &
[Q]: What object is at \textit{\{(x, y)\}}? [A]: It refers to \textit{\{object\}}. \newline
[Q]: Which object does \textit{\{(x, y)\}} point to? [A]: \textit{\{object\}} \newline
[Q]: What does the point at \textit{\{(x, y)\}} refer to? [A]: This point indicates the \textit{\{part\}}. \newline
[Q]: Which point is closer to the camera? [A]: \textit{\{Blue point/Red point\}} \newline
[Q]: Rank the marked points by their 3D distance to the camera. [A]: \textit{\{order\}} \\
\hline

\textbf{Cross Occlusion Visibility} &
[Q]: Is the point marked in image1 still visible in image2? [A]: \textit{\{Yes/No\}} \newline
[Q]: Is the location in image1 out of view in image2? [A]: \textit{\{Yes/No\}} \newline
[Q]: Which objects become occluded in image2? [A]: \textit{\{object list\}} \newline
[Q]: Which point corresponds to an occluded part in image2? [A]: \textit{\{color point\}} \\
\hline

\textbf{Cross Correspondence} &
[Q]: Does the point in image1 correspond to the point in image2? [A]: \textit{\{Yes/No\}} \newline
[Q]: Which point in image2 aligns with the point in image1? [A]: \textit{\{color point\}} \newline
[Q]: Which point corresponds to the same 3D location? [A]: \textit{\{color point\}} \newline
[Q]: Where is the corresponding point for \textit{\{(x, y)\}} in image1? [A]: \textit{\{(x', y')\}} \newline
[Q]: Where can you find the point matching the red dot in image1? [A]:\textit{\{(x, y)\}} \\
\hline

\textbf{Cross Spatial Transformation} &
[Q]: How did the camera rotate? [A]: The rotation direction is {\textit{\{direction\}}}. \newline
[Q]: In which direction is the viewpoint rotating? [A]: It rotated toward \textit{\{direction\}}. \newline
[Q]: Which direction would the cameraman need to move? [A]: \textit{\{direction\}}. \newline
[Q]: What movement of the camera would produce the second image? [A]:\textit{\{direction\}}. \\ 
\hline

\textbf{Cross Depth Variation} &
[Q]: Is the corresponding point closer or farther in image2? [A]: \textit{\{Closer/Farther\}} \newline
[Q]: For the same physical point, is it nearer or farther in image2? [A]: \textit{\{answer\}} \newline
[Q]: The corresponding 3D
location appears at what relative position? [A]: \textit{\{answer\}} \\
\hline

\end{tabular}
\label{tab:template} 
\end{table*}

\begin{itemize}
    \item \textbf{QA Generation:} We treat the cross-view correspondence pairs obtained in the previous stage as the base units for all multi-view tasks, and consider each individual image as the base unit for single-view tasks. Across both settings, we employ several families of question–answer templates, some of which contain placeholders. These placeholders are populated using object categories, action types, or pixel-level coordinates extracted in the previous stage. In addition, depth-based judgments and camera-motion assessments are computed directly from the camera parameters. The template specification is shown in Tab.~\ref{tab:template}. Following this procedure, we generate a total of 378k high-quality QA samples aligned with the semantics of each task type.

\end{itemize}

\subsection{CrossPoint-Bench Details}
CrossPoint-Bench is constructed using the same data generation pipeline as CrossPoint-378K (see Sec.~\ref{sec:pipeline}). Specifically, we select 100 scenes from ScanNet and ScanNet++ that do not appear in CrossPoint-378K. After manual verification, we obtain 1,000 unambiguous and high-quality samples that serve as our benchmark.

For CrossPoint-LR-Bench, we build the benchmark through manual construction, drawing on ETH3D and ARKitScenes as source datasets. We curate 90 long-range reasoning samples tailored to evaluate the model’s capability under extended spatial correspondence conditions.

\section{Implementation Details for CroPond}
\subsection{Architecture}
The model is based on Qwen-2.5‑VL~\cite{Qwen2.5-VL}, comprising a vision encoder, a large language model, and a multimodal projector. The vision encoder is a ViT with patch size 14, enhanced with 2D‑RoPE for spatially aware positional encoding and supporting dynamic image resolutions, which is crucial for fine-grained point prediction. The LLM is initialized from the pre-trained Qwen2.5. To improve efficiency, four adjacent patch features are grouped and projected via a two-layer MLP into the LLM embedding dimension, reducing sequence length while preserving spatial structure.

\subsection{Training Data}

\begin{figure}[t]
    \centering
    \includegraphics[width=\linewidth]{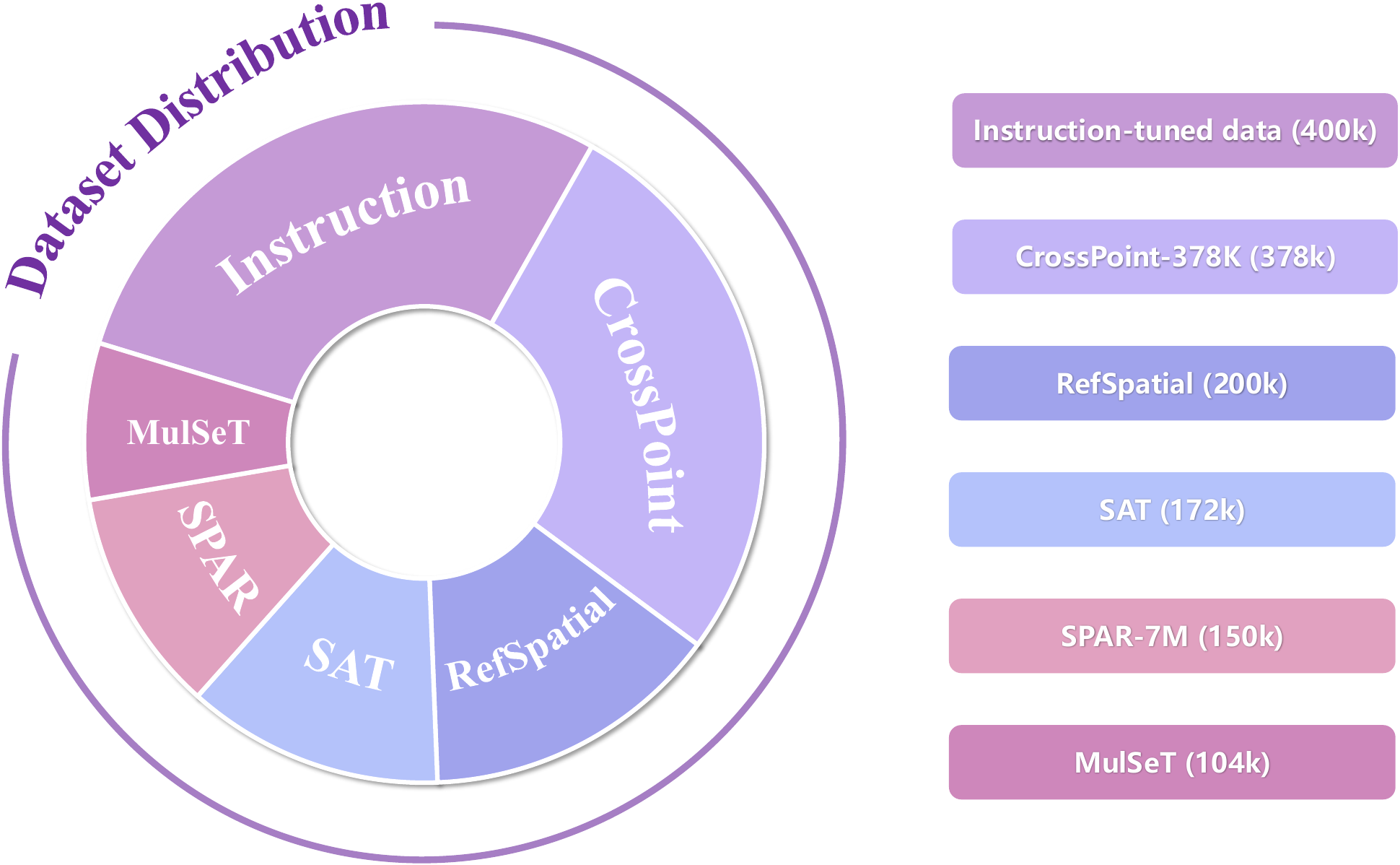}
    \caption{Training data distribution of CroPond. By jointly leveraging multi-source data, CroPond achieves CVPC capability while generalizing across spatial and general tasks.}

    \label{fig:training_data}
\end{figure}

We train CroPond using a comprehensive multi-source dataset, as illustrated in Fig.~\ref{fig:training_data}. CrossPoint-378K provides the supervision needed for acquiring core CVPC capabilities. To further enhance spatial reasoning, we incorporate 200k samples from RefSpatial~\cite{zhou2025roborefer} and 172k samples from the SAT~\cite{SAT} training set, which strengthen the model’s understanding of spatial relations, distance estimation, and related concepts. We also include auxiliary multi-view data, consisting of 104k samples from MulSeT~\cite{zhang2025mllms} and 150k from multi view part of SPAR-7M~\cite{zhang2025from}. Finally, to improve instruction-following behavior and general visual-language understanding, we integrate 400k instruction-tuned samples derived from LLaVA-1.5~\cite{liu2024improved}.

\subsection{SFT Training Details}
We perform the supervised fine-tuning (SFT) stage on our model using a dataset 
$D$ consisting of image-text pairs $(O, Q, A)$, where $O$ represents an image (RGB), $Q$ is a textual prompt or question, and $A$ is the corresponding answer. The training objective is to maximize the likelihood of generating the answer given the input pair:

\begin{equation}
\mathcal{L}_{\text{SFT}} = - \mathbb{E}_{(O, Q, A) \sim D} \sum_{t=1}^{T} \log \pi_\theta(y_t \mid O, Q, y_{<t}),
\end{equation}
where $\pi_\theta$ is the model's token distribution.

All model parameters are updated during fine-tuning. We adopt full-parameter fine-tuning with a per-device batch size of 1 and gradient accumulation over 2 steps, effectively resulting in a batch size of 2. The training uses a cosine learning rate scheduler with a maximum learning rate of $1\times 10^{-5}$, weight decay of 0.1, and 100 warm-up steps. The maximum sequence length is set to 4096 tokens. To accelerate training and reduce memory footprint, we enable bf16 precision and utilize DeepSpeed with ZeRO Stage 3 optimization.

\section{Experimental Setting and Details}
\label{sec:experimental_setting}
\subsection{Experiments Compute Resources}
All experiments were conducted on a server equipped with 2 Intel Xeon Platinum 8468 CPUs, each with 48 cores, and 8 NVIDIA H100 80GB HBM3 GPUs. The 3B model required approximately 39 hours of training, while the 7B model required around 49 hours.

\subsection{Effectiveness of CrossPoint-378K}

We first evaluate the effectiveness of CrossPoint-378K on the CrossPoint-Bench. As shown in Tab.~\ref{tab:crosspointbench}, fine-tuning Qwen2.5-VL-3B solely on CrossPoint-378K leads to dramatic improvements across all metrics. The overall score increases from 25.40\% to 72.90\%, with particularly notable gains in Correspondence-Judge and Correspondence-Pointing. These results demonstrate that CrossPoint-378K provides strong cross-view geometric supervision, enabling the base model to build substantially more robust and consistent point-level correspondences across viewpoints.

To assess whether CrossPoint-378K also yields benefits for other cross-view and spatial-understanding tasks, we further evaluate the model on three spatial understanding benchmarks: CV-Bench~\cite{CV-Bench}, BLINK~\cite{fu2024blink} validation set, and SAT~\cite{SAT}. As summarized in Tab.~\ref{tab:cvbench}, Tab.~\ref{tab:blink}, and Tab.~\ref{tab:sat_effective}, the fine-tuned model consistently outperforms the base model on the majority of tasks. These improvements indicate that CrossPoint-378K not only strengthens CVPC capability, but also enhances broader cross-view understanding and spatial perception.

\begin{table}[t]
\centering
\scriptsize
\setlength{\tabcolsep}{4pt}
\renewcommand{\arraystretch}{1.2}
\caption{CrossPoint-Bench results for Qwen2.5-VL-3B and the model trained with CrossPoint-378K.}
\begin{tabular}{lccccc}
\hline
Model & Overall & Ground. & Visibility. & Judge. & Point. \\
\hline
Base Model & 25.40 & 52.80 & 47.27 & 33.97 & 2.59 \\
\ \ +CrossPoint-378K & \textbf{72.90} & \textbf{56.52} & \textbf{79.09} & \textbf{80.13} & \textbf{73.22} \\
\hline
$\Delta$ Gain & \textbf{+47.50} & \textbf{+3.72} & \textbf{+31.82} & \textbf{+46.16} & \textbf{+70.63} \\
\hline
\end{tabular}
\label{tab:crosspointbench}
\end{table}

\begin{table}[t]
\centering
\scriptsize
\setlength{\tabcolsep}{4pt}
\renewcommand{\arraystretch}{1.2}
\caption{CV-Bench performance comparison between Qwen2.5-VL-3B and the model enhanced with CrossPoint-378K.}
\begin{tabular}{lcccc}
\hline
Model & Overall & 3D-Depth & 3D-Distance & 2D-Relation \\
\hline
Base Model & 67.08 & 77.17 & 52.50 & 71.23 \\
\ \ +CrossPoint-378K & \textbf{74.00} & \textbf{81.50} & \textbf{68.33} & \textbf{72.31} \\
\hline
$\Delta$ Gain & \textbf{+6.92} & \textbf{+4.33} & \textbf{+15.83} & \textbf{+1.08} \\
\hline
\end{tabular}
\label{tab:cvbench}
\end{table}

\begin{table}[t]
\centering
\scriptsize
\setlength{\tabcolsep}{4pt}
\renewcommand{\arraystretch}{1.2}
\caption{BLINK\textsubscript{val} performance comparison between Qwen2.5-VL-3B and the model enhanced with CrossPoint-378K.}
\begin{tabular}{lccccc}
\hline
Model & Overall & Multi-view & Rel.Depth & SpatialRel. & Vis.Corr. \\
\hline
Base Model & 55.24 & 44.36 & 62.90 & 79.72 & 37.79 \\
\ \ +CrossPoint-378K & \textbf{77.80} & \textbf{96.24} & \textbf{71.77} & 60.84 & \textbf{81.98} \\
\hline
$\Delta$ Gain & \textbf{+22.56} & \textbf{+51.88} & \textbf{+8.87} & -18.88 & \textbf{+44.19} \\
\hline
\end{tabular}
\label{tab:blink}
\end{table}

\begin{table}[t]
\centering
\scriptsize
\setlength{\tabcolsep}{4pt}
\renewcommand{\arraystretch}{1.2}
\caption{SAT performance comparison between Qwen2.5-VL-3B and the model enhanced with CrossPoint-378K.}
\begin{tabular}{lcccccc}
\hline
Model & Overall & Action & EgoMove & GoalAim & ObjMove & Persp. \\
\hline
Base Model & 56.67 & 63.51 & 52.17 & 66.18 & 47.83 & 48.48 \\
\ \ +CrossPoint-378K & \textbf{63.00} & 43.24 & \textbf{100.00} & \textbf{76.47} & 47.83 & \textbf{56.06} \\
\hline
$\Delta$ Gain & \textbf{+6.33} & -20.27 & \textbf{+47.83} & \textbf{+10.29} & 0.00 & \textbf{+7.58} \\
\hline
\end{tabular}
\label{tab:sat_effective}
\end{table}

\subsection{Effectiveness of Vanilla CoT}

We further investigate the effect of CoT prompting (i.e., \emph{``Please output your reasoning process step by step''}) on the performance of various models on CrossPoint-Bench. We select three representative models, and the results are summarized in Tab.~\ref{tab:cot}. We observe no consistent trend across models: although some sub-tasks show minor improvements, the overall gains remain limited. This suggests that CoT prompting alone does not provide substantial benefits for CVPC tasks, highlighting the need to leverage CrossPoint-378K to enforce cross-view consistency in spatial reasoning.

\begin{table}[t]
\centering
\scriptsize
\setlength{\tabcolsep}{4pt}
\renewcommand{\arraystretch}{1.2}

\caption{\textbf{Performance comparison on CrossPoint-Bench with and without CoT.}
Arrows indicate performance changes compared to the non-CoT setting. Qwen3-VL-235B refers to Qwen3-VL-235B-A22B-Thinking, 
Qwen2.5-VL-7B refers to Qwen2.5-VL-7B-Instruct.}
\label{tab:cot}

\begin{tabular}{lccccc}
\hline
\textbf{Model} & \textbf{Overall} & \textbf{Ground.} &
\textbf{Visibility.} & \textbf{Judge.} &
\textbf{Point.} \\
\hline

\makecell[l]{Gemini-2.5-Pro}
& 37.10 & 32.92 & 67.27 & 60.26 & 16.41 \\

\ \ + CoT
& 35.90 ↓ & 36.65 ↑ & 63.64 ↓ & 58.33 ↓ & 14.90 ↓ \\
\hline

\makecell[l]{Qwen3-VL-235B}
& 52.70 & 65.22 & 62.27 & 66.03 & 39.31 \\

\ \ + CoT
& 54.35 ↑ & 66.88 ↑ & 59.55 ↓ & 63.46 ↓ & 42.45 ↑ \\
\hline

\makecell[l]{Qwen2.5-VL-7B}
& 26.80 & 48.45 & 57.27 & 41.03 & 0.00 \\

\ \ + CoT
& 27.30 ↑ & 44.72 ↓ & 55.91 ↓ & 45.51 ↑ & 1.51 ↑ \\
\hline

\end{tabular}
\end{table}

\renewcommand{\arraystretch}{1.25} 
\setlength{\tabcolsep}{3pt} 
\begin{table*}[htbp]
\footnotesize
    \centering
    \caption{\textbf{Performance of different models on SPAR-Bench.} Top-1 \& Top-2 accuracies are represented using \textbf{bold text}, and \underline{underlined}. SPAR-Bench (\textit{tiny}) refers to a subset of the full benchmark. Results are taken from the original paper of SPAR-Bench~\cite{zhang2025from}.}
    \small
    \resizebox{1.0\textwidth}{!}{
    \begin{tabular}{l| c c | c c c c c c c c c | c c c c | c c c c c c c c c c}
        \toprule
        \textbf{Model} & \rotatebox{75}{\textbf{Rank}} & \normalsize\rotatebox{75}{\textbf{Avg.}} & \normalsize\rotatebox{75}{\textbf{Low}} & 
       \scriptsize \rotatebox{75}{Depth-OC} & \scriptsize \rotatebox{75}{Depth-OC-MV} & \scriptsize \rotatebox{75}{Depth-OO} & \scriptsize \rotatebox{75}{Depth-OO-MV} & \scriptsize \rotatebox{75}{Dist-OC} & \scriptsize \rotatebox{75}{Dist-OC-MV} & \scriptsize \rotatebox{75}{Dist-OO} & \scriptsize \rotatebox{75}{Dist-OO-MV} &
       \normalsize \rotatebox{75}{\textbf{Medium}} &
       \scriptsize \rotatebox{75}{PosMatch} & \scriptsize \rotatebox{75}{CamMotion} & \scriptsize \rotatebox{75}{ViewChgI} &
       \normalsize \rotatebox{75}{\textbf{High}} & \scriptsize \rotatebox{75}{DistI-OO} & \scriptsize \rotatebox{75}{DistI-OO-MV} & \scriptsize \rotatebox{75}{ObjRel-OC-MV} & \scriptsize \rotatebox{75}{ObjRel-OO} & \scriptsize \rotatebox{75}{ObjRel-OO-MV} & \scriptsize \rotatebox{75}{SpImag-OC} & \scriptsize \rotatebox{75}{SpImag-OC-MV} & \scriptsize \rotatebox{75}{SpImag-OO} & \scriptsize \rotatebox{75}{SpImag-OO-MV} \\
        \midrule
        \rowcolor[HTML]{F2F2F2} \multicolumn{26}{l}{\textbf{Baseline}}\\
        Chance Level (Random) & -& -& -& -& -& -& -& -& -& -& -&- &22.65&24.50& -& 25.09 &23.82& 22.02 & 31.25 & 25.27 & 22.16 & 25.81 & 24.42& 24.17 & 26.89\\
        Chance Level (Frequency) & - & 32.74 & 31.19 & 43.09 & 43.51 & 17.38 & 13.05 & 41.90 & 30.99 & 27.40 & 32.17 & 38.25 &29.01 & 26.75 & 59.00 &32.29& 52.94 & 50.60 & 28.25 & 26.92 & 26.59 & 26.34 & 26.74 & 26.49 & 25.77 \\
        \midrule
        \rowcolor[HTML]{F2F2F2} \multicolumn{26}{l}{\textbf{SPAR-Bench (tiny) API}}\\
        Human Level & 1 & \textbf{67.27} & \textbf{55.31} & \textbf{72.75} & \textbf{74.25} & \textbf{28.75} & \textbf{36.25} & \textbf{78.25} & \underline{52.25} & \textbf{66.5} & \textbf{33.5} & \textbf{72.32} & \textbf{92} & \textbf{64} & \textbf{60.97} & \textbf{76.22} & \textbf{80} & \textbf{94} & \textbf{70} & \textbf{92} & \textbf{80} & \textbf{78} & \textbf{82} & \textbf{50} & \textbf{60} \\
        GPT-4o & 8 & 36.39 & 29.25 & \underline{53.8} & 45 & 15 & 13.6 & 37.4 & 34.4 & 23.4 & \underline{24.4} & \underline{24.93} & 30 & 16 & \underline{28.8} & 45.11 & 64 & 64 & 58 & 46 & 46 & \underline{32} & \underline{44} & \underline{30} & 22 \\
        Claude-3.7-Sonnet & 17 & 21.77 & 25.43 & 41 & 45.4 & 11.2 & 12.2 & 42.6 & 19.6 & 26 & 5.4 & 7.33 & 16 & 6 & 0 & 23.33 & 40 & 48 & 22 & 36 & 14 & 12 & 20 & 6 & 12 \\
        Qwen2-VL-72B & 10 & 35.62 & 35.28 & 45.4 & \underline{49.8} & 13.8 & 10 & \underline{54.6} & 49.4 & \underline{36.8} & 22.4 & 23.39 & \underline{42} & \underline{18} & 10.16 & 40 & 60 & 68 & 50 & 38 & 44 & 18 & 28 & 18 & 36 \\
        Qwen2.5-VL-72B & 5 & \underline{39.4} & \underline{35.35} & 53.2 & 46.8 & \underline{17.8} & \underline{29} & 49.6 & \textbf{57.4} & 14.4 & 14.6 & 23.05 & 40 & 16 & 13.16 & \underline{48.44} & \underline{74} & \underline{74} & \underline{60} & \underline{56} & \underline{50} & 20 & 34 & 24 & \underline{44} \\
        \midrule
        \rowcolor[HTML]{F2F2F2} \multicolumn{26}{l}{\textbf{SPAR-Bench (full)}}\\
        GPT-4o & 6 & 38.11 & \underline{36.88} & \textbf{51.22} & \textbf{44.69} & 21.21 & 19.33 & 41.4 & \underline{44.9} & \underline{36.34} & \underline{35.96} & 26.49 & 27.74 & 25.25 & 19.99 & 43.8 & 65 & 64.88 & 44.75 & 50.82 & 43.21 & 29.84 & 32.56 & 27.81 & 35.29 \\
        GPT-4.1 & 4 & 41.6 & \textbf{41.95} & \underline{48.25} & 42.66 & 20.86 & \textbf{23.58} & \textbf{58.29} & \textbf{52.33} & \textbf{48.65} & \textbf{40.44} & 44.02 & 59.29 & 28.75 & 22.02 & 42.93 & \underline{71.76} & \textbf{67.26} & 46.25 & 54.95 & 41 & 30.38 & 29.65 & 20.2 & 24.93 \\
        InternVL2.5-2B & 16 & 30.14 & 25.79 & 39.67 & 39.72 & 12.12 & 15.03 & 30.94 & 29.59 & 20.22 & 19.02 & 22.93 & 37.91 & 24.25 & 6.64 & 36.41 & 51.47 & 56.85 & 50.25 & 33.79 & 24.1 & 27.15 & 35.17 & 26.49 & 22.41 \\
        InternVL2.5-4B & 15 & 30.55 & 25.66 & 29.06 & 32.97 & 21.77 & 16.83 & 20.84 & 26.85 & 28.13 & 28.79 & 29.75 & 47.07 & 33.25 & 8.92 & 35.16 & 54.12 & 58.93 & 35.5 & 29.67 & 34.63 & 24.73 & 31.39 & 19.21 & 28.29 \\
        InternVL2.5-8B & 9 & 36.28 & 29.46 & 25.78 & 29.31 & \underline{23.79} & 18.76 & \underline{46.82} & 42.68 & 22.62 & 25.89 & 31.88 & 61.32 & 28 & 6.32 & 43.8 & 59.71 & 56.85 & 51.75 & 44.23 & 41.55 & 36.56 & 41.57 & 22.52 & 39.5 \\
        InternVL2.5-26B & 11 & 34.11 & 24.18 & 41.83 & 36.28 & 18.09 & 17.96 & 22.62 & 17.97 & 22.01 & 16.63 & 50.11 & 69.47 & 30.75 & 7.03 & 42.4 & 62.94 & 58.33 & 45.25 & 54.67 & 50.97 & 24.46 & 25 & 27.48 & 32.49 \\
        InternVL2.5-38B & 12 & 33.83 & 26 & 42.03 & 38.81 & 17.1 & 17.77 & 17.58 & 17.84 & 29.37 & 27.51 & 30.89 & 44.53 & 17.25 & 9.7 & 44.13 & 69.12 & \underline{66.67} & 43.75 & 64.29 & 37.67 & 25.27 & 31.98 & 31.79 & 26.61 \\
        Qwen2.5-VL-7B & 14 & 33.07 & 28.75 & 31.33 & 33.66 & 21.99 & 14.97 & 42.88 & 37.73 & 23.83 & 23.64 & 22.97 & 33.33 & 28.75 & 6.83 & 40.27 & 58.24 & 51.49 & 44.75 & 50 & 32.13 & 33.87 & 32.85 & 27.15 & 31.93 \\
        Qwen2.5-VL-32B & 13 & 33.09 & 27.09 & 37.47 & 35.25 & 15.91 & 15.75 & 34.2 & 33.01 & 26.2 & 18.96 & 33.92 & 34.1 & 33.75 & 13.15 & 40.44 & 58.82 & 61.31 & 37.75 & 51.1 & 34.35 & 26.08 & 33.14 & 25.83 & 35.57 \\
        Qwen2.5-VL-72B & 7 & 37.01 & 29.94 & 37.47 & \underline{43} & 19.52 & 18.36 & 38.72 & 36.44 & 27.8 & 18.25 & 44.61 & 56.49 & 32.75 & 17.27 & 43.8 & 58.82 & 61.9 & 40.75 & 53.57 & 45.98 & 26.88 & 35.17 & 34.11 & 36.97 \\
        \midrule
        \rowcolor[HTML]{F2F2F2} \multicolumn{26}{l}{\textbf{CroPond Variants (full)}}\\
        CroPond-3B & 3 & \underline{52.67} & 22 & 30.56 & 30.31 & \textbf{26.88} & 20.97 & 10.43 & 16.71 & 23.97 & 16.16 & \underline{57.77} & \underline{75.57} & \underline{71} & \underline{26.75} & \underline{78.23} & \textbf{77.06} & 66.37 & \underline{89.25} & \underline{86.54} & \textbf{88.37} & \underline{75} & \underline{68.9} & \underline{68.54} & \underline{84.03} \\
        CroPond-7B & 2 & \textbf{53.64} & 19.22 & 23.33 & 27.81 & 21.51 & \underline{21.24} & 5.34 & 14.79 & 21.21 & 18.52 & \textbf{66.55} & \textbf{79.39} & \textbf{80.25} & \textbf{40} & \textbf{79.94} & 69.41 & 61.9 & \textbf{89.75} & \textbf{89.01} & \underline{87.53} & \textbf{75.81} & \textbf{76.74} & \textbf{82.45} & \textbf{86.83} \\
        \bottomrule
    \end{tabular}
}
    \label{tab:sparbench_res}
\end{table*}

\begin{table}[htbp]
\centering
\caption{\textbf{Performance of different models on SAT.} Top-1 \& Top-2 accuracies are represented using \textbf{bold text}, and \underline{underlined}. Persp. denotes perspective.}
\label{tab:sat}
\resizebox{\columnwidth}{!}{%
\begin{tabular}{lrrrrrr}
\hline
    \textbf{Model} & \textbf{Overall} & \textbf{Action} & \textbf{EgoMove} & \textbf{GoalAim} & \textbf{ObjMove} & \textbf{Persp.} \\
    \midrule
    \rowcolor[HTML]{F2F2F2} \multicolumn{7}{l}{\textbf{Proprietary Models}}\\
    Claude-3.7-Sonnet              & 60.00 & 59.46 & 60.87 & 70.59 & 63.04 & 46.97 \\
    Gemini-2.5-Pro                 & \underline{79.00} & \textbf{85.14} & \textbf{100.00} & 77.94 & \textbf{73.91} & \textbf{62.12} \\
    GPT-4o                         & 53.67 & 54.05 & 41.30 & 54.41 & 65.22 & 53.03 \\
    \midrule
    \rowcolor[HTML]{F2F2F2} \multicolumn{7}{l}{\textbf{Open-Source Vision-Language Models}}\\
    Qwen2.5-VL-3B-Instruct         & 56.67 & 63.51 & 52.17 & 66.18 & 47.83 & 48.48 \\
    Qwen2.5-VL-7B-Instruct         & 56.00 & 50.00 & 78.26 & 70.59 & 50.00 & 36.36 \\
    Qwen2.5-VL-32B-Instruct        & 65.00 & 72.97 & 63.04 & 82.35 & 47.83 & 51.52 \\
    Qwen2.5-VL-72B-Instruct        & 69.00 & 75.68 & 69.57 & 85.29 & 60.87 & 50.00 \\
    Qwen3-VL-30B-A3B-Instruct      & 69.39 & 70.27 & \underline{93.18} & 80.88 & 64.29 & 43.94 \\
    Qwen3-VL-235B-A22B-Instruct    & \textbf{80.00} & \underline{83.78} & \textbf{100.00} & 88.24 & \underline{71.74} & \underline{59.09} \\
    \midrule
    \rowcolor[HTML]{F2F2F2} \multicolumn{7}{l}{\textbf{CroPond Variants}}\\
    CroPond-3B                  & 78.33 & 82.43 & \textbf{100.00} & \textbf{92.65} & 58.70 & 57.58 \\
    CroPond-7B                  & 73.00 & 58.11 & \textbf{100.00} & \underline{89.71} & 65.22 & \underline{59.09} \\
\hline
\end{tabular}%
}
\end{table}

\subsection{Evaluation of CrossPoint-LR-Bench}
\label{sec:LR-Bench}

We provide the evaluation details of CrossPoint-LR-Bench, with the exact prompting procedure illustrated in Fig.~\ref{fig:prompt_lr}. Each sample consists of three rounds of questions, corresponding to the three stages defined in the CVPC task. A prediction is considered correct only if the model answers all three rounds correctly, making the benchmark markedly more challenging.

\subsection{Evaluation on Spatial VLM Benchmarks}
\label{sec:Spatial_Bench}
We evaluate several spatial understanding benchmarks, including CV-Bench~\cite{CV-Bench}, the BLINK validation set~\cite{fu2024blink}, SAT~\cite{SAT}, and SPAR-Bench~\cite{zhang2025from}. We select all sub-tasks that are closely aligned with CVPC, along with other spatial reasoning tasks, and exclude non-spatial tasks from our evaluation, such as 2D Counting in CV-Bench or Art Style and IQ Test in BLINK. All evaluations follow the official protocols of the corresponding benchmarks. For SPAR-Bench, we directly report the results provided in~\cite{zhang2025from}. In this section, we additionally include the detailed results for individual sub-tasks in SAT and SPAR-Bench that were not expanded upon in the main paper, as shown in Tab.~\ref{tab:sparbench_res}, ~\ref{tab:sat}.

CroPond shows solid generalization across spatial reasoning tasks, with strong performance on those closely related to CVPC (such as Position Matching in SPAR-Bench and Visual Correspondence in BLINK). It does not show notable performance on the Low-level category of SPAR-Bench, as these tasks involve precise metric estimation, which is not the focus of CroPond.

\subsection{Evaluation on General VLM Benchmarks}
\label{sec:General_Bench}

We further compare CroPond-3B with its base model, Qwen2.5-VL-3B, on several general vision–language benchmarks to study how spatially aware supervision affects overall VQA performance. The detailed results are presented in Tab.~\ref{tab:general}. The results show that CroPond-3B achieves comparable performance across a wide range of general tasks, and even surpasses the base model such as OK-VQA and MMBench\textsubscript{en\_dev}. This demonstrates that spatial VQA training on CrossPoint-378K, together with other multi-source visual instruction datasets, can substantially improve the model’s spatial understanding and CVPC capability without compromising its general vision–language abilities.

\begin{table}[htbp]
  \centering
  \caption{Comparison of CroPond and base model performance on general benchmarks.}
  \setlength{\tabcolsep}{6pt}
  \renewcommand{\arraystretch}{1.1}

  {%
  \fontsize{8}{10}\selectfont   

  \begin{tabular}{lcc}
    \toprule
    \textbf{Benchmark} & \textbf{Qwen2.5-VL-3B} & \textbf{Cropond-3B} \\
    \midrule
    OK\-VQA~\cite{marino2019ok}                             & 42.56     & \textbf{54.93} \\
    POPE~\cite{li2023evaluating}                            & 87.59     & \textbf{87.82} \\
    GQA~\cite{hudson2019gqa}                                & 59.93     & \textbf{62.44} \\
    MME~\cite{fu2025mme}                                    & 2150      & 2131  \\
    ScienceQA\textsubscript{img}~\cite{lu2022learn}         & 81.16     & 80.27 \\
    MMBench\textsubscript{en\_dev}~\cite{liu2024mmbench}    & 78.44     & \textbf{81.10} \\
    MMMU\textsubscript{val}~\cite{yue2024mmmu}              & 46.80     & \textbf{47.40} \\
    MMStar~\cite{chen2024we}                                & 56.39     & 56.12 \\
    TextVQA~\cite{singh2019towards}                         & 78.67     & 77.35 \\
    Ai2d~\cite{kembhavi2016diagram}                         & 78.69     & 78.63 \\
    MMMU\textsubscript{pro}~\cite{yue2025mmmu}              & 31.04     & 30.17 \\
    \bottomrule
  \end{tabular}
  }
\label{tab:general}
\end{table}

\section{Exploratory Experiments on Error Patterns}
    \label{sec:exploratory}

    To better understand why current models fail on CrossPoint-Bench, we categorize their errors into three patterns: \emph{Frame Transfer Failure}, \emph{Spatial Reconstruction Failure} and \emph{Semantic--Point Decoupling}. We then conduct an exploratory intervention in which we augment the original prompts with \emph{error-aware suffixes} that explicitly draw attention to the pitfalls associated with each pattern.

\subsection{Experimental Setup}

    We start from a manually annotated subset of CrossPoint-Bench, where each example is labeled as one of the three error types or \emph{No Error}. For every example that was mispredicted, we re-evaluate the corresponding model while appending an additional natural-language suffix to the original prompt. The suffix is chosen according to the previously assigned error type:
    
    \begin{itemize}
        \item \textbf{Frame Transfer Failure.} \\
        \emph{``Use the second image as the reference when outputting the coordinates, ensuring that all positions are defined consistently within the coordinate system of this second image and not any of the others.''}
        \item \textbf{Spatial Reconstruction Failure.} \\
        \emph{``Pay close attention to occlusions and the relative positions between objects, carefully reasoning about which parts are hidden, which are visible, and how objects overlap or align with each other in the scene.''}
        \item \textbf{Semantic--Point Decoupling.} \\
        \emph{``Precisely determine the location in the image based on your reasoning, giving coordinates that follow logically from the visual evidence and your step-by-step analysis of the scene structure, and ensure that the location you infer during this reasoning is consistent with the red dot marker shown in image 1.''}
    \end{itemize}

    All images and answer formats are kept unchanged; only the textual prompt is modified. After re-running the models, we re-label each case using the same set of categories as in our annotation protocol and report results for three model groups: \emph{closed-source}, \emph{open-source ($>$7B)}, and \emph{open-source ($\leq$7B)}.

\subsection{Quantitative Results}

    \begin{table}[t]
        \centering
        \caption{\textbf{Error-aware prompting scores before and after adding suffixes for different model groups.} Frame, Spatial, and Point denote Frame Transfer Failure, Spatial Reconstruction Failure, and Semantic--Point Decoupling, respectively.}
        \label{tab:d1-error-distribution}
        {\fontsize{8}{9.6}\selectfont
        \begin{tabular}{l l c c c}
            \toprule
            \textbf{Model group} & \textbf{Setting} & \textbf{Frame} & \textbf{Spatial} & \textbf{Point} \\
            \midrule
            \multirow{2}{*}{Closed-source} 
                & Before & 12 & 28 & 26 \\
                & After  & 12 & 26 & 19 \\
            \midrule
            \multirow{2}{*}{Open-source ($>$7B)} 
                & Before & 16 & 18 & 24 \\
                & After  & 15 & 14 & 10 \\
            \midrule
            \multirow{2}{*}{Open-source ($\leq$7B)} 
                & Before & 27 & 29 & 16 \\
                & After  & 27 & 24 & 12 \\
            \bottomrule
        \end{tabular}
        }
    \end{table}
    
    Tab.~\ref{tab:d1-error-distribution} summarizes the error-aware prompting results for each group. Across all model families, adding the suffixes leads to consistent but only modest improvements in the proportion of correctly solved cases: for the closed-source models, the proportion of correct answers increases from 34\% to 43\%; for open-source models with more than 7B parameters, it increases from 42\% to 61\%; and for open-source models with at most 7B parameters, it increases from 28\% to 37\%.

    A more fine-grained view reveals several notable patterns: \textit{\textbf{1) Semantic Point Decoupling is highly prompt sensitive for larger models.}} For closed source models, Semantic Point Decoupling decreases from 26\% to 19\%, and for open source models with more than 7B parameters it decreases from 24\% to 10\%, indicating that many failures stem from weak coupling between intermediate reasoning and final coordinate prediction. \textit{\textbf{2) Spatial Reconstruction Failure is only moderately improved.}} Rates decline only slightly across model families, which suggests that reminders about occlusions and relative positions only partially address multiview three dimensional reasoning limitations. \textit{\textbf{3) Frame Transfer Failure remains essentially unchanged.}} Error-aware prompting has almost no effect on frame transfer errors, suggesting that this type of failure is difficult to alleviate through prompt-level modifications alone and may require stronger geometric priors or architectural changes in VLMs.

\subsection{Insights from Error-Aware Prompting}

    The error-aware prompting experiment provides several insights into the nature of the three error patterns and their implications for the design of future cross-view VLMs. Taken together, the results clarify which failures are mainly due to misaligned reasoning and which instead reflect deeper limitations in the underlying geometric representations.
    
    \textbf{First, the experiment exposes a clear distinction between prompt-sensitive and prompt-insensitive failures.} Semantic--Point Decoupling is strongly affected by the added instructions, especially in larger models, indicating that many of these errors arise from insufficient coupling between the reasoning process and the final coordinate prediction rather than from missing semantic or visual knowledge. Spatial Reconstruction Failure is only moderately reduced, which suggests that textual hints about occlusions and relative positions are not sufficient to overcome deficiencies in multi-view 3D reasoning. Frame Transfer Failure is essentially unchanged, implying that viewpoint-dependent coordinate frames are not robustly encoded and that this error type is difficult to address through prompt-level modifications alone.
    
    \textbf{Second, the effects of error-aware prompting differ systematically across model capacity regimes.} Open-source models with more than 7B parameters benefit the most, primarily through reductions in Semantic--Point Decoupling, which is consistent with the view that higher-capacity models already possess strong visual and semantic representations but under-utilize them without explicit guidance. Smaller models with at most 7B parameters show weaker gains and remain dominated by Spatial Reconstruction Failure and Frame Transfer Failure, pointing to more fundamental capacity and representation constraints beyond instruction following.
    
    \textbf{Third, error-aware prompting serves as a lightweight diagnostic tool rather than a complete solution.} Because it operates purely at the prompt level, without any architectural changes or finetuning, the degree to which a model improves under this intervention is informative about where its main bottlenecks lie. Large improvements indicate that failures are primarily due to misaligned attention and reasoning, whereas persistent error modes, such as Frame Transfer Failure, are more likely rooted in limitations of the models' underlying visual and geometric representations. 
    
    To better address the error patterns identified above, CrossPoint-378K can be leveraged as a supervised signal for CVPC, while also exploring richer training strategies and architectures that explicitly encode multi-view geometry. Combining advances in supervision and model design can lead to notable improvements in cross-view geometric reasoning.

\section{More Visualization}
\label{sec:more_visualization}

\subsection{Visualization of Prompts}
\begin{figure*}[t]
    \centering
    \includegraphics[width=\linewidth]{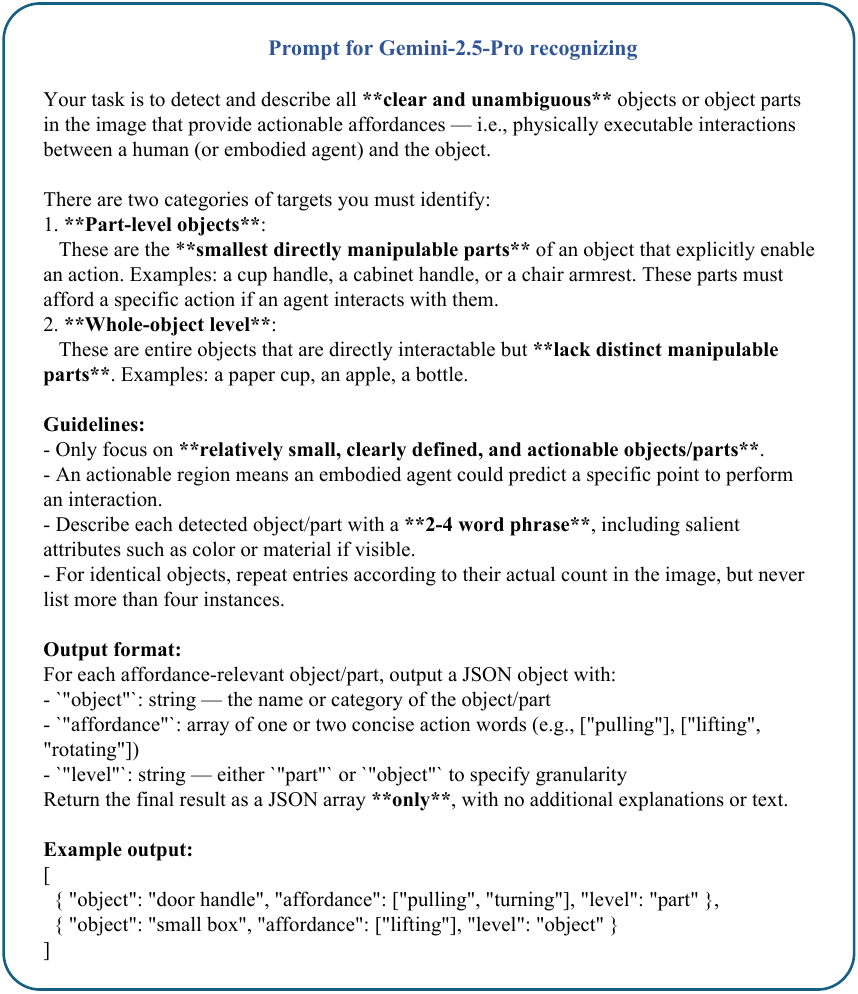}
    \caption{Prompt for Gemini-2.5-Pro recognizing.}
    \label{fig:prompt_recognize}
\end{figure*}

\begin{figure*}[t]
	\centering
	\includegraphics[width=\linewidth]{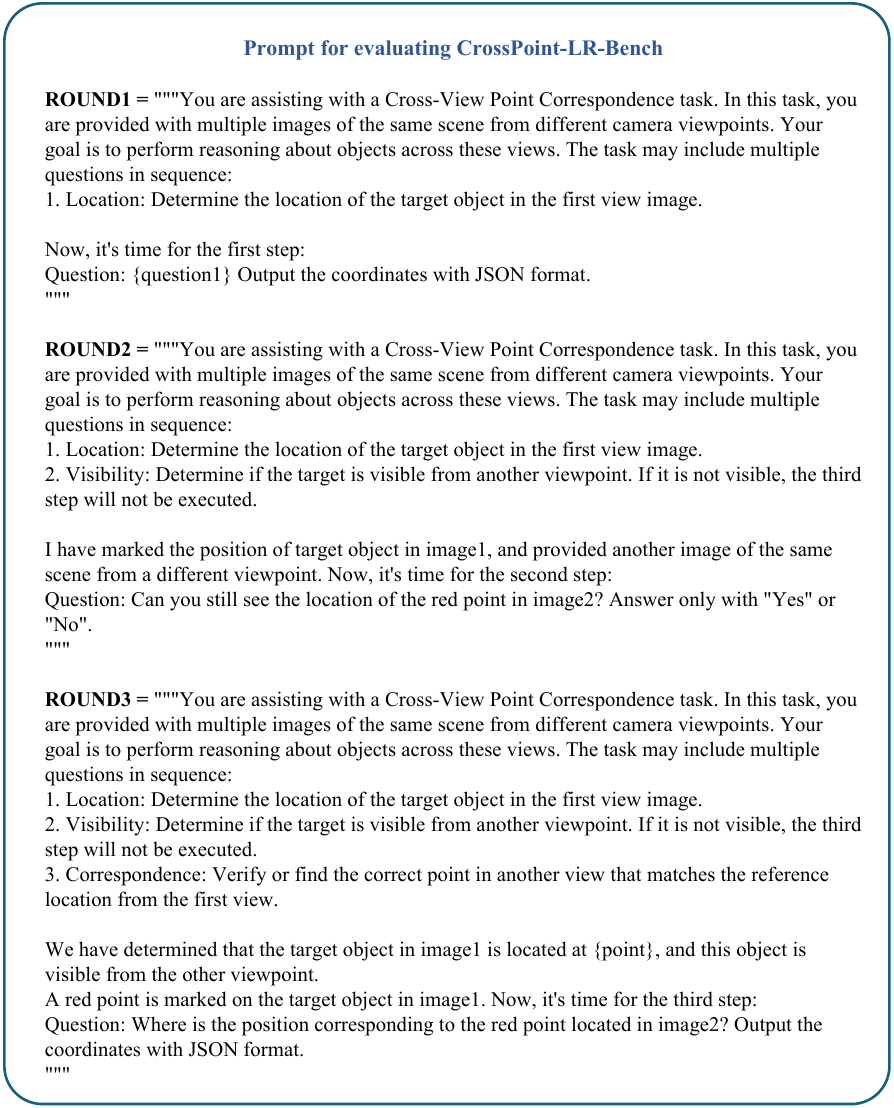}
    \caption{Prompt for evaluating CrossPoint-LR-Bench}
    \label{fig:prompt_lr}
\end{figure*}

We provide the Gemini-2.5-Pro prompt used in the dataset pipeline in Fig.~\ref{fig:prompt_recognize}. The evaluation prompts for CrossPoint-LR-Bench are shown in Fig.~\ref{fig:prompt_lr}.

\subsection{Visualization of CrossPoint-Bench}
\begin{figure*}[htbp]
    \centering
    \includegraphics[width=1.0\linewidth]{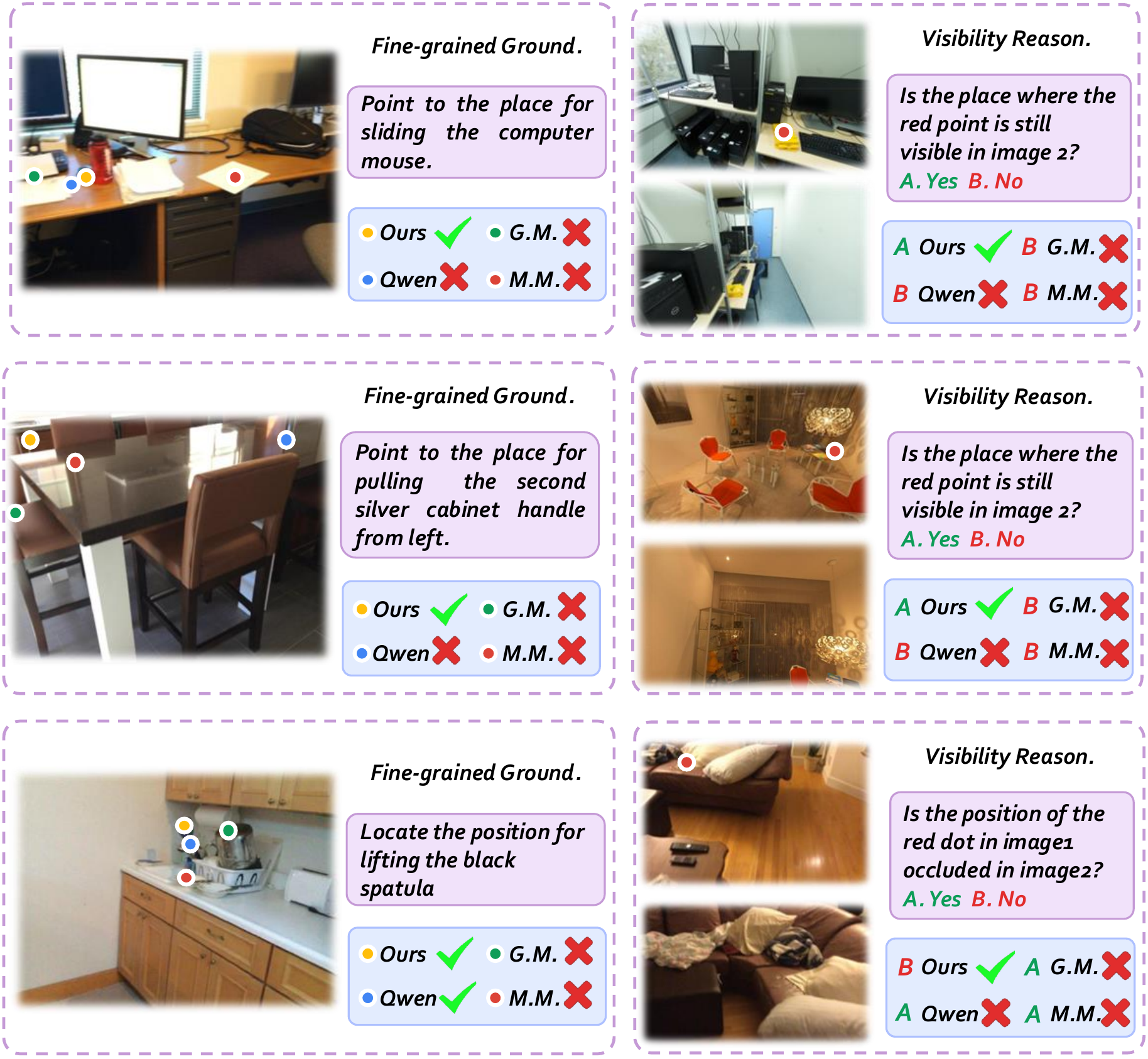}
    \caption{Visualization of CrossPoint-Bench (Fine-grained Grounding and Visibility Reasoning)}
    \label{fig:bench_12}
\end{figure*}

\begin{figure*}[htbp]
    \centering
    \includegraphics[width=1.0\linewidth]{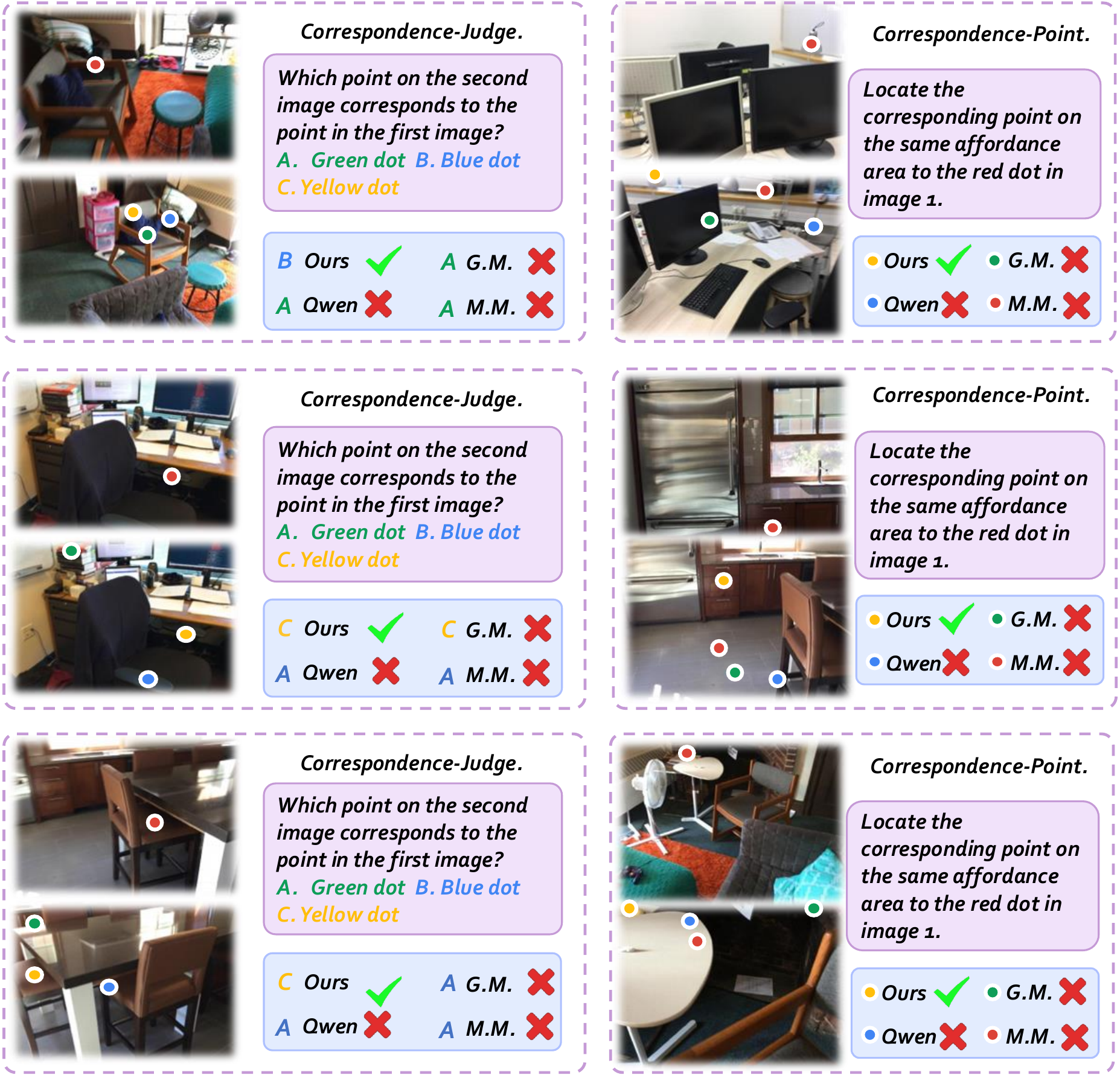}
    \caption{Visualization of CrossPoint-Bench (Correspondence-Judgment and Correspondence-Pointing)}
    \label{fig:bench_34}
\end{figure*}

\begin{figure*}[htbp]
    \centering
    \includegraphics[width=1.0\linewidth]{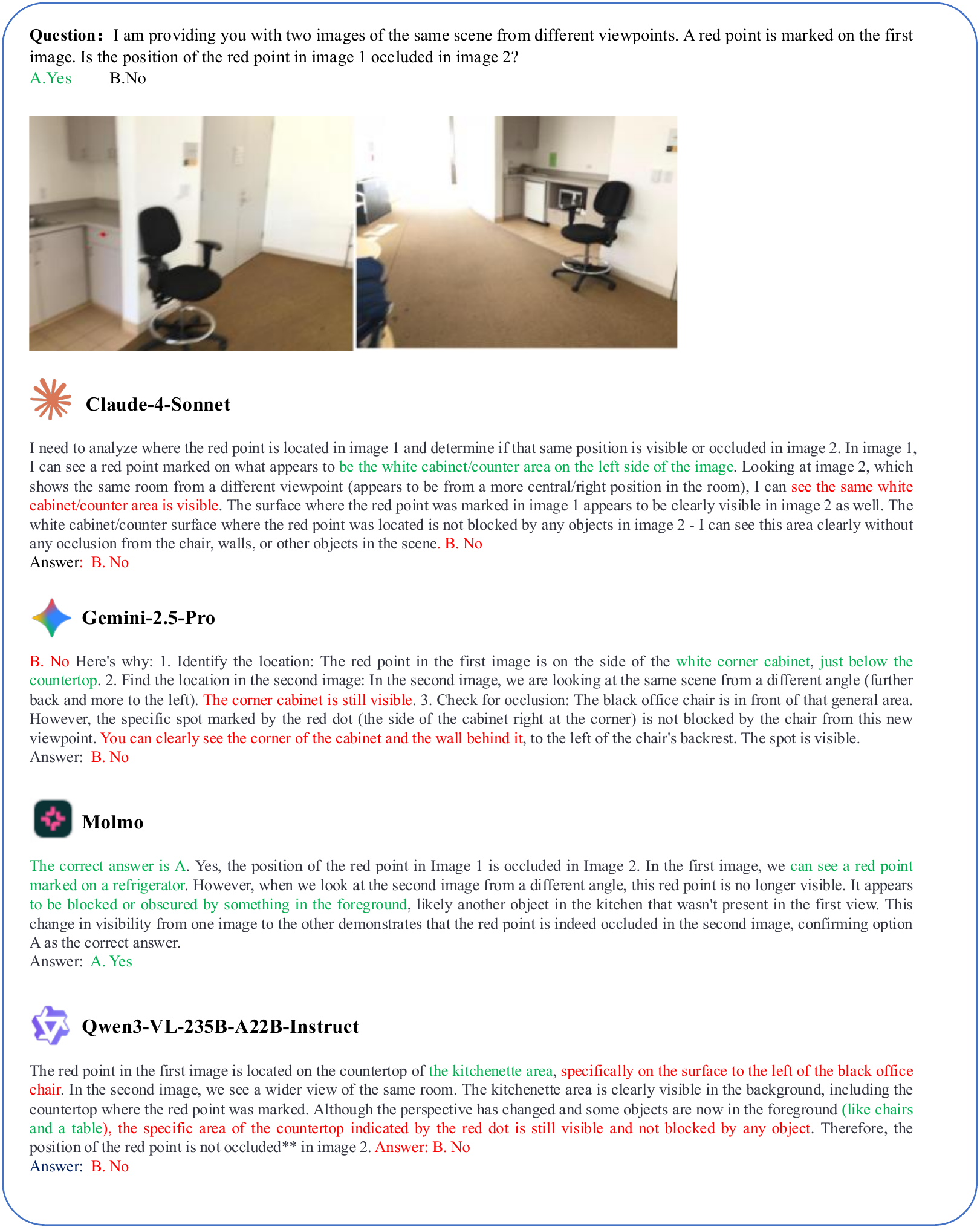}
    \caption{Visualization of CrossPoint-Bench (with reasoning process)}
    \label{fig:bench_cot_1}
\end{figure*}

\begin{figure*}[htbp]
    \centering
    \includegraphics[width=1.0\linewidth]{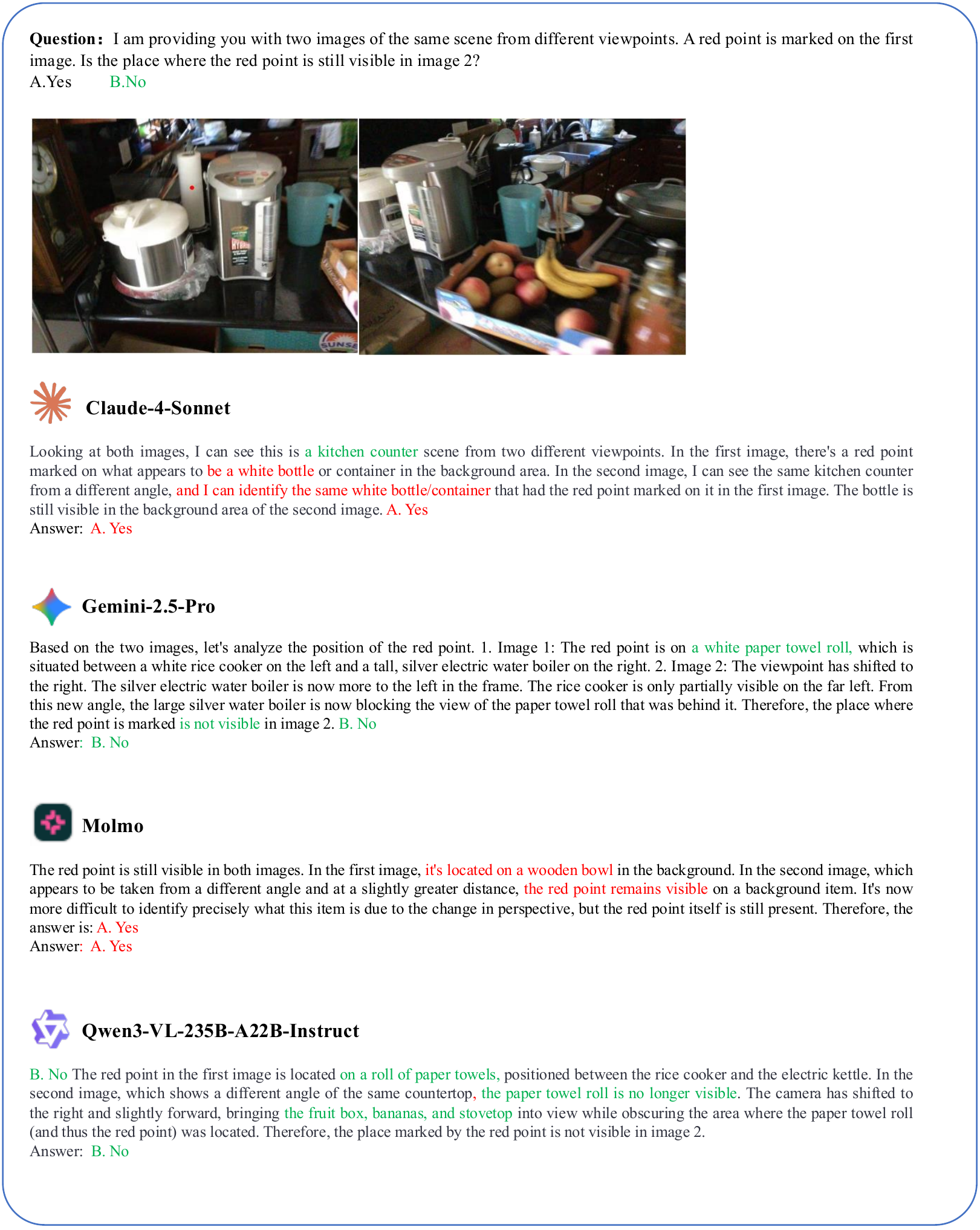}
    \caption{Visualization of CrossPoint-Bench (with reasoning process)}
    \label{fig:bench_cot_2}
\end{figure*}

\begin{figure*}[htbp]
    \centering
    \includegraphics[width=1.0\linewidth]{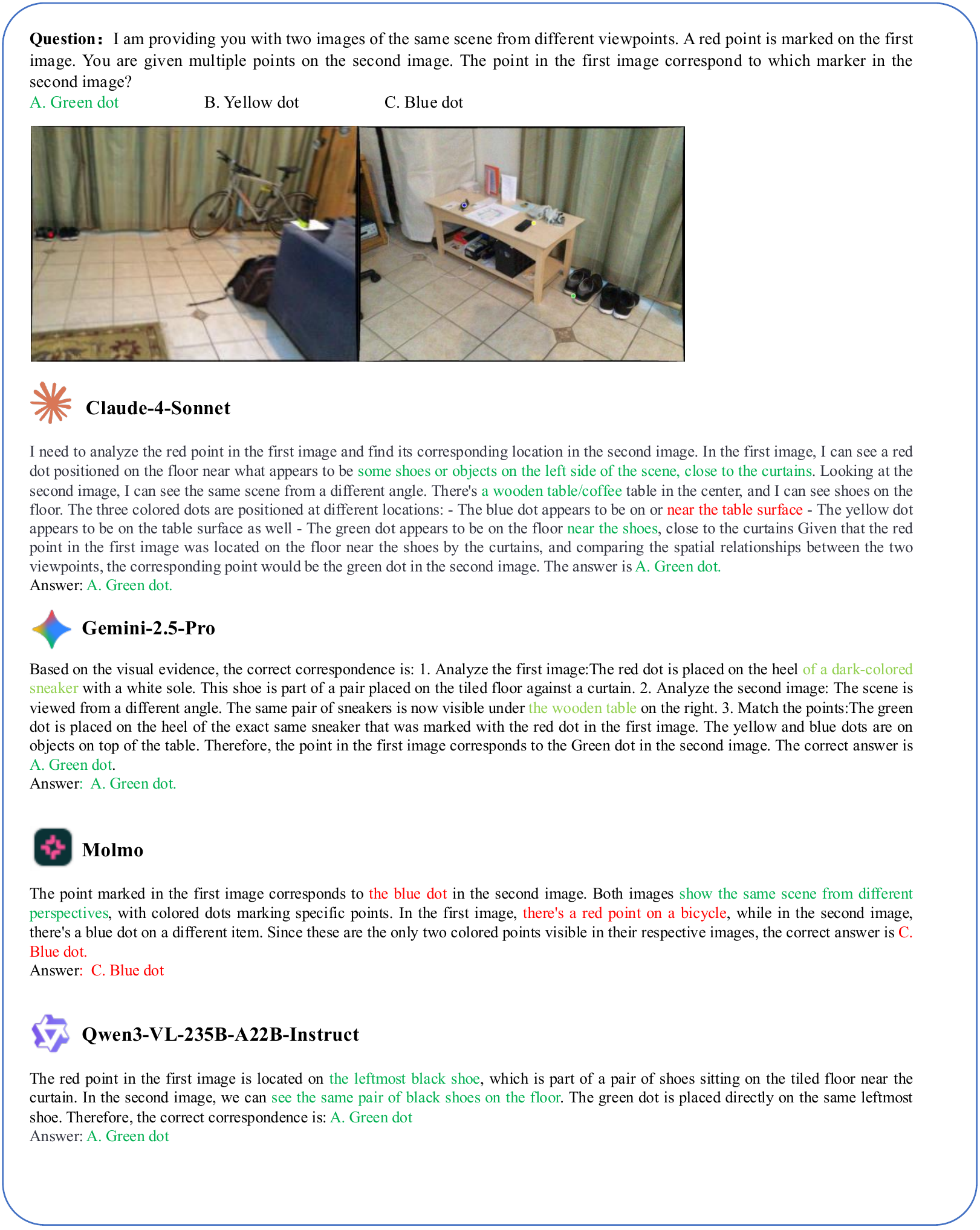}
    \caption{Visualization of CrossPoint-Bench (with reasoning process)}
    \label{fig:bench_cot_3}
\end{figure*}

\begin{figure*}[htbp]
    \centering
    \includegraphics[width=1.0\linewidth]{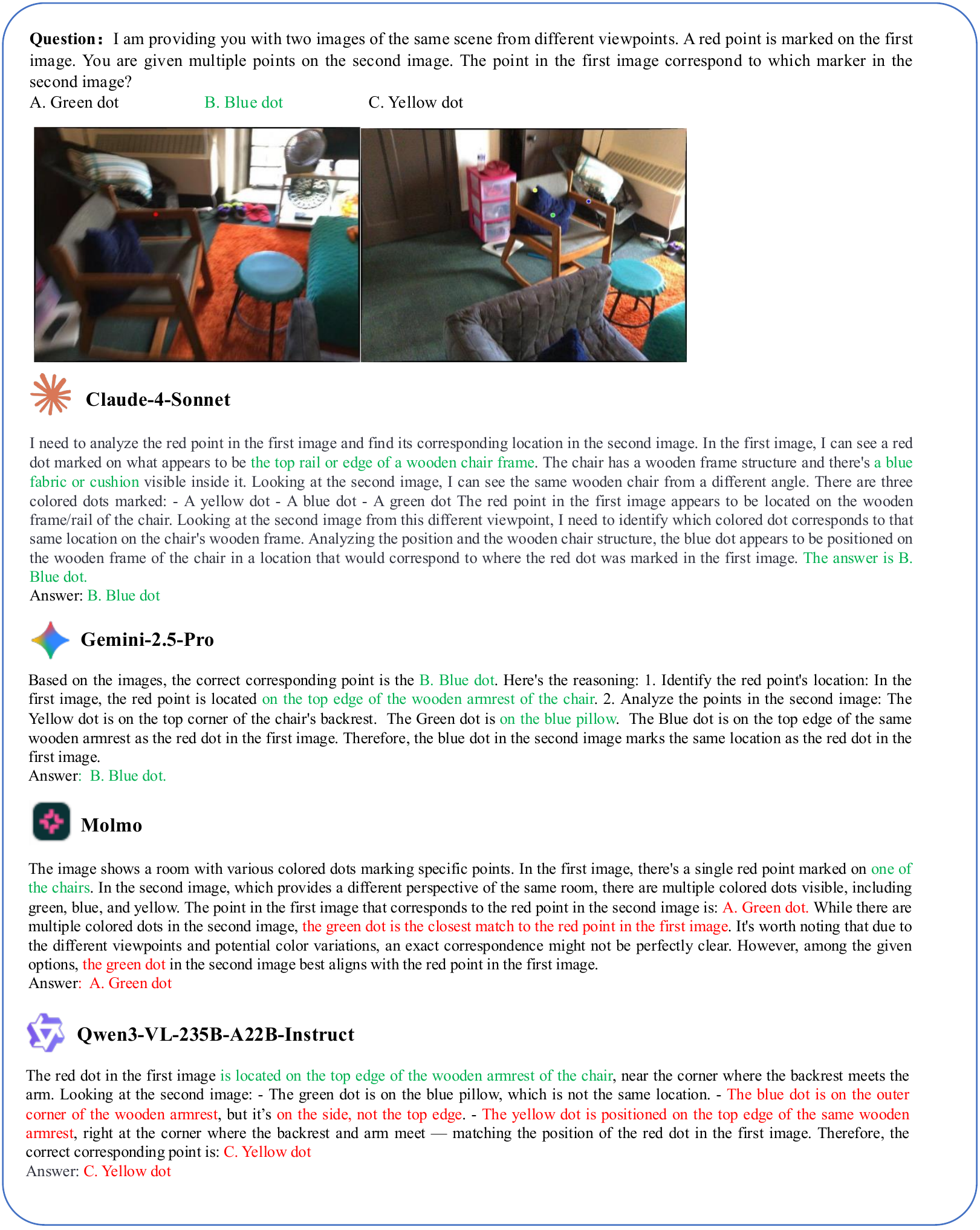}
    \caption{Visualization of CrossPoint-Bench (with reasoning process)}
    \label{fig:bench_cot_4}
\end{figure*}

We present representative examples from CrossPoint-Bench in Fig.~\ref{fig:bench_12},~\ref{fig:bench_34} illustrating comparisons between CroPond and other models. Additional examples in Fig.~\ref{fig:bench_cot_1},~\ref{fig:bench_cot_2},~\ref{fig:bench_cot_3},~\ref{fig:bench_cot_4} depict the detailed multi-step reasoning process.

\subsection{Visualization of CrossPoint-378K}
\begin{figure*}
	\centering
	\includegraphics[width=0.9\textwidth]{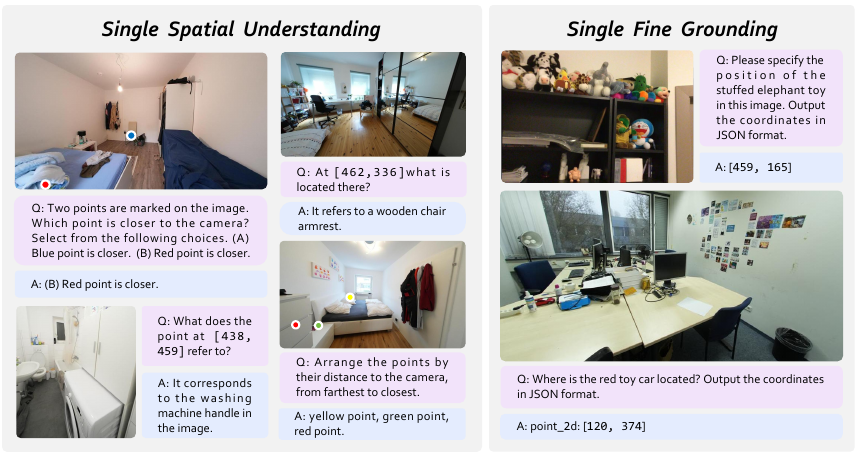}
	\includegraphics[width=0.9\textwidth]{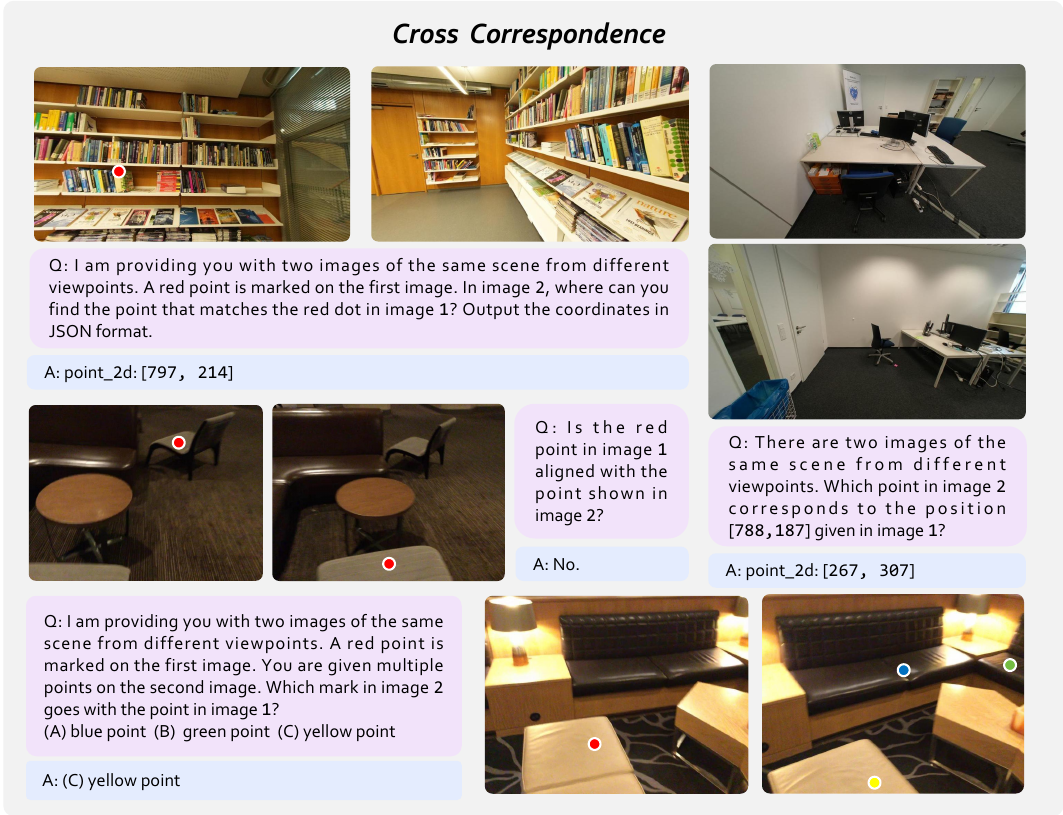}
    \caption{Visualization of CrossPoint-378K}
    \label{fig:CrossPoint-378K_1}
\end{figure*}

\begin{figure*}
	\centering
	\includegraphics[width=0.83\textwidth]{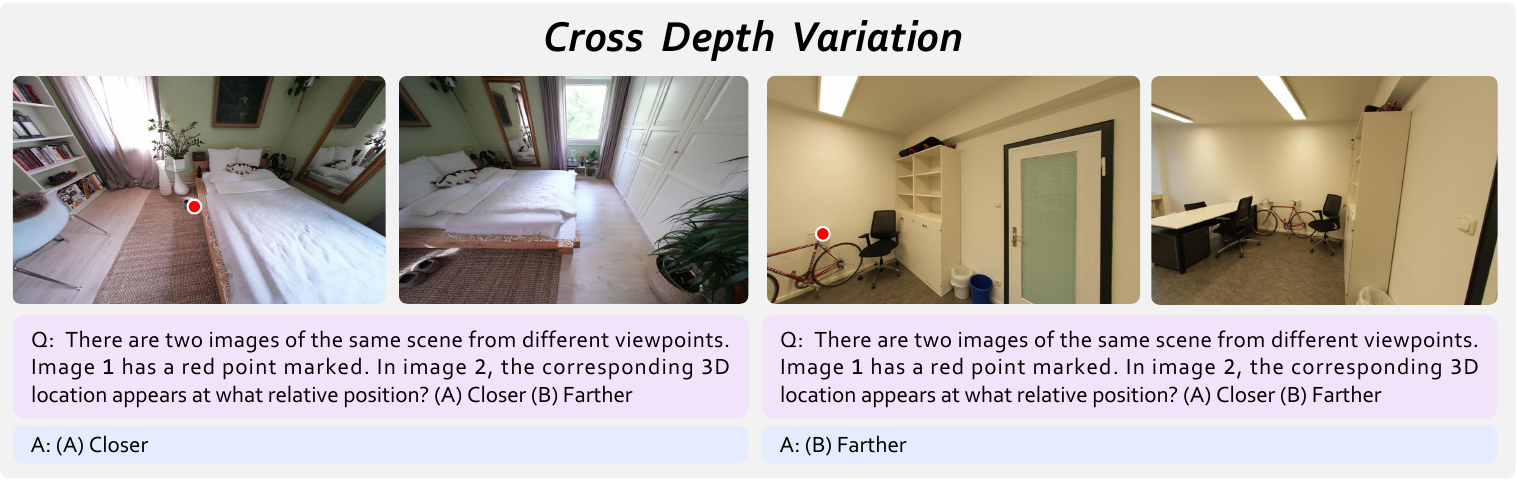}
	\includegraphics[width=0.83\textwidth]{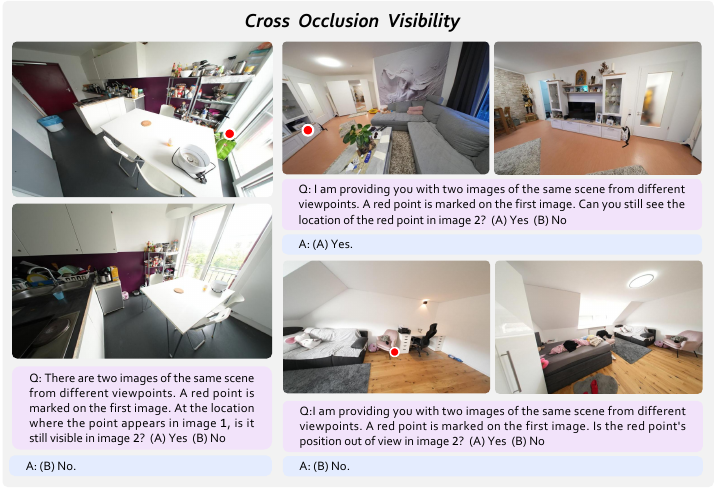}
	\includegraphics[width=0.83\textwidth]{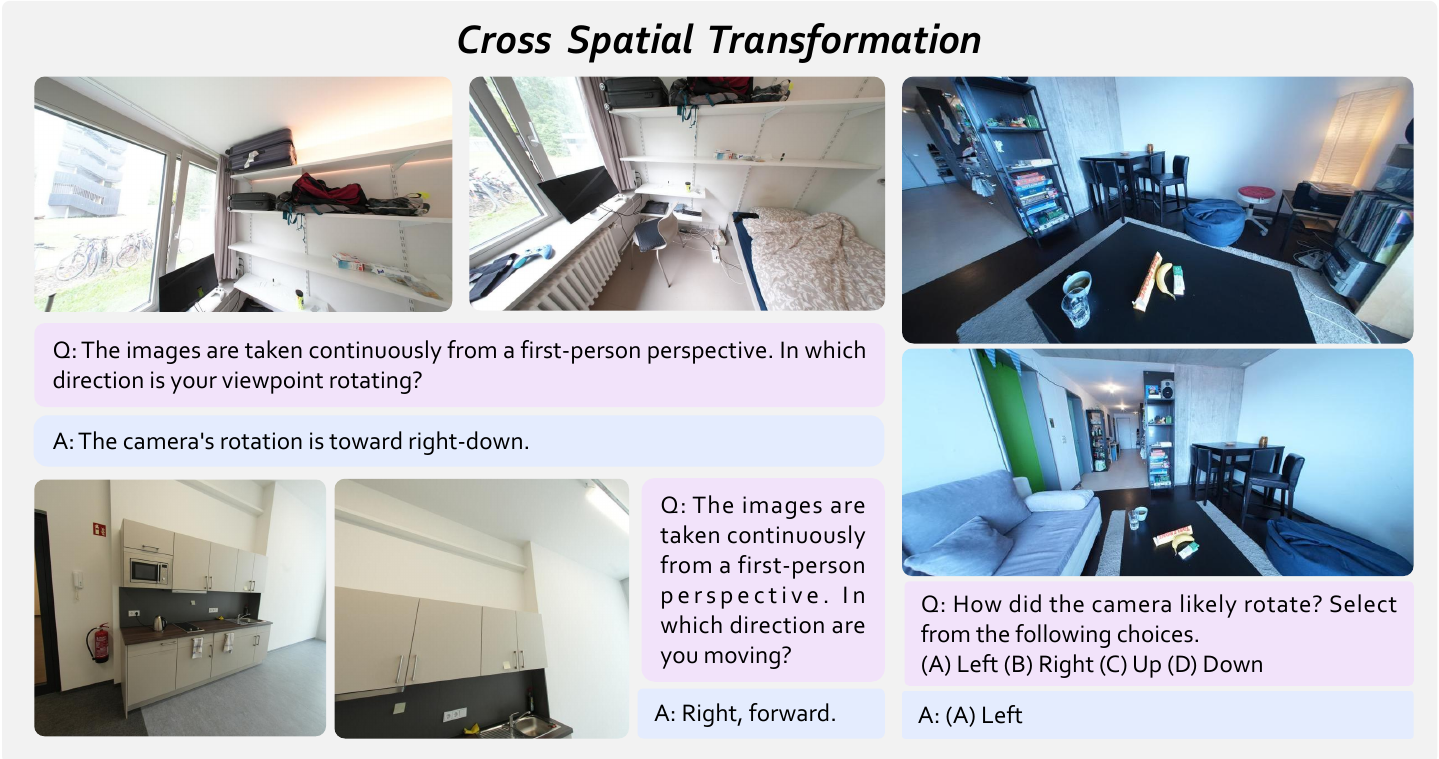}
    \caption{Visualization of CrossPoint-378K}
    \label{fig:CrossPoint-378K_2}
\end{figure*}

We show dataset examples in Fig.~\ref{fig:CrossPoint-378K_1} and Fig.~\ref{fig:CrossPoint-378K_2}, which span six distinct categories across both single-view and cross-view settings.

\section{Future Work}

CrossPoint-Bench provides a comprehensive benchmark for Cross-View Point Correspondence, and CroPond establishes the first strong baseline designed for the CVPC task in realistic indoor scenes, yet several promising directions remain open for future exploration.

\textbf{Advancing Training Paradigms for CVPC.}
While CroPond is trained using a straightforward supervised fine-tuning scheme, future work will explore more expressive optimization strategies aimed at strengthening multi-step spatial reasoning. Reinforcement learning with geometry-aware reward structures may enable the model to self-correct inaccurate correspondences and maintain long-horizon geometric consistency.

\textbf{From Perception to Planning.}
Current applications of CroPond in multi-agent systems remain limited to perceptual understanding. A key direction is to couple CVPC with task-planning modules, enabling agents to leverage precise cross-view point reasoning for more intelligent decision-making and coordinated interactions.

We release CrossPoint-Bench, CrossPoint-378K, and CroPond to support community-driven extensions and encourage broader exploration of CVPC.

\newpage

\clearpage

\newpage
\newpage

\end{document}